%% file: main.tex
\documentclass{article}

     \PassOptionsToPackage{numbers, compress}{natbib}

\usepackage[preprint]{neurips_2020}

\makeatletter
\renewcommand{\@noticestring}{%
  }
\makeatother

\usepackage[utf8]{inputenc} 
\usepackage[T1]{fontenc}    
\usepackage{hyperref}       
\usepackage{url}            
\usepackage{booktabs}       
\usepackage{amsfonts}       
\usepackage{nicefrac}       
\usepackage{microtype}      

\usepackage{todonotes}
\usepackage{placeins}
\usepackage{amsthm}
\usepackage{array} 
\usepackage{enumitem}

\usepackage{float}
\usepackage{amsmath}
\usepackage{amssymb}
\usepackage[ruled,vlined]{algorithm2e}
\usepackage{subcaption}
\usepackage{xcolor}
\definecolor{ForestGreen}{rgb}{0, 0.6, 0.33}

\newcommand{\R}{\mathbb{R}}
\long\def\comment#1{}

\newcommand{\zahra}[1]{\textcolor{red}{#1}}

\title{Solving Linear Inverse Problems \\ Using the Prior Implicit in a Denoiser}

\author{%
  Zahra Kadkhodaie\\
      Center for Data Science, and \\
        Howard Hughes Medical Institute\\
     New York University \\
    \texttt{zk388@nyu.edu}\\
   \And
  Eero P. Simoncelli\\
       Center for Neural Science, \\
       Courant Inst. of Mathematical Sciences, and\\
       Howard Hughes Medical Institute\\
       New York University\\
    \texttt{eero.simoncelli@nyu.edu}\\
   }

\begin{document}

\maketitle

\begin{abstract}
Prior probability models are a fundamental component of many image processing problems, but density estimation is notoriously difficult for high-dimensional signals such as photographic images. Deep neural networks have provided state-of-the-art solutions for problems such as denoising, which implicitly rely on a prior probability model of natural images. Here, we develop a robust and general methodology for making use of this implicit prior. We rely on a statistical result due to Miyasawa (1961), who showed that the least-squares solution for removing additive Gaussian noise can be written directly in terms of the gradient of the log of the noisy signal density. We use this fact to develop a stochastic coarse-to-fine gradient ascent procedure for drawing high-probability samples from the implicit prior embedded within a CNN trained to perform blind (i.e., with unknown noise level) least-squares denoising. A generalization of this algorithm to constrained sampling provides a method for using the implicit prior to solve any linear inverse problem, with no additional training.  We demonstrate this general form of transfer learning in multiple applications, using the same algorithm to produce state-of-the-art levels of unsupervised performance for deblurring, super-resolution, inpainting, and compressive sensing.
\end{abstract}

\section{Introduction}
\label{sec:intro}
\comment{ZK: We need a name for our method. It helps when other people want to mention it or compare to it}
\comment{EPS: Some potential acronyms:
Stochastic linear inverse via denoiser-embedded prior  SLIvaDEP
Inference via a denoiser-embedded prior  IVADEP (or IDEP)
Linear inverse via a denoiser-embedded prior  LIVADEP (or LIDEP)
Inference via a denoiser-implicit prior  IVADIP
Linear inverse via a denoiser-implicit prior  LIVADIP
I like LIVADEP. It's easy to pronounce and it conveys the core of the method. 
EPS: How about "SLIVADEP"?
Also note: system diagram is called Universal Linear Inverse Sampler (ULIS).
}
Many problems in image processing and computer vision rely, explicitly or implicitly, on prior probability models. Describing the full density of natural images is a daunting problem, given the high dimensionality of the signal space.
Traditionally, models have been developed by combining assumed symmetry properties (e.g., translation-invariance, dilation-invariance), with simple parametric forms (e.g., Gaussian, exponential, Gaussian mixtures), often within pre-specified transformed coordinate systems (e.g., Fourier transform, multi-scale wavelets). 
While these models have led to steady advances in problems such as denoising (e.g., \cite{Donoho95,Simoncelli96c,Moulin99,Hyvarinen99,Romberg01,Sendur02,Portilla03}), they are too simplistic to generate complex features that occur in our visual world, or to solve more demanding statistical inference problems.

In recent years, nearly all problems in image processing and computer vision have been revolutionalized by the use of deep Convolutional Neural Networks (CNNs). These networks are generally trained to perform tasks in a supervised fashion, without explicit use of an image prior. Their phenomenal performance suggests that they embed far more sophisticated priors than traditional methods. But these implicit priors arise from a combination of the distribution of the training data, the architecture of the network \cite{Ulyanov2020DeepImagePrior}, regularization terms included in the optimization objective, and the optimization algorithm.  Moreover, they are intertwined with the task for which they are optimized. 
\comment{***EPS: Redundant with first sentence of next par....
Our goal in this work is to extract and make use of a prior embedded in a trained CNN.}

Nearly a decade ago, a strategy known as ``plug and play'' was proposed 
for using the prior embedded in a denoiser to solve other inverse problems \cite{venkatakrishnan2013plug}, and a number of recent extensions \citep{Romano17, zhang2017learning, 8237460, 7744574, mataev2019deepred, teodoro2019image, 8616843, 8528509} have used this concept to develop MAP solutions for linear inverse problems. Generally, the objective is decoupled into a data fidelity term and a regularization, introducing a slack variable for use in a proximal optimization algorithm (e.g., ADMM). In place of a proximal operator, a denoiser is substituted for the gradient of the regularization term in the objective function. A parallel thread of research has focused on the use of generative models based on score matching \citep{alain2014regularized,Saremi18,Li19,Song19, bigdeli2020learning}. The connection between score matching \cite{hyvarinen2005estimation} and denoising autoencoders \cite{vincent2008extracting} was first shown in \cite{Vincent11}, by proving that the training criterion of a denoising autoencoder is equivalent to matching the score of the model and a Parzan density estimate of the data. This has been used as the basis for drawing samples from the prior implicit in the denoiser \cite{Song19}. In Section \ref{sec:related work} we elaborate on how these methods are related to our results.

\comment{
Generative Adversarial Networks (GANs) \cite{goodfellow2014generative} have been shown capable of synthesizing novel high-quality images, that may be viewed as samples of an implicit prior density model, and recent methods have been developed to use these priors in solving inverse problems
\cite{Shah2018,bora2017compressed}.
Although GANs can produce visually impressive synthetic samples, a number of results suggest that these samples are not representative of the full image density (a problem sometimes referred to as ``mode collapse'') \cite{richardson2018gans}. \zahra{Maybe we should also mention that GANs are generally tricky to train?}}

Here, we derive a general algorithm for solving linear inverse problems using the prior implicit in a denoiser. We start with a result from classical statistics \citep{Miyasawa61} that states that a denoiser that aims to minimize squared error of images corrupted by additive Gaussian noise may be interpreted as computing the gradient of the log of the density of noisy images. This result is related to score matching, but provides more direct relationship between least-squares optimal denoising and the embedded prior \cite{Raphan10}.
Starting from a random initialization, we develop a stochastic ascent algorithm that uses this denoiser-estimated gradient to draw high-probability samples from the embedded prior. The gradient step sizes and the amplitude of injected noise are jointly and adaptively controlled by the denoiser, producing robust and efficient convergence. More generally, we combine this procedure with constraints arising from any linear measurement of an image to draw samples from the prior conditioned on this measurement, thus providing a stochastic solution for any linear inverse problem.  We demonstrate that our method, using the prior implicit in a pre-trained state-of-the-art CNN image denoiser, produces high-quality results on image synthesis, inpainting, super-resolution, deblurring and recovery of missing pixels. We also apply our method to recovering images from projections onto a random low-dimensional basis, demonstrating results that greatly improve on those obtained using sparse union-of-subspace priors typically assumed in the compressive sensing literature. \footnote{A software impliementation of the sampling and linear inverse algorithms is available at\\ \url{https://github.com/LabForComputationalVision/universal_inverse_problem}}

\subsection{Image priors, manifolds, and noisy observations}
Digital photographic images lie in a high-dimensional space ($\R^N$, where $N$ is the number of pixels), and simple thought experiments suggest that they are concentrated on or near low-dimensional manifolds. For a given photograph, applying any of a variety of local continuous deformations (e.g., translations, rotations, dilations, intensity changes) yields a low-dimensional family of natural-appearing images.  These deformations follow complex curved trajectories in the space of pixels, and thus lie on a manifold.  In contrast, images generated with random pixels are almost always feature and content free, and thus not considered to be part of this manifold. We can associate with this a prior probability model, $p(x)$, by assuming that images within the manifold have constant or slowly-varying probability, while unnatural or distorted images (which lie off the manifold) have low or zero probability. 

Suppose we make a noisy observation of an image, $y = x + z$, where $x\in R^N$ is the original image drawn from $p(x)$, and  $z \sim \mathcal{N}(0, \sigma^2 I_N)$ is a sample of Gaussian white noise. The observation density $p(y)$ (also known as the {\em prior predictive density}) is related to the prior $p(x)$ via marginalization:
\begin{equation}
p(y) = \int p(y|x) p(x) dx = \int g(y-x) p(x) dx ,
\label{eq:measDist}
\end{equation}
where the noise distribution is
\[
g(z) = \frac{1}{(2\pi \sigma^2)^{N/2}} e^{-||z||^2/{2\sigma^2}} .
\]
Equation~(\ref{eq:measDist}) is in the form of a convolution, and thus $p(y)$ is a Gaussian-blurred version of the signal prior, $p(x)$. 
Moreover, the family of observation densities over different noise variances, $p_\sigma(y)$, forms a Gaussian scale-space representation of the prior~\citep{Koenderink84,Lindeberg94}, analogous to the temporal evolution of a diffusion process.

\subsection{Least squares denoising and CNNs}

Given a noisy observation, $y$, the least squares estimate (also called "minimum mean squared error", MMSE) of the true signal is well known to be the conditional mean of the posterior:  
\begin{equation}
\hat{x}(y) = \int x p(x|y) dx = \int x \frac{p(y|x) p(x)}{p(y)} dx
\label{eq:BLS}
\end{equation}
Traditionally, one obtains such estimators by choosing a prior probability model, $p(x)$ (often with parameters fit to sets of images), combining it with a likelihood function describing the noise, $p(y|x)$, and solving.  For example, the Wiener filter is derived by assuming a Gaussian prior in which variance falls inversely with spatial frequency~\citep{wiener1950extrapolation}.  Modern denoising solutions, on the other hand, are often based on discriminative training.  One expresses the estimation function (as opposed to the prior) in parametric form, and sets the parameters by minimizing the denoising MSE over a large training set of example signals and their noise-corrupted counterparts \citep{HelOr07,elad2006image,jain2009natural,burger2012image}. The size of the training set is virtually unlimited, since it can be constructed automatically from a set of photographic images, and does not rely on human labelling. 

Current state-of-the-art denoising results using CNNs are far superior, both numerically and visually, to results of previous methods \citep{zhang2017beyond,huang2017densnet,chen2017trainable}.
Recent work \citep{MohanKadkhodaie19b} demonstrates that these architectures can be simplified by removing all additive bias terms, with no loss of performance. The resulting {\em bias-free} networks offer two important advantages. First, they automatically generalize to all noise levels: a network trained on images with barely noticeable levels of noise can produce high quality results when applied to images corrupted by noise of any amplitude.  Second, they may be analyzed as adaptive linear systems, which reveals that they perform an approximate projection onto a low-dimensional subspace. In our context, we interpret this subspace as a tangent hyperplane of the image manifold at a specific location. Moreover, the dimensionality of these subspaces falls inversely with $\sigma$, and for a given noise sample, the subspaces associated with different noise amplitude are nested, with high-noise subspaces lying within their lower-noise counterparts. In the limit as the noise variance goes to zero, the subspace dimensionality grows to match that of the manifold at that particular point. 

\subsection{Exposing the implicit prior through Empirical Bayes estimation}
\label{Miyasawa}

The trained CNN denoisers mentioned above embed detailed prior knowledge of image structure.  Given such a denoiser, how can we obtain access to this implicit prior? Recent results have derived relationships between Score matching density estimates and denoising~\cite{Bengio13,saremi2019neural,Song19,li2019annealed}, and have used these relationships to make use of implicit prior information. Here, we exploit a much more direct but little-known result from the literature on Empirical Bayesian estimation. The idea was introduced in \cite{Robbins56}, extended to the case of Gaussian additive noise in \cite{Miyasawa61}, and generalized to many other measurement models in \cite{Raphan10}. For the Gaussian noise case, the least-squares estimate of Eq.~(\ref{eq:BLS}) may be rewritten as:
\begin{equation}
    \hat{x}(y) = y + \sigma^2 \nabla_{\!y} \log p(y) .
    \label{eq:miyasawa}
\end{equation}

The proof of this result is relatively straightforward. The gradient of the observation density expressed in Eq.~(\ref{eq:measDist}) is:
\begin{align*}
 \nabla_{\!y}\ p(y) &= 
 \frac{1}{\sigma^2} \int (x-y) g(y-x) p(x) dx \\
  &= \frac{1}{\sigma^2} \int (x-y) p(y,x) dx .   
 \end{align*}
Multiplying both sides by $\sigma^2/p(y)$ and separating the right side into two terms gives:
\begin{align}
\label{eq:miyasawa-grad}
\sigma^2 \frac{ \nabla_{\!y}\ p(y)}{p(y)}
&= \int x p(x|y) dx - \int y p(x|y) dx \\
&= \hat{x}(y) - y . \nonumber
\end{align}
Rearranging terms and using the chain rule to compute the gradient of the log gives Miyasawa's result, as expressed in Eq.~(\ref{eq:miyasawa}).

Intuitively, the Empirical Bayesian form in Eq.~(\ref{eq:miyasawa}) suggests that denoisers use a form of gradient ascent, removing noise from an observed signal by moving up a probability gradient.  But note that: 1) the relevant density is not the prior, $p(x)$, but the noisy {\em observation density}, $p(y)$; 2) the gradient is computed on the {\em log} density (the associated ``energy function''); and 3) the adjustment is not iterative - the optimal solution is achieved in a single step, and holds for any noise level, $\sigma$.

\section{Drawing high-probability samples from the implicit prior}
\label{sec:sampling}
Suppose we wish to draw a sample from the prior implicit in a denoiser. Equation~(\ref{eq:miyasawa-grad}) allows us to generate an image proportional to the gradient of $\log p(y)$ by computing the denoiser residual, $f(y) = \hat{x}(y) - y$. 
Previous work \citep{Saremi18,Song19} developed a related computation in a Markov chain Monte Carlo (MCMC) scheme, combining gradient steps derived from Score Matching and injected noise in a Langevin sampling algorithm to draw samples from a sequence of densities $p_\sigma(y)$, while reducing $\sigma$ in discrete steps, each associated with an appropriately trained denoiser. In contrast, starting from a random initialization, $y_0$, we aim to find a high-probability image (i.e., an image from the manifold) using a simpler and more efficient stochastic gradient ascent procedure. 

We compute gradients using the residual of a bias-free universal CNN denoiser, which automatically adapts to each noise level. On each iteration, we take a small step in the direction specified by the denoiser, which moves closer to the image manifold, thereby reducing the amplitude of the effective noise.  The reduction of noise is achieved by decreasing the amplitude in the directions orthogonal to the observable manifold while retaining the amplitude of image in the directions parallel to manifold which gives rise to synthesis of image content. As the effective noise decreases, the observable dimensionality of the image manifold increases~\cite{MohanKadkhodaie19b}, allowing the synthesis of detailed structures. Since the family of observation densities, $p_\sigma(y)$ forms a scale-space representation of $p(x)$, the algorithm may be viewed as an adaptive form of coarse-to-fine optimization~\citep{Lucas81,Kirckpatrick83,Geman84,Blake87}. Assuming the step sizes are adequately-controlled, the procedure converges to a point on the manifold.  
Figure \ref{fig:2Dvisualization} illustrates this process in two dimensions.

Each iteration operates by taking a deterministic step in the direction of the gradient (as obtained from the denoising function) and injecting some additional noise:
\begin{equation}
y_t = y_{t-1} + h_t f(y_{t-1}) + \gamma_t z_t ,
\end{equation}
where $f(y) = \hat{x}(y)-y$ is the residual of the denoising function, which is proportional to the gradient of $p(y)$, from Eq.~(\ref{eq:miyasawa-grad}).
The parameter $h_t\in[0,1]$ controls the fraction of the denoising correction that is taken, and $\gamma_t$ controls the amplitude of a sample of white Gaussian noise, $z_t \sim \mathcal{N}(0,I)$, whose purpose is to avoid getting stuck in local maxima.
The effective noise variance of image $y_t$ is:
\begin{equation}
\sigma_t^2 = (1-h_t)^2 \sigma_{t-1}^2 + \gamma_t^2 ,
\label{eq:sigma_t}
\end{equation}
where the first term is the variance of the noise remaining after the denoiser correction , and the second term is the variance arising from the injected noise. We assume the denoiser step is highly effective in estimating the direction and magnitude of noise, reducing the noise standard deviation by a factor of $(1-h_t)$ - this is justified in the case of recent DNN denoisers, as shown empirically in \citep{MohanKadkhodaie19b}.

To ensure convergence, we require the effective noise variance on each time step to be reduced, despite the injection of additional noise. For this purpose, we introduce a parameter $\beta\in[0,1]$ to control the proportion of injected noise ($\beta=1$ indicates no noise), and enforce the convergence by requiring that:
\begin{equation}
\sigma_t^2 = (1-\beta h_t)^2 \sigma_{t-1}^2 .
\label{eq:sigma_schedule}
\end{equation}
Combining this with Eq.~(\ref{eq:sigma_t}) yields an expression for $\gamma_t$ in terms of $h_t$:
\begin{eqnarray}
\gamma_t^2 &=& \left[(1- \beta h_t)^2 - (1-h_t)^2\right] \sigma_{t-1}^2 \nonumber \\
           &=& \left[(1- \beta h_t)^2 - (1-h_t)^2\right] \left\|f(y_{t-1})\right\|^2/N ,
\label{eq:iteration}
\end{eqnarray}
where the second line assumes that the magnitude of the denoising residual provides a good estimate of the effective noise standard deviation, as was found in \cite{MohanKadkhodaie19b}.
This allows the denoiser to adaptively control the gradient ascent step sizes, reducing them as the $y_t$ approaches the manifold (see Figure~\ref{fig:2Dvisualization} for a visualization in 2D). This automatic adjustment results in efficient and reliable convergence, as demonstrated empirically in Figure \ref{fig:convergence_expected}.

We found that initial implementations with a small constant fractional step size $h_t=h_0$ produced high quality results, but required many iterations. Inuitively, step sizes that are a fixed proportion of the distance to the manifold lead to exponential decay - a form of Zeno's paradox.  To accelerate convergence, we introduced a schedule for increasing the step size proportion according to  
$h_t = \frac{h_0 t}{1 + h_0(t-1)}$, starting from $h_0 \in [0,1]$. Note that the injected noise amplitude, $\gamma_t$, is adjusted using Eq.~(\ref{eq:iteration}), maintaining convergence, albeit at a slower rate.
The sampling algorithm is summarized below (Algorithm 1), and is laid out in a block diagram in Figure~\ref{fig:BlockDiagram} in the appendix.  Example convergence behavior is shown in Figure~\ref{fig:convergence_expected}.

\begin{algorithm}
    \SetAlgoLined
     parameters: $\sigma_0$, $\sigma_L$, $h_0$, $\beta$ \\
     initialization: $t=1$, \ draw $y_{0} \sim \mathcal{N}(0.5,\sigma_0^2 I)$\\
     \While{$ \sigma_{t-1} \leq \sigma_L $}{
     $h_t = \frac{h_0 t}{1 + h_0 (t-1)}$\;
     $d_t = f(y_{t-1})$\;
     $\sigma_{t}^2 =\frac{||d_t||^2}{N}$\;    
     $\gamma_t^2 = \left((1-\beta h_t)^2 - (1-h_t)^2\right) \sigma_{t}^2$\;
     Draw $ z_t \sim \mathcal{N}( 0,I)$\;               
     $y_{t} \leftarrow y_{t-1} + h_t d_t + \gamma_t z_t$\;
     $t \leftarrow t+1$
       {
      }
     }
     \caption{Coarse-to-fine stochastic ascent method for sampling from the implicit prior of a denoiser, using denoiser residual $f(y)=\hat{x}(y)-y$. }
    \label{alg:sampling}
\end{algorithm}   

\subsection{Image synthesis examples}
\label{sec:synthesis}

We examined the properties of images synthesized using this stochastic coarse-to-fine ascent algorithm. For a denoiser, we used BF-CNN~\citep{MohanKadkhodaie19b}, a bias-free variant of DnCNN~\citep{zhang2017beyond}. We trained the same network on three different datasets: $40 \times 40$ patches cropped from Berkeley segmentation training set \cite{MartinFTM01}, in color and grayscale, and MNIST dataset \cite{lecun-mnisthandwrittendigit-2010} (see Appendix \ref{sec:bfcnn} for further details).
We obtained similar results (not shown) using other bias-free CNN denoisers (see \cite{MohanKadkhodaie19b}).
For the sampling algorithm, we chose parameters $\sigma_0=1, \sigma_L=0.01$, and $h_0=0.01$ for all experiments.

\begin{figure}
\centering
\begin{tabular}{c}
  \includegraphics[width=.9\linewidth]{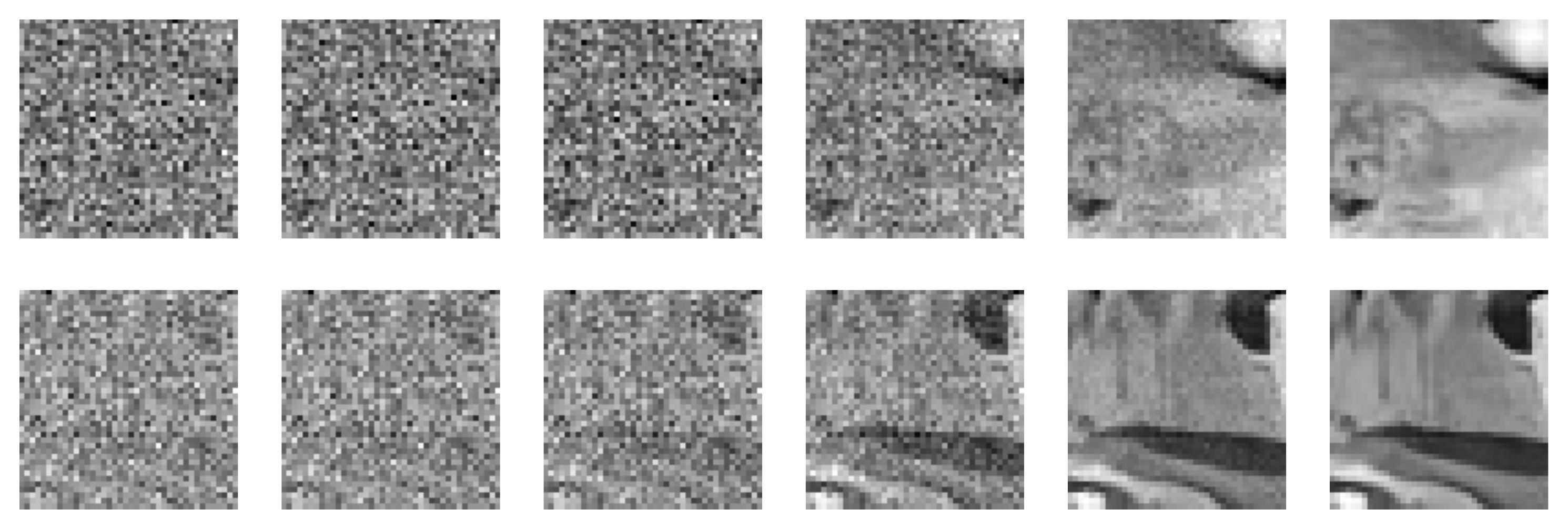}\\
  \includegraphics[width=.89\linewidth]{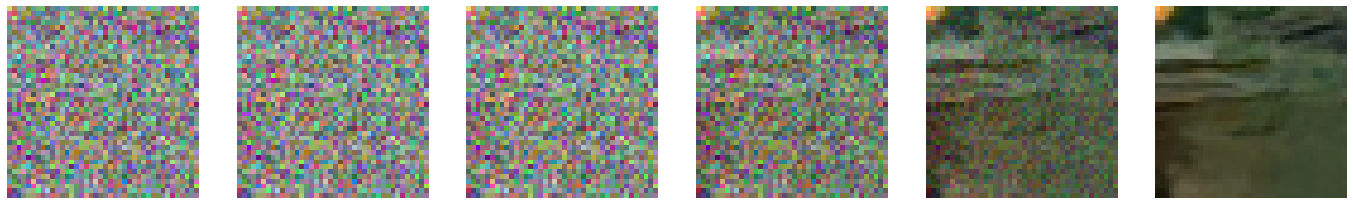}\\[0.2ex]
   \includegraphics[width=.89\linewidth]{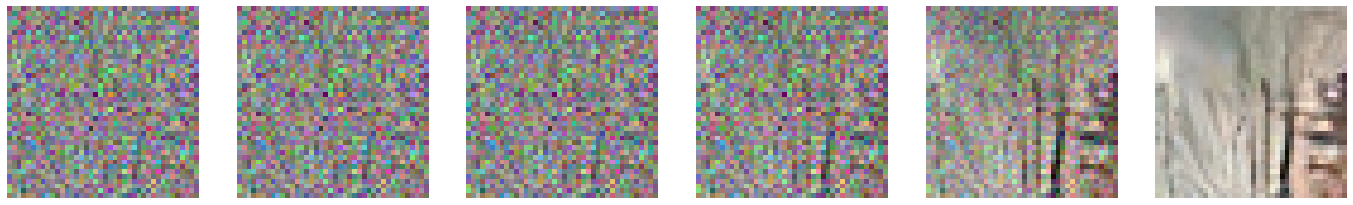}\\
\end{tabular}
\caption{Sampling from the implicit prior embedded in BF-CNN trained on grayscale (two top rows) and color (two buttom rows) Berkeley segmentation datasets. Each row shows a sequence of images, $y_t, t=1,9,17,25,\ldots$, from the iterative sampling procedure, with different initializations, $y_0$, and no added noise ($\beta=1$).
} 
\label{fig:synthesis_trajectory}
\end{figure}

\begin{figure}
\centering
\def\fsz{0.48\linewidth}
\includegraphics[width=\fsz]{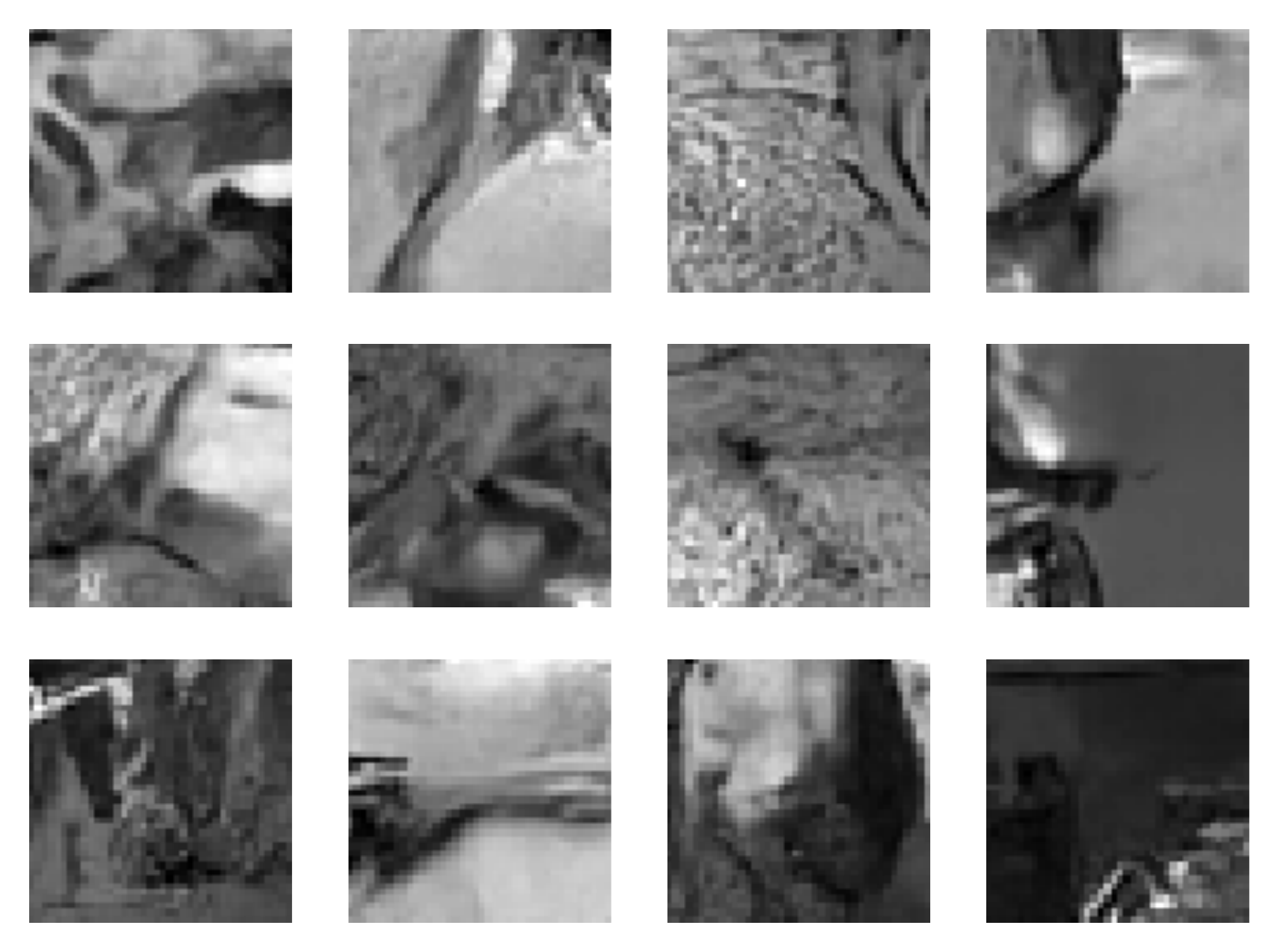}
\hfill
\includegraphics[width=\fsz]{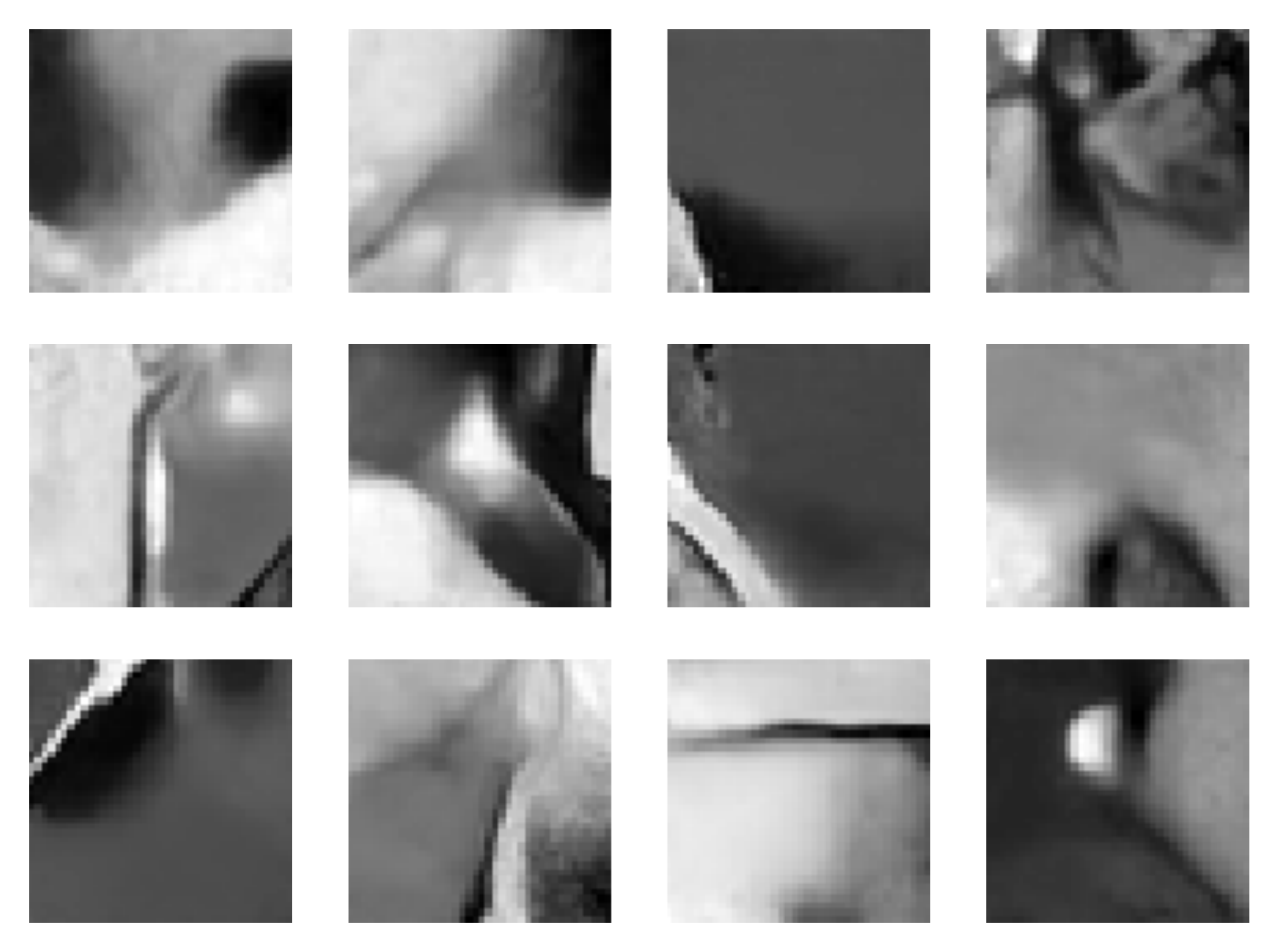} \\[-0.5ex]
\caption{Samples arising from different inializations, $y_0$. Left: A moderate level of noise ($\beta=0.5$) is injected in each iteration. Right: A high level of injected noise ($\beta=0.1$).
} 
\label{fig:synthesis_samples}
\end{figure}

Figure~\ref{fig:synthesis_trajectory} illustrates the iterative generation of four images, starting from different random initializations, $y_0$, with no additional noise injected (i.e, $\beta=1$), demonstrating the way that the algorithm amplifies and "hallucinates" structure found in the initial (noise) images.
Convergence is typically achieved in less than 40 iterations with stochasticity disabled ($\beta = 1$). The left panel in Figure~\ref{fig:synthesis_samples} shows samples drawn with different initializations, $y_0$, using a moderate level of injected noise ($\beta=0.5$). Images contain natural-looking features, with sharp contours, junctions, shading, and in some cases, detailed texture regions. 
The right panel in Figure~\ref{fig:synthesis_samples}
shows a set of samples drawn with more substantial injected noise ($\beta=0.1$). The additional noise helps to avoid local maxima, and arrives at 
images that are smoother and higher probability, 
but still containing sharp boundaries. As expected, the additional noise also lengthens convergence time (see Figure~\ref{fig:convergence_expected}). 
Figure \ref{fig:mnist_beta} shows a set of samples drawn from the implicit prior of a denoiser trained on the MNIST dataset of handwritten digits.

\section{Solving  linear inverse problems using the implicit prior}

Many applications in signal processing can be expressed as deterministic linear inverse problems -
deblurring, super-resolution, estimating missing pixels (e.g., inpainting), and compressive sensing are all examples.
Given a set of linear measurements of an image, $x^c = M^T x$, where $M$ is a low-rank measurement matrix, one attempts to recover the original image. In Section \ref{sec:sampling}, we developed a stochastic gradient-ascent algorithm for obtaining a high-probability sample from $p(x)$. Here, we modify this algorithm to solve for a high-probability sample from the conditional density $p(x|M^T x = x^c)$.

\subsection{Constrained sampling algorithm}

Consider the distribution of a noisy image, $y$, conditioned on the linear measurements, $x^c=M^Tx$.
Without loss of generality, we assume the columns of the matrix M are orthogonal unit vectors (if not, we  can re-parameterize to an equivalent constraint using the SVD). In this case, $M$ is the pseudo-inverse of $M^T$, and that matrix $MM^T$ can be used to project an image onto the measurement subspace. 
Using Bayes' rule, we write the conditional density of the noisy image conditioned on the linear measureement as
$$
p(y|x^c) = p(y^c, y^u|x^c) = p(y^u|y^c,x^c)p(y^c|x^c) = p(y^u|x^c)p(y^c|x^c)
$$
where $y^c = M^T y$, and $y^u = \bar{M}^T y$ (the  projection of $y$ onto the orthogonal complement of $M$).  \comment{Without loss of generality, we assume the measurement matrix has singular values that are equal to one (i.e., columns of $M$ are orthogonal unit vectors, and thus $M^TM = I$)\footnote{The assumption that $M$ has orthogonal columns does not restrict our solution. Assume an arbitrary linear constraint, $W^Tx = x_w$, and  the singular value decomposition $W=USV^T$. Setting $M=U$, the constraint holds if and only if $M^T x = x_c$, where $x_c = S^{\#} V^T x_w$. The columns of $M$ are orthogonal unit vectors, $M^TM =I$, and $MM^T$ is a projection matrix.}. It follows that $M$ is the pseudo-inverse of $M^T$, and that matrix $MM^T$ can be used to project an image onto the measurement subspace.}  As with the algorithm of Section \ref{sec:sampling}, we wish to obtain a local maximum of  this function using stochastic coarse-to-fine gradient ascent. Applying the operator $\sigma^2 \nabla \log(\cdot)$ yields
$$
\sigma^2 \nabla_{\!y} \log p(y|x^c) =
  \sigma^2 \nabla_{\!y} \log p(y^u|x^c) + \sigma^2 \nabla_{\!y} \log p(y^c|x^c) .
$$
The second term is the gradient of the observation noise distribution, projected into the measurement space.  If this is Gaussian with variance $\sigma^2$, it reduces to $M(y^c - x^c)$. 
The first term is the gradient of a function defined only within the subspace orthogonal to the measurements, and thus can be computed by projecting the measurement subspace out of the full gradient.  Combining these gives:
\begin{align}
\sigma^2 \nabla_{\!y} \log p(y) &=
(I-MM^T) \sigma^2 \nabla_{\!y} \log p(y) + M(x^c-y^c) \nonumber \\
  &= (I-MM^T) f(y) + M(x^c- M^Ty) .
  \label{eq:constrainedGradient}
\end{align}

Thus, we see that the gradient of the conditional density is partitioned into two orthogonal components, capturing the gradient of the (log) noisy density, and the  deviation from the constraints, respectively. 
To draw a high-probability sample from $p(x|x^c)$, we use the same algorithm described in Section~\ref{sec:sampling}, 
substituting Eq.~(\ref{eq:constrainedGradient}) for the deterministic update vector, $f(y)$ (see Algorithm 2, and Figure \ref{fig:BlockDiagram}). 
\comment{*** make consistent use of leftarrow and equalities. Can we make main equation indicate that noise term includes factor of ||f|| ? maybe elim sigma_t-1, and write ||d_t||/sqrt(N) into update rule?}
    \begin{algorithm}
    \SetAlgoLined
     parameters: $\sigma_0$, $\sigma_L$, $h_0$, $\beta$, $M$, $x^c$\\
     initialization: t=1; draw $ y_{0} \sim \mathcal{N}(0.5(I-MM^T)e+Mx^c,\ \sigma_0^2 I)$\\
     \While{$ \sigma_{t-1} \leq \sigma_L $}{
       $h_t  = \frac{h_0 t}{1 + h_0 (t-1)}$\;
       $d_t = (I-MM^T) f(y_{t-1}) + M(x^c - M^T y_{t-1})$\;
       $\sigma_{t}^2 = \frac{|| d_t ||^2}{N}$\;
       $\gamma_t^2 = \left((1-\beta h_t)^2 -(1-h_t)^2\right)\sigma_t^2$\;
       Draw $ z_{t} \sim \mathcal{N}(0,I)$\;
       $y_{t} \leftarrow y_{t-1} +  h_t d_t + \gamma_t z_t$\;
       $t \leftarrow t+1$
       {
      }
     }
     \caption{Coarse-to-fine stochastic ascent method for sampling from $p(x|M^Tx = x^c)$, based on the residual of a denoiser, $f(y)=\hat{x}(y)-y$. Note: $e$ is an image of ones.}
     \label{algorithm2}
    \end{algorithm}

\subsection{Linear inverse examples}            
We demonstrate the results of applying our method to several linear inverse problems. The same algorithm and parameters are used on all problems - only the measurement matrix $M$ and measured values $M^T x$ are altered. In particular, as in section \ref{sec:synthesis}, we used  BF-CNN \cite{MohanKadkhodaie19b}, and chose parameters  $\sigma_0=1, \sigma_L=0.01, h_0=0.01, \beta=0.01$. 
\comment{Note that this method is general and not limited to these specific tasks. It can be used to find a solution for any linear inverse problem, as long as the measurement model is known. }
For each example, we show a row of original images ($x$), a row of direct least-squares reconstructions ($MM^Tx$), and a row of restored images generated by our algorithm. For these applications, comparisons to ground truth are not particularly meaningful, at least when the measurement matrix is of very low rank.  In these cases, the algorithm relies heavily on the prior to "hallucinate" the missing information, and the goal is not so much to reproduce the original image, but to create an image that looks natural while being consistent with the measurements. Thus, the best measure of performance is a judgement of perceptual quality by a human observer.

\begin{figure}
\def\fsz{0.15\linewidth} 
\centering
\includegraphics[width=\fsz]{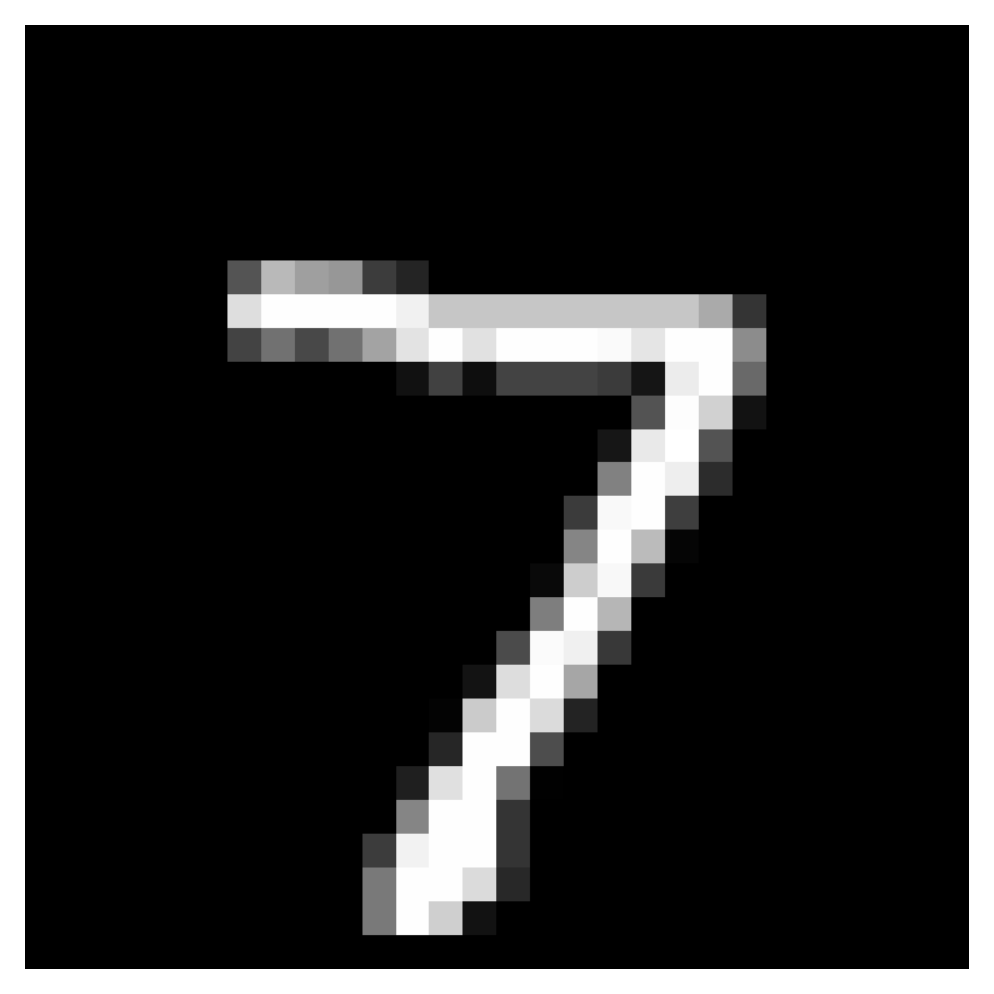} \hspace{1ex} 
\includegraphics[width=\fsz]{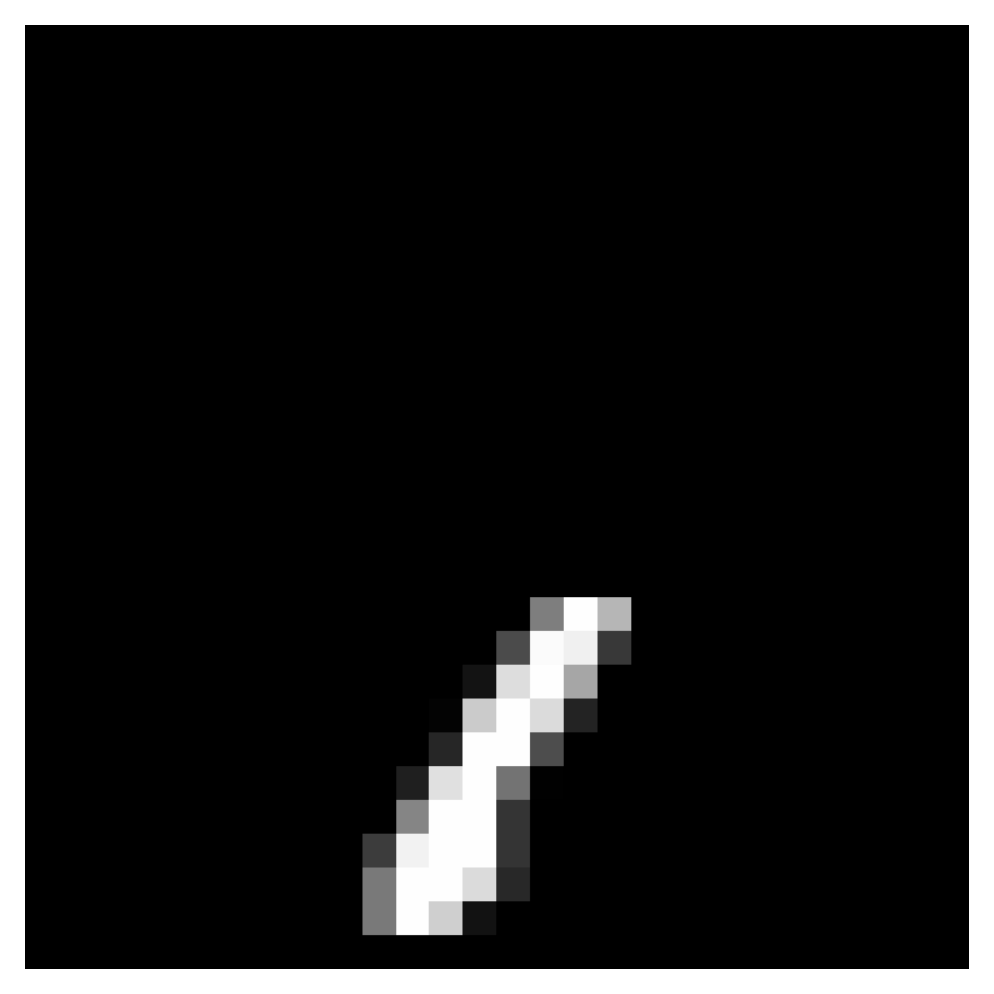} \hspace{3ex}
\includegraphics[width=\fsz]{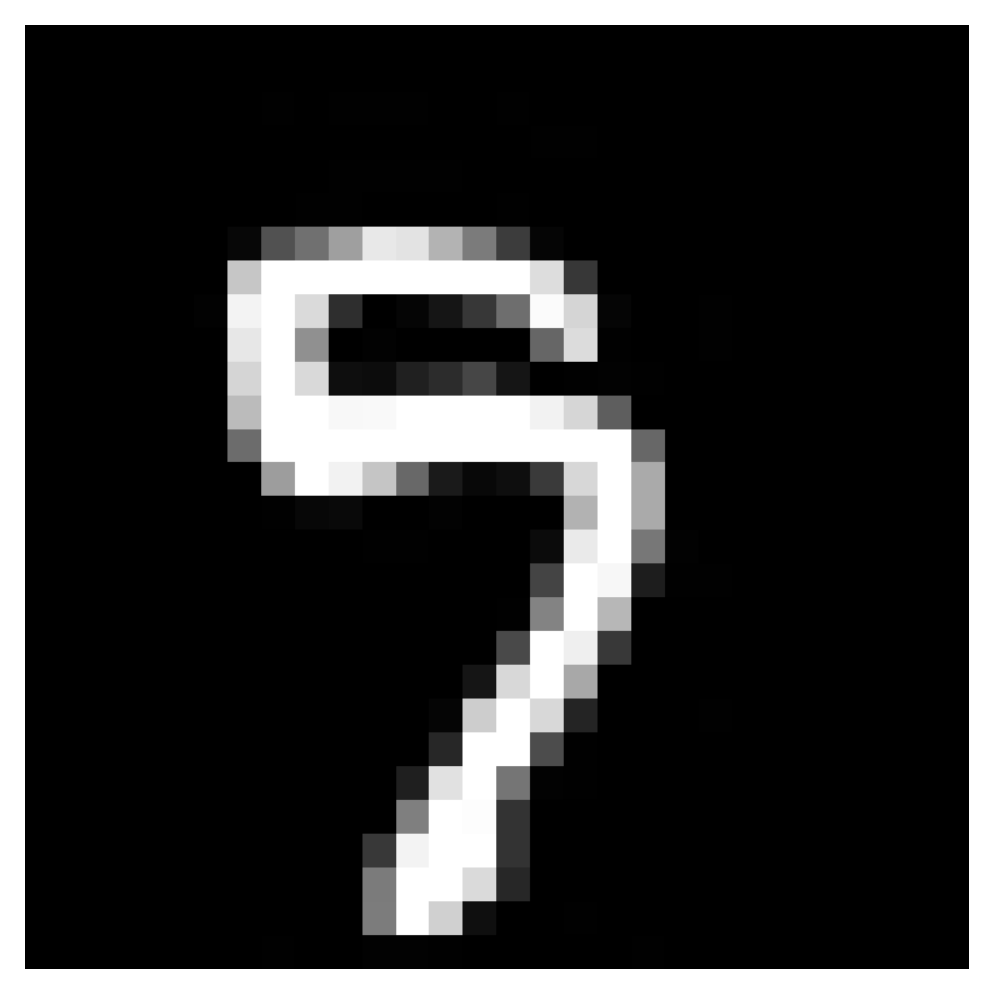} \hspace{1ex}
\includegraphics[width=\fsz]{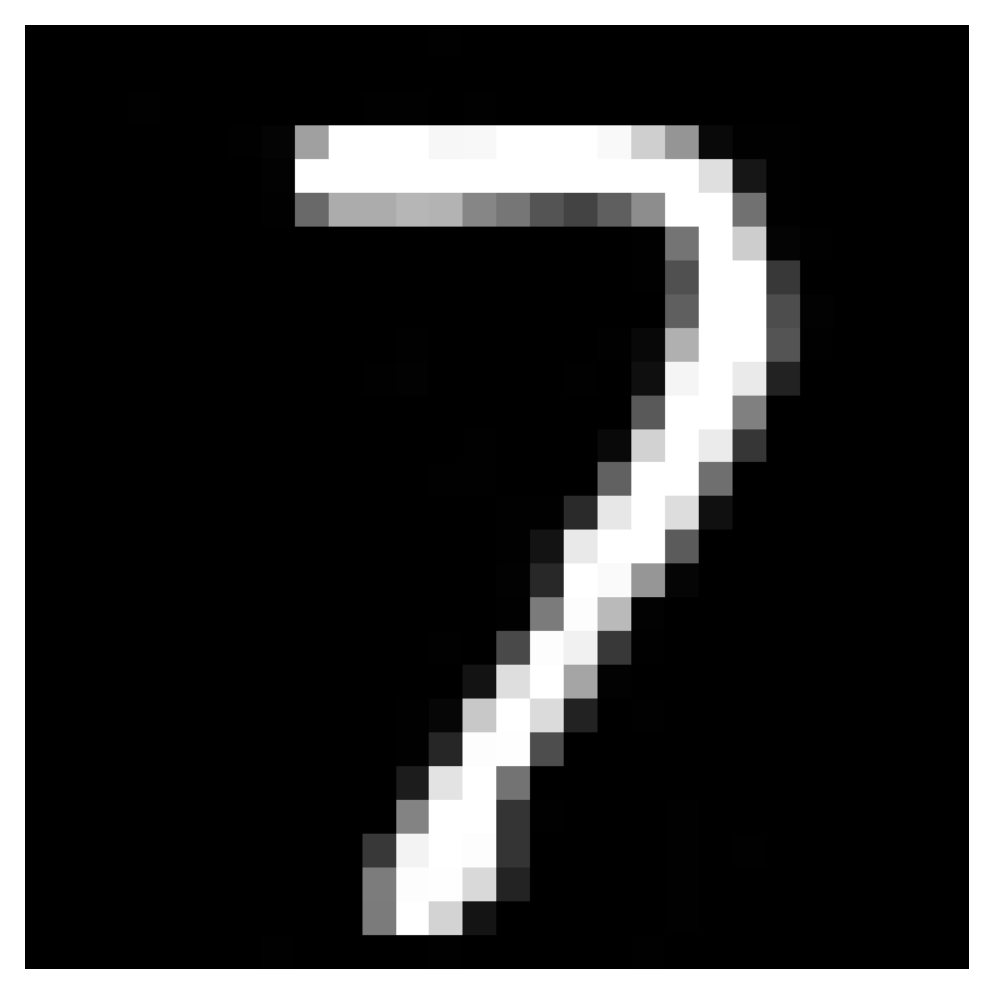} \hspace{1ex}
\includegraphics[width=\fsz]{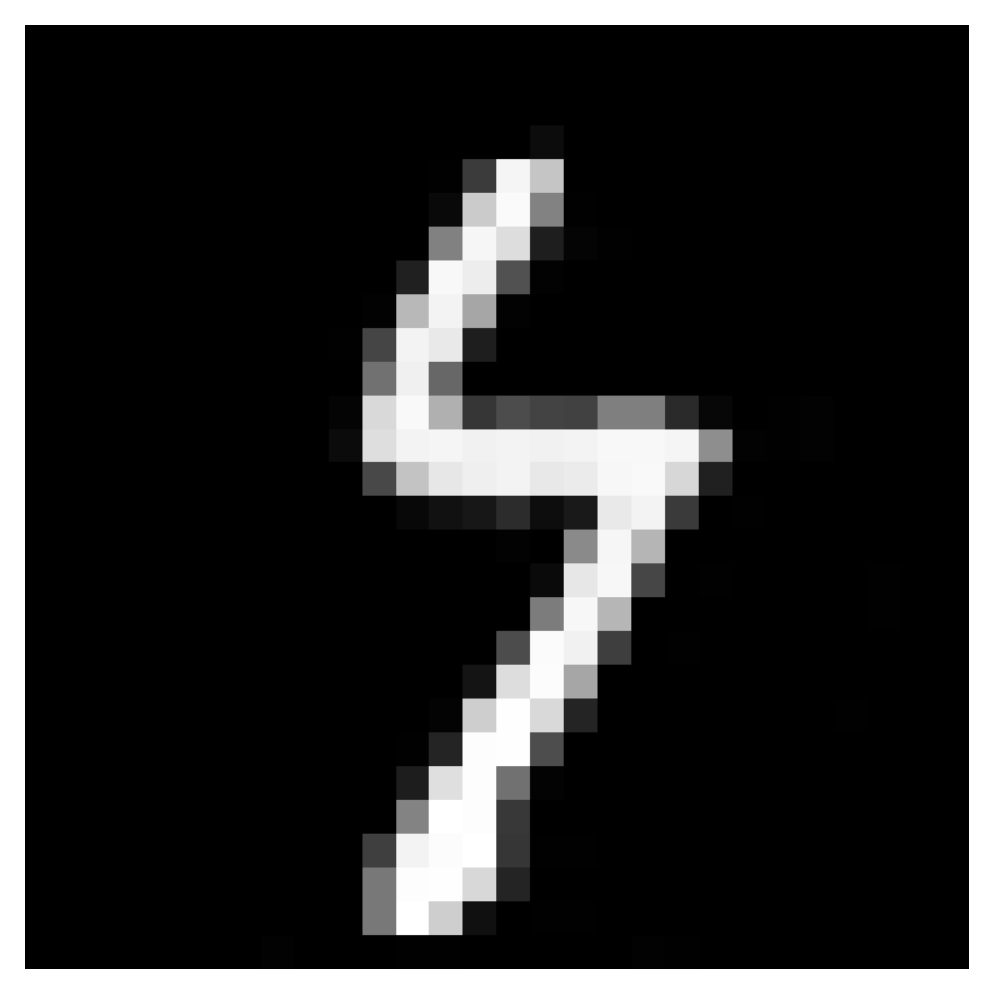} \\
\includegraphics[width=\fsz]{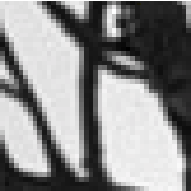} \hspace{1ex} 
\includegraphics[width=\fsz]{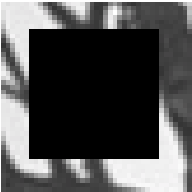} \hspace{3ex}
\includegraphics[width=\fsz]{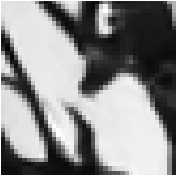} \hspace{1ex}
\includegraphics[width=\fsz]{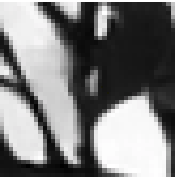} \hspace{1ex}
\includegraphics[width=\fsz]{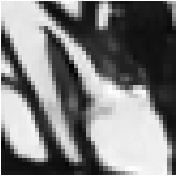} \\
\includegraphics[width=\fsz]{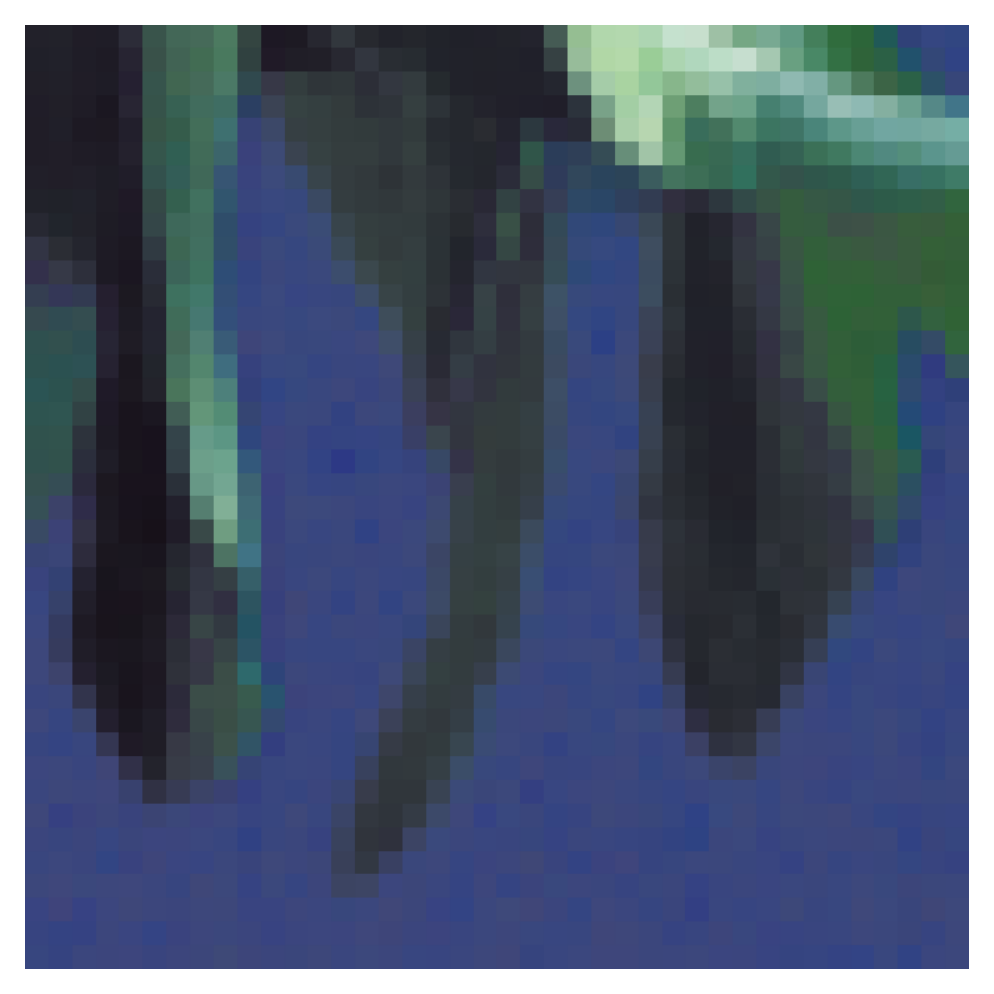} \hspace{1ex} 
\includegraphics[width=\fsz]{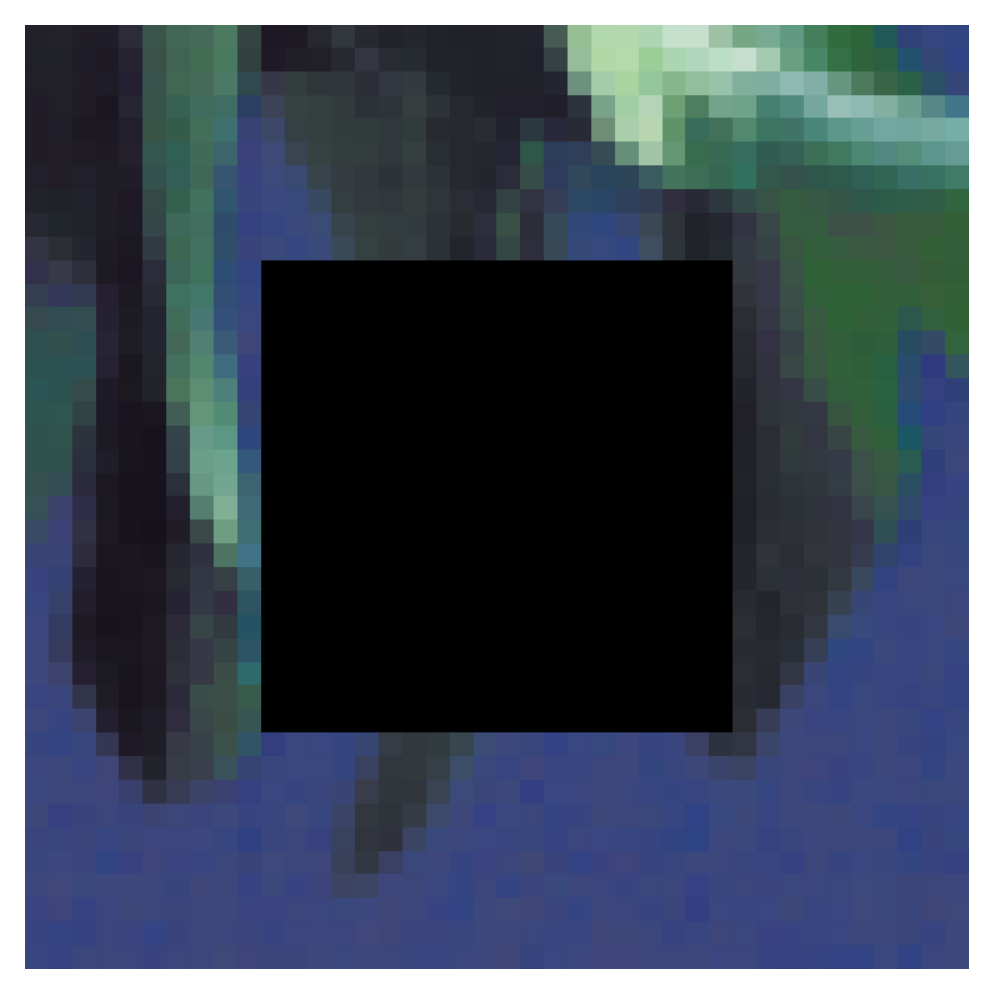} \hspace{3ex}
\includegraphics[width=\fsz]{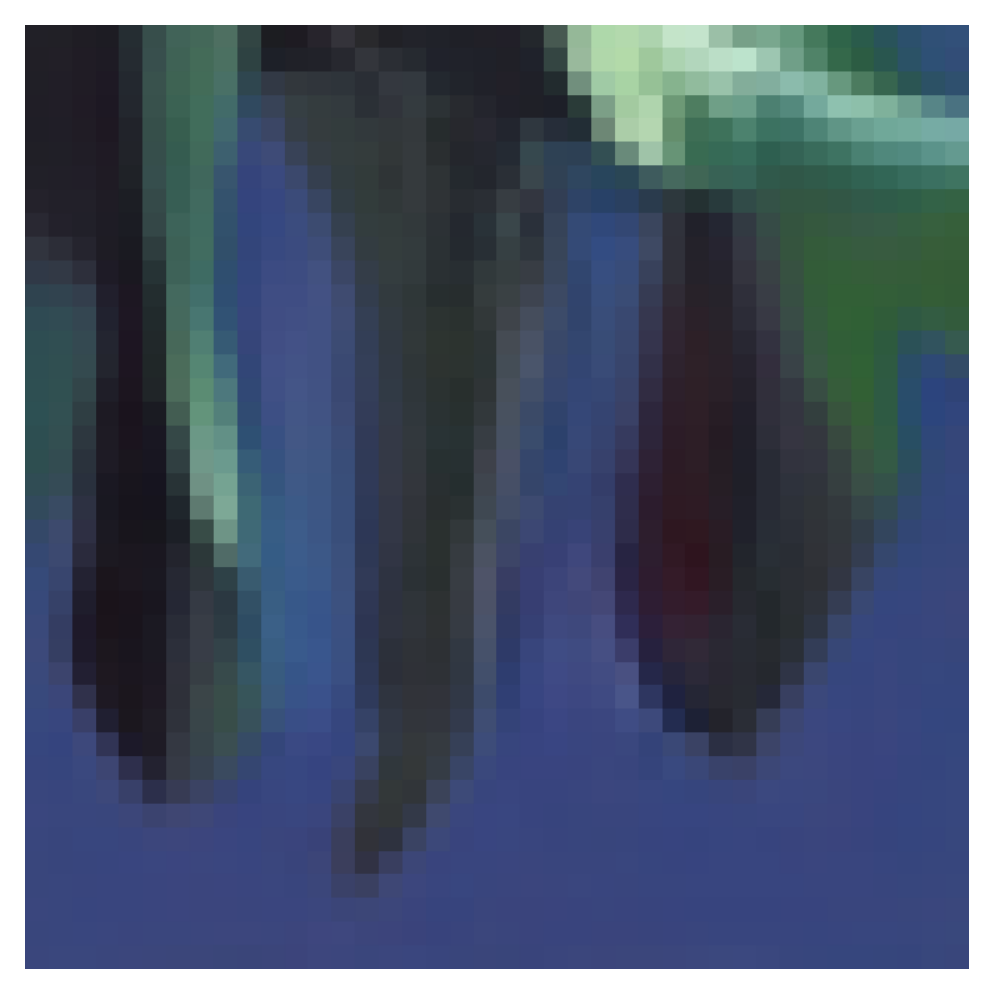} \hspace{1ex}
\includegraphics[width=\fsz]{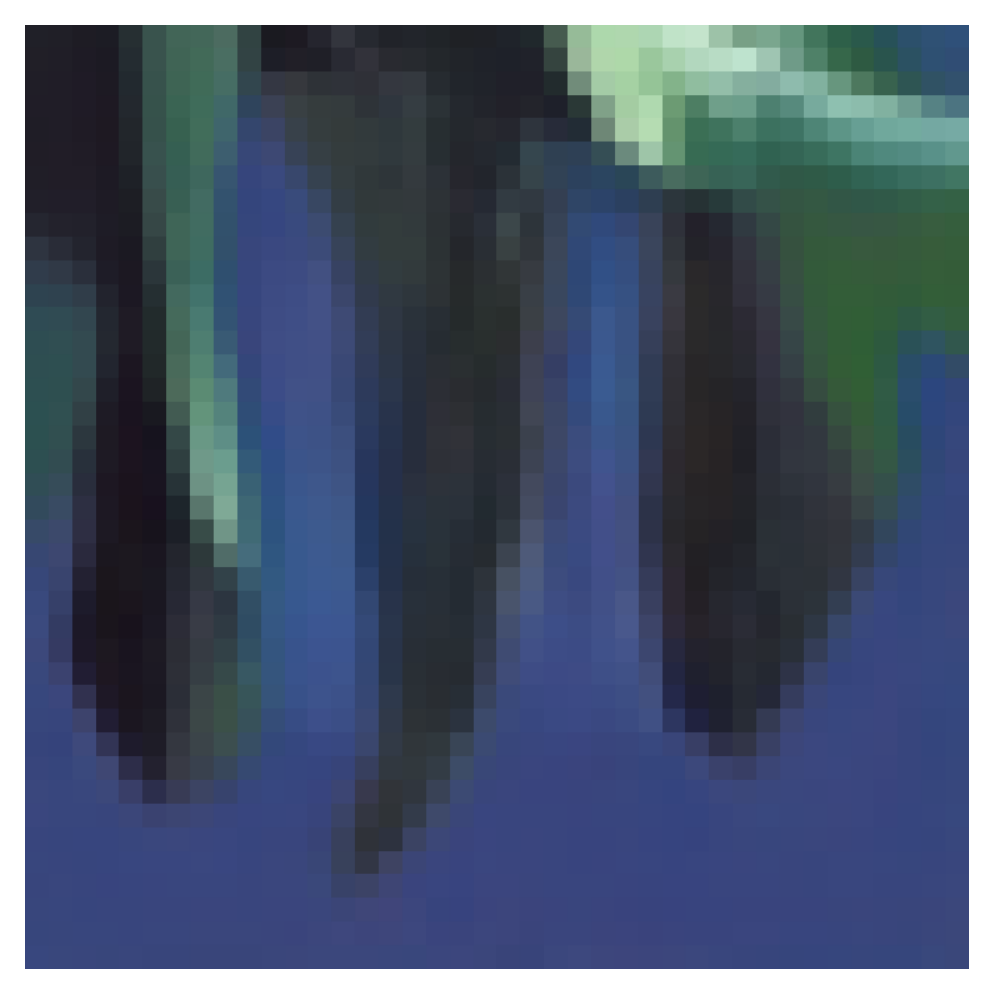} \hspace{1ex}
\includegraphics[width=\fsz]{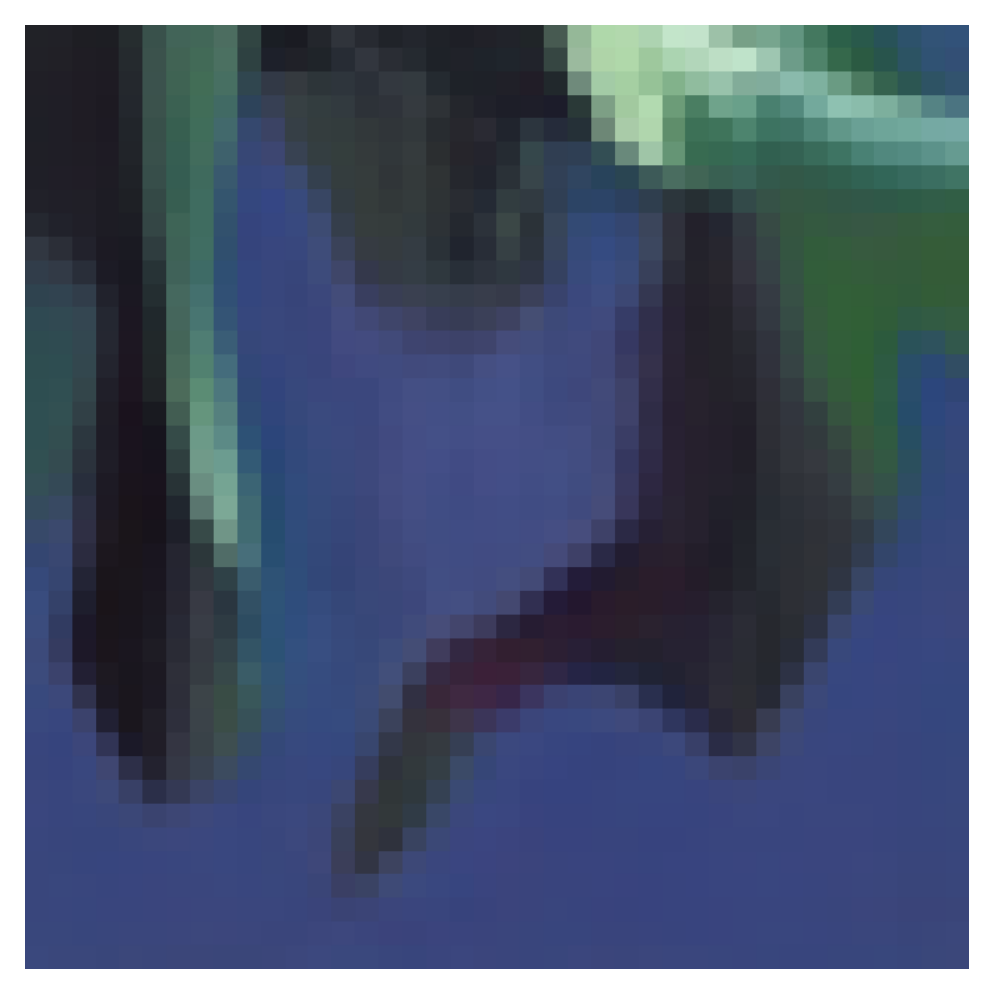} 
\caption{Inpainting examples generated using three BF-CNN denoisers trained on (1) MNIST, (2) Berkeley grayscale segmentation dataset (3) Berkeley color segmentation dataset. First column: original images.  Second column: partially measured images. Right three columns: Restored examples, with different random initializations, $y_0$. Each initialization results in a different restored image, corresponding to a different point on the intersection of the manifold and the constraint hyper plane. 
} 
\label{fig:inpainting_diversity}
\end{figure}

\begin{figure}
\centering
\def\fsz{0.16\linewidth}
\includegraphics[width=0.15\linewidth]{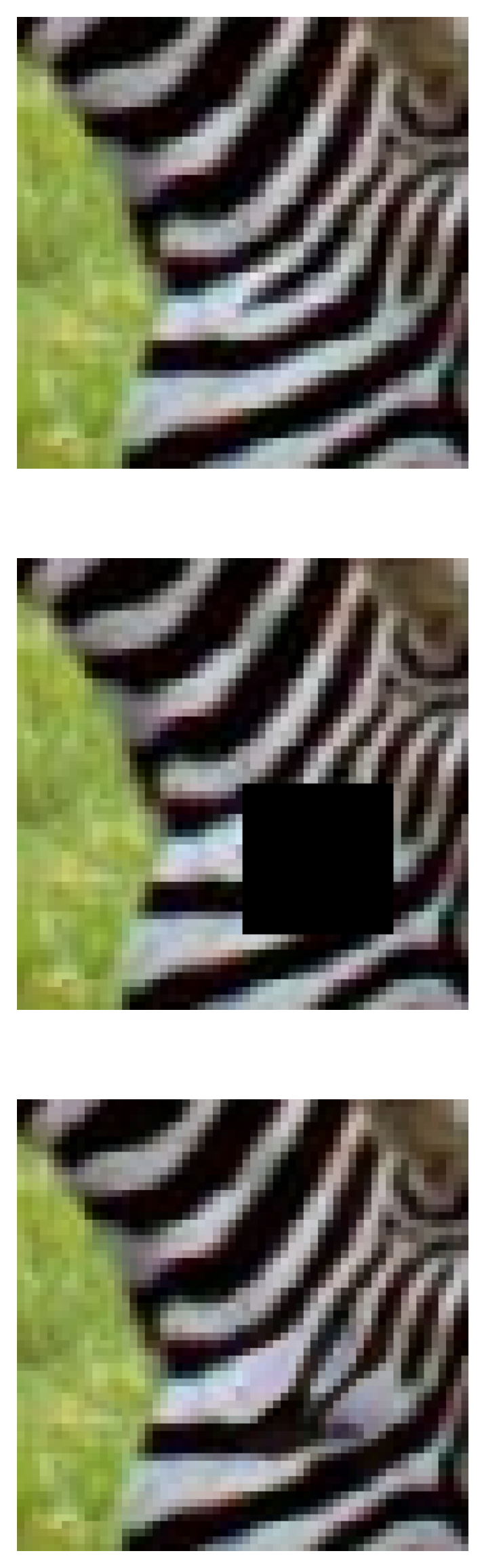}\hspace{1ex} 
\includegraphics[width=0.15\linewidth]{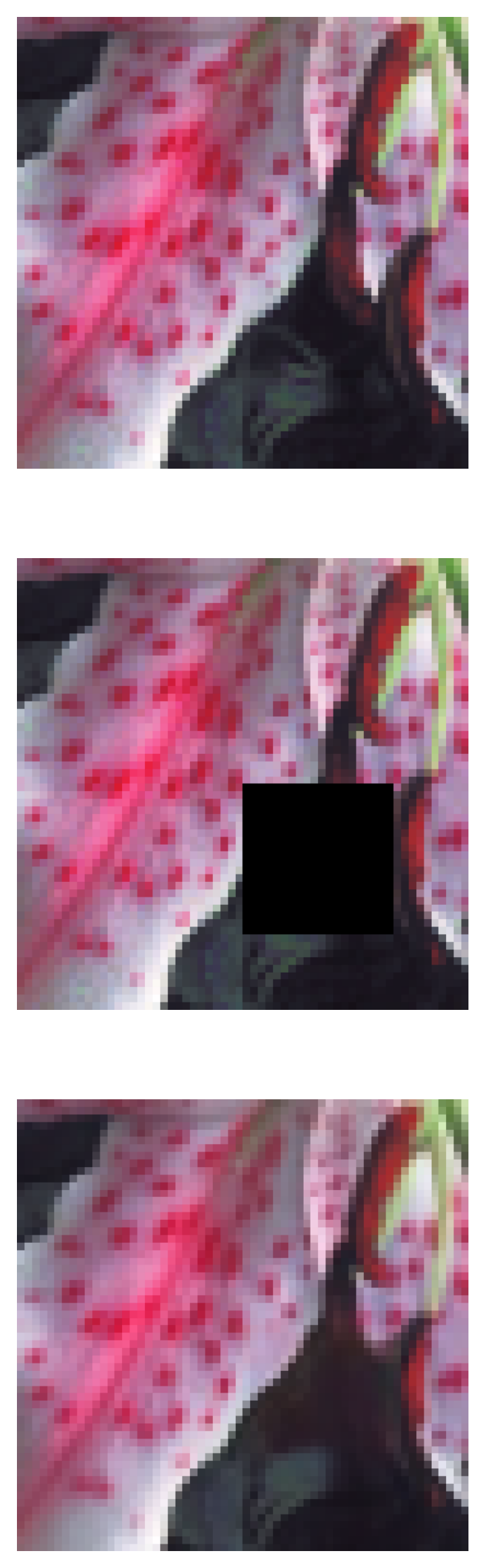}\hspace{1ex} 
\includegraphics[width=0.15\linewidth]{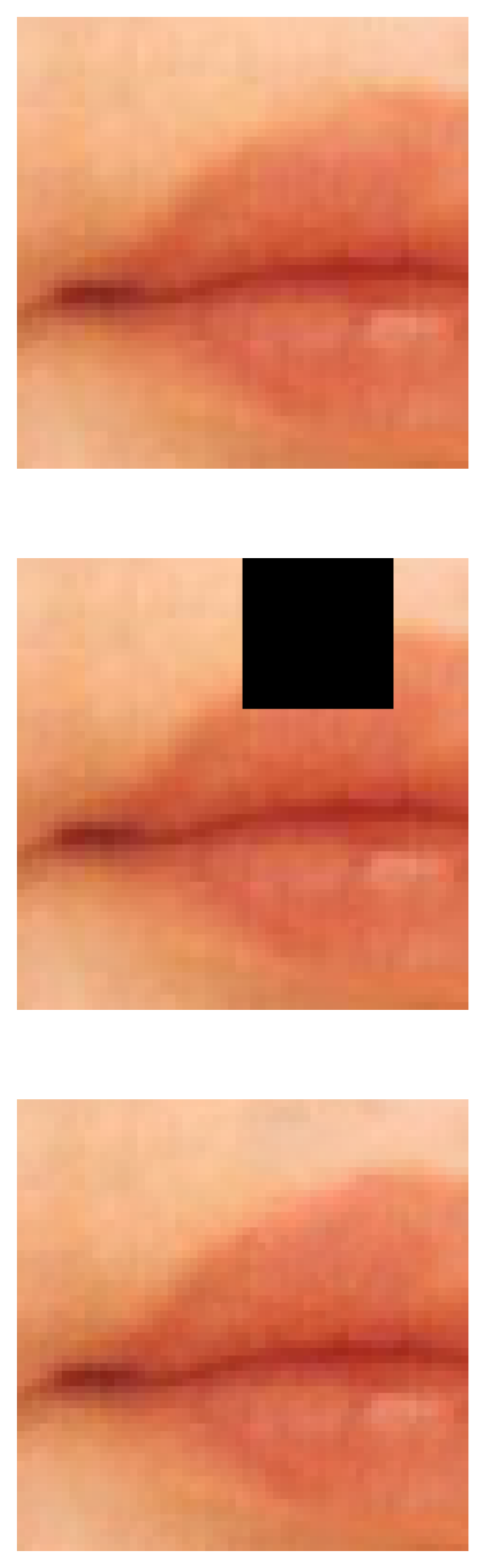}\hspace{1ex} 
\includegraphics[width=\fsz]{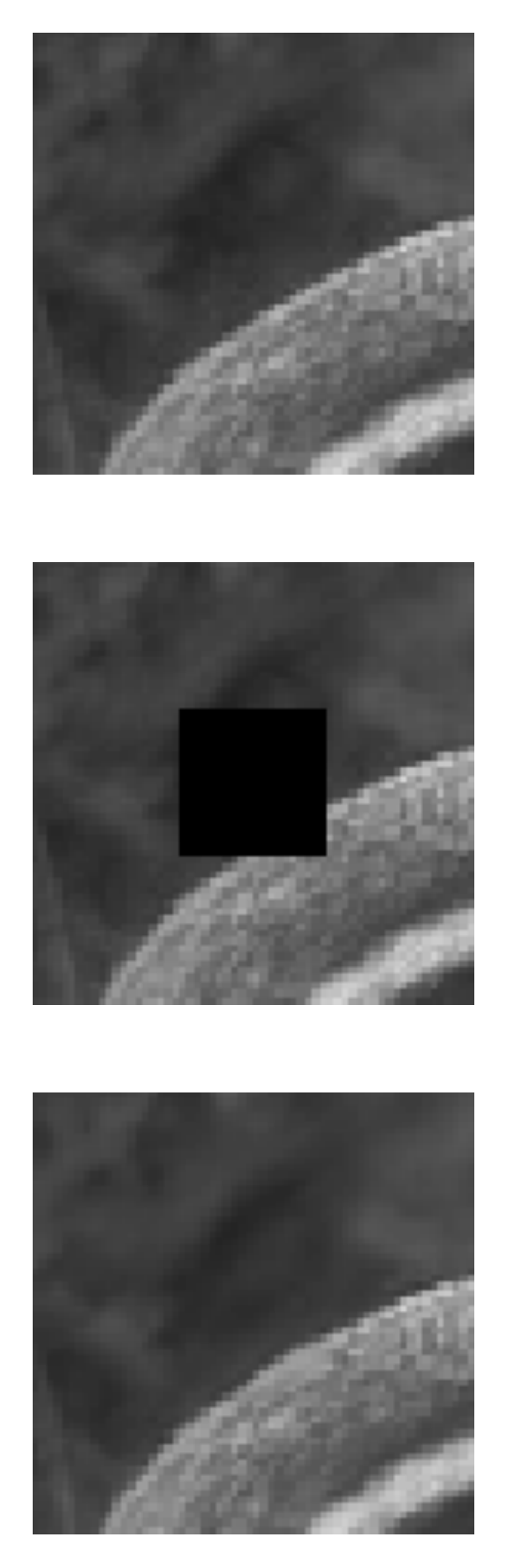}
\includegraphics[width=\fsz]{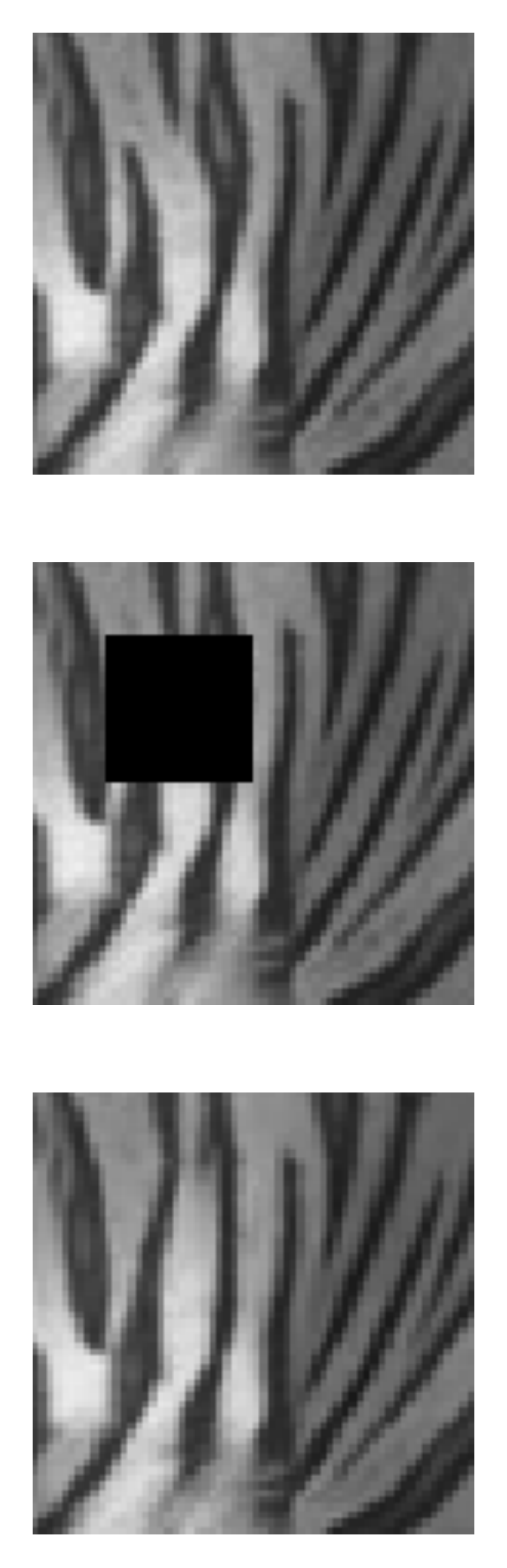}
\includegraphics[width=\fsz]{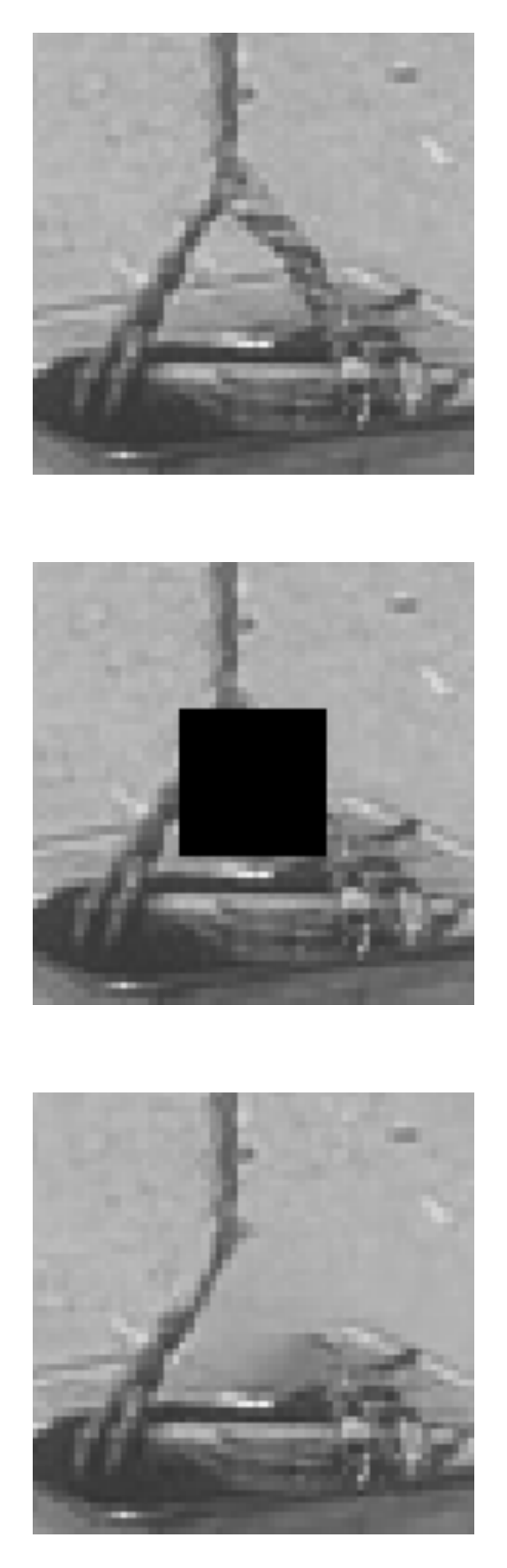}\\
\caption{Inpainting examples.  Top row: original images ($x$).  Middle: Images corrupted with blanked region ($MM^Tx$). Bottom: Images restored using our algorithm.
} 
\label{fig:inpainting2}
\end{figure}

\textbf{Inpainting.}
A simple example of a linear inverse problem involves restoring a block of missing pixels, conditioned on the surrounding content.  Here, the columns of the measurement matrix $M$ are a subset of the identity matrix, corresponding to the measured (outer) pixel locations. 
We choose a missing block of size $30 \times 30$ pixels, which is less than the size of the receptive field of the BF-CNN network ($40 \times 40$), the largest extent over which this denoiser can be expected to directly capture joint statistical relationships. There is no single correct solution for this problem: Figure~\ref{fig:inpainting_diversity} shows multiple samples, resulting from different initalizations.  Each appears plausible and consistent with the surrounding content. Figure \ref{fig:inpainting2} shows additional examples. 

\begin{figure}
\centering
\def\fsz{0.155\linewidth}
\includegraphics[width=\fsz]{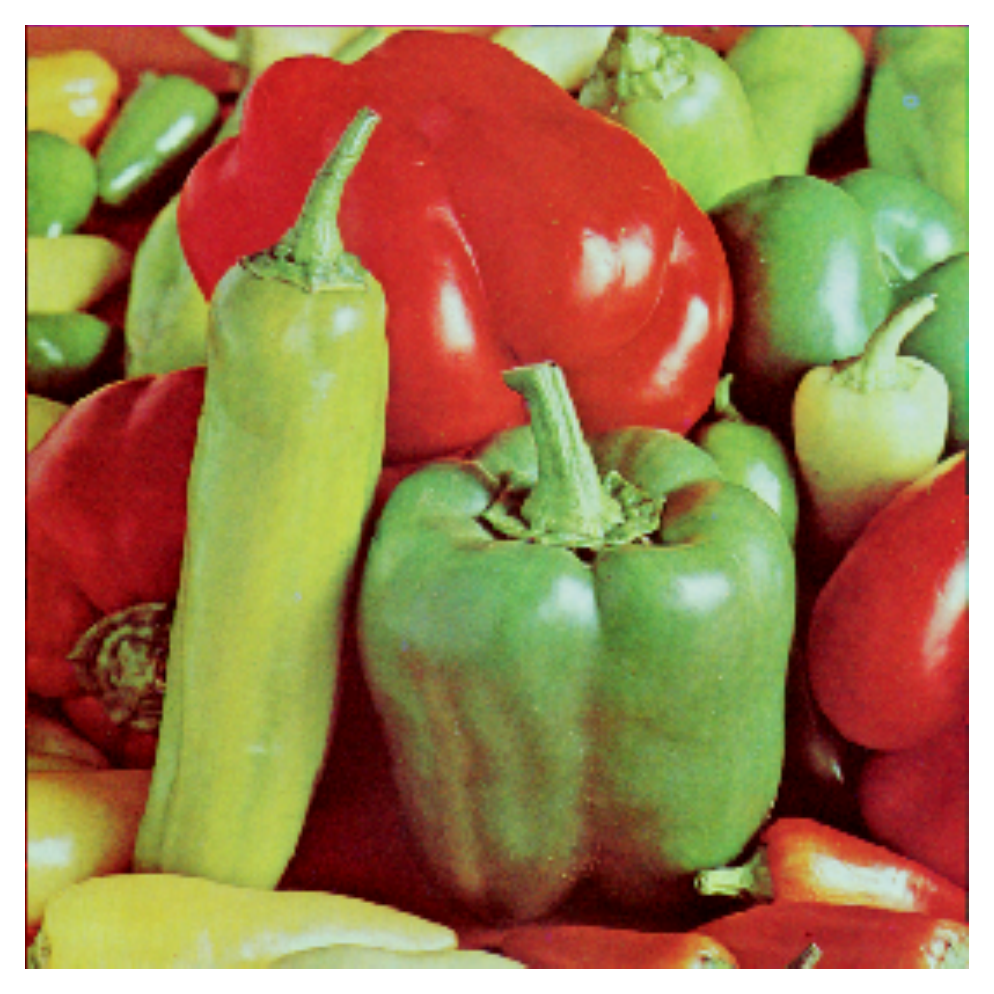} 
\includegraphics[width=\fsz]{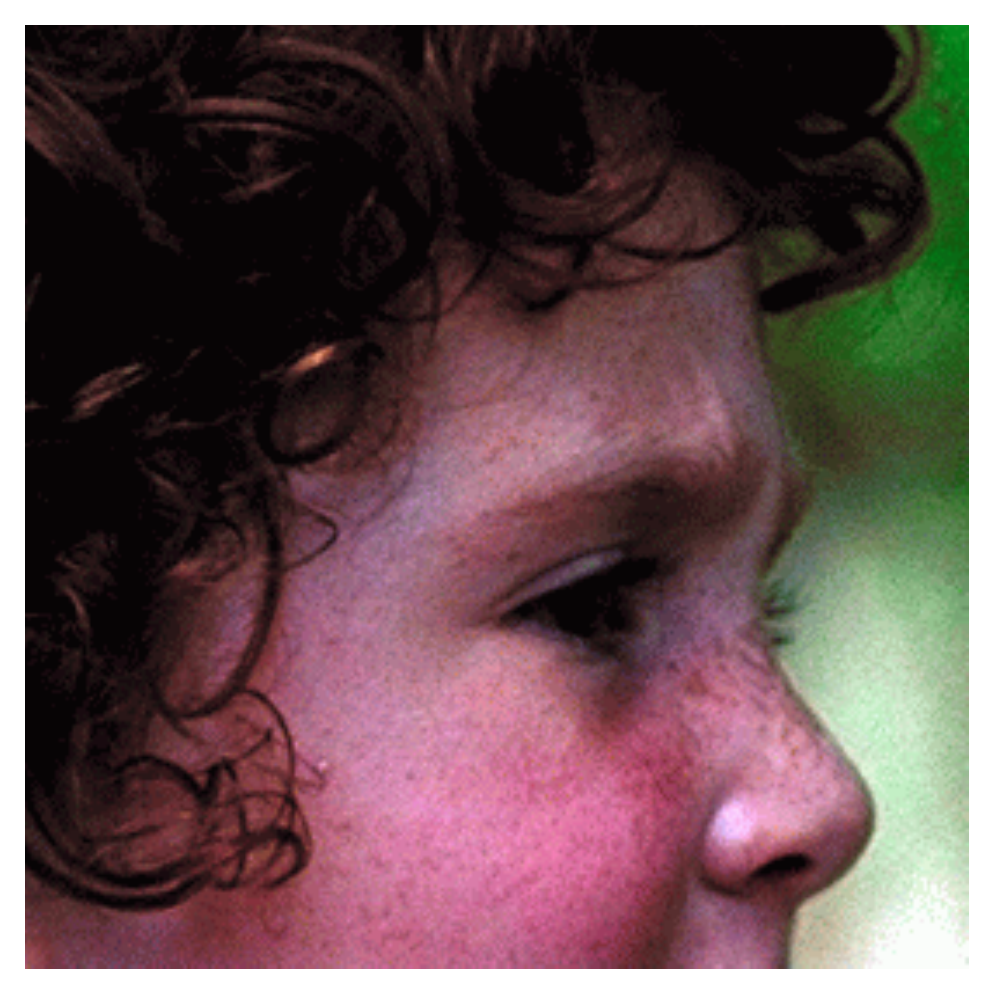}
\includegraphics[width=\fsz]{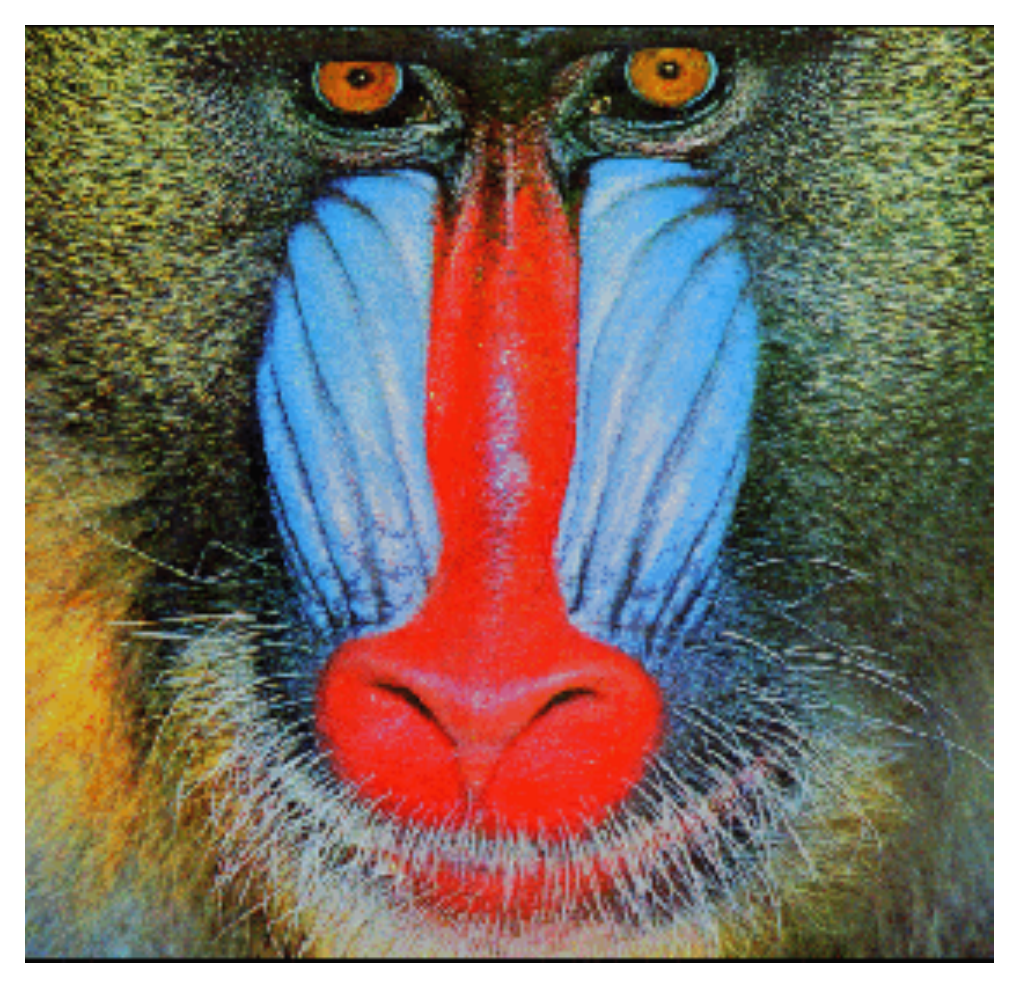} \hfil
\includegraphics[width=\fsz]{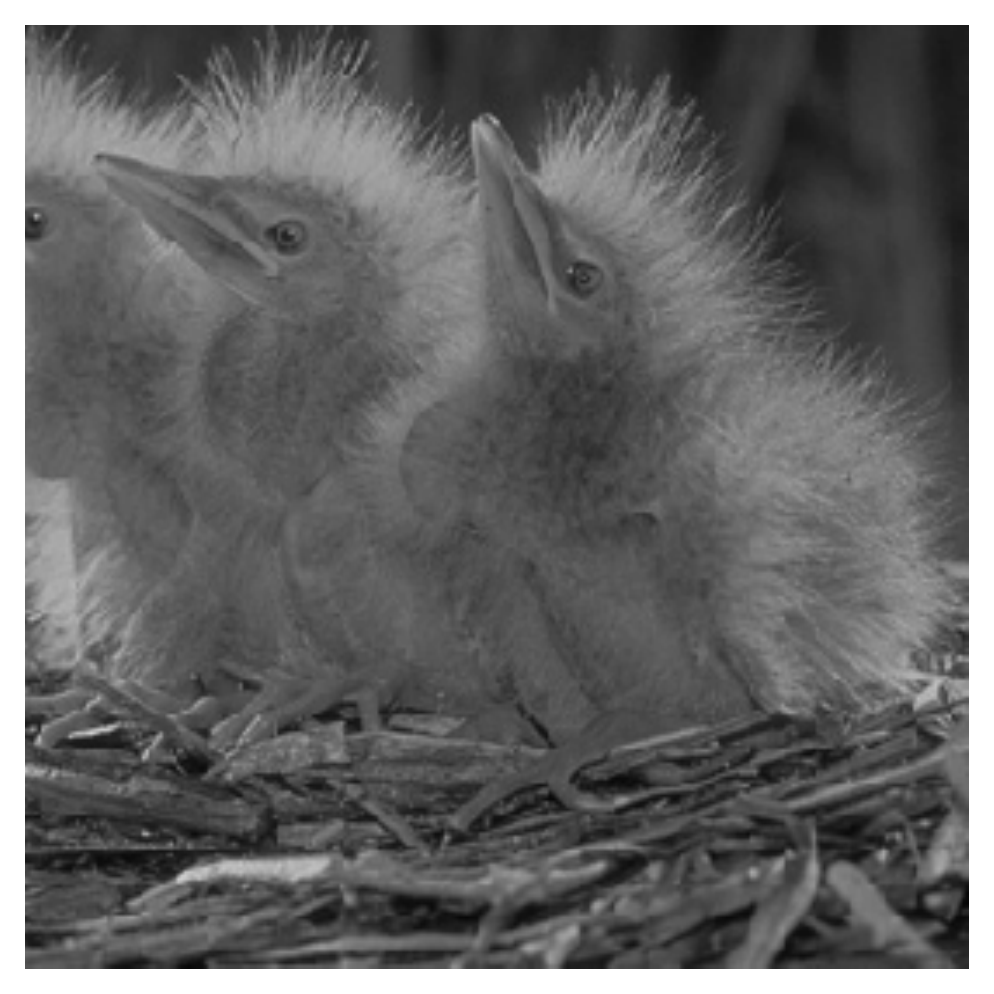} 
\includegraphics[width=\fsz]{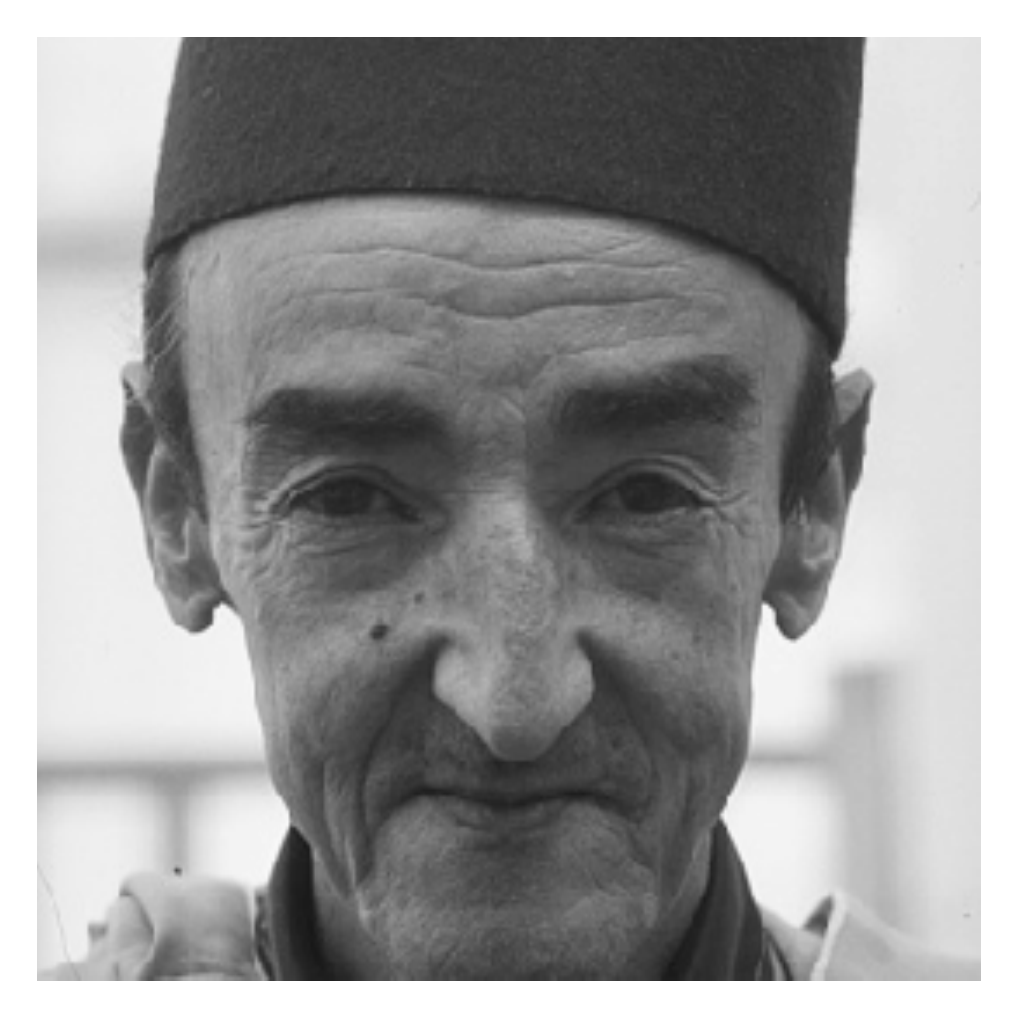}
\includegraphics[width=\fsz]{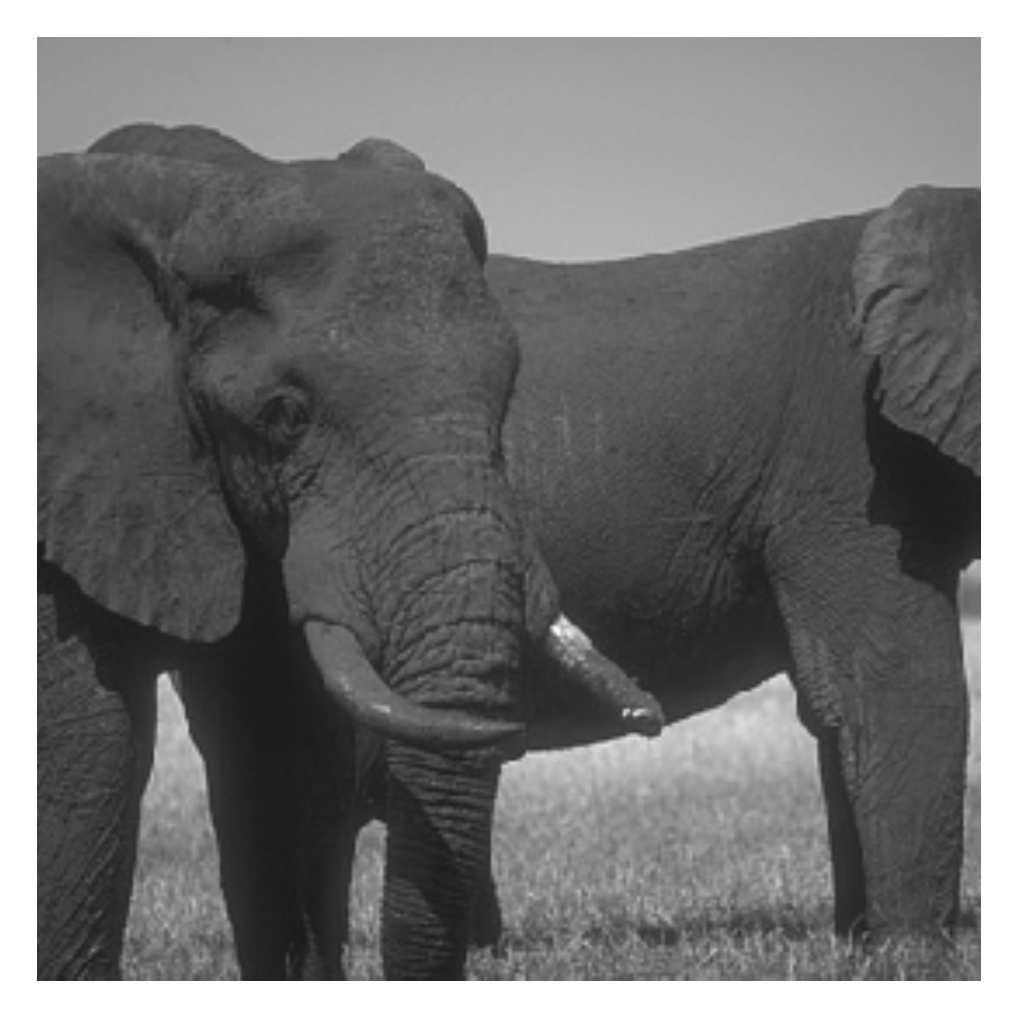} \\
\includegraphics[width=\fsz]{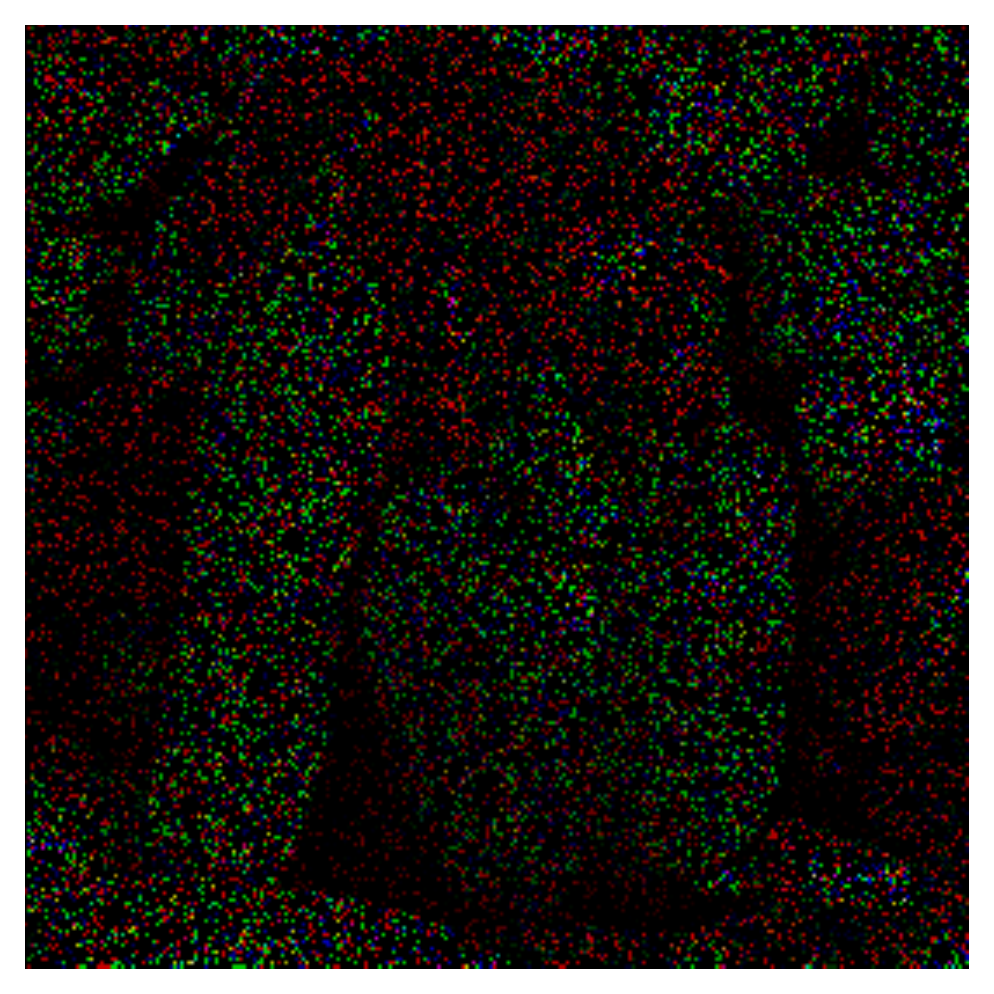} 
\includegraphics[width=\fsz]{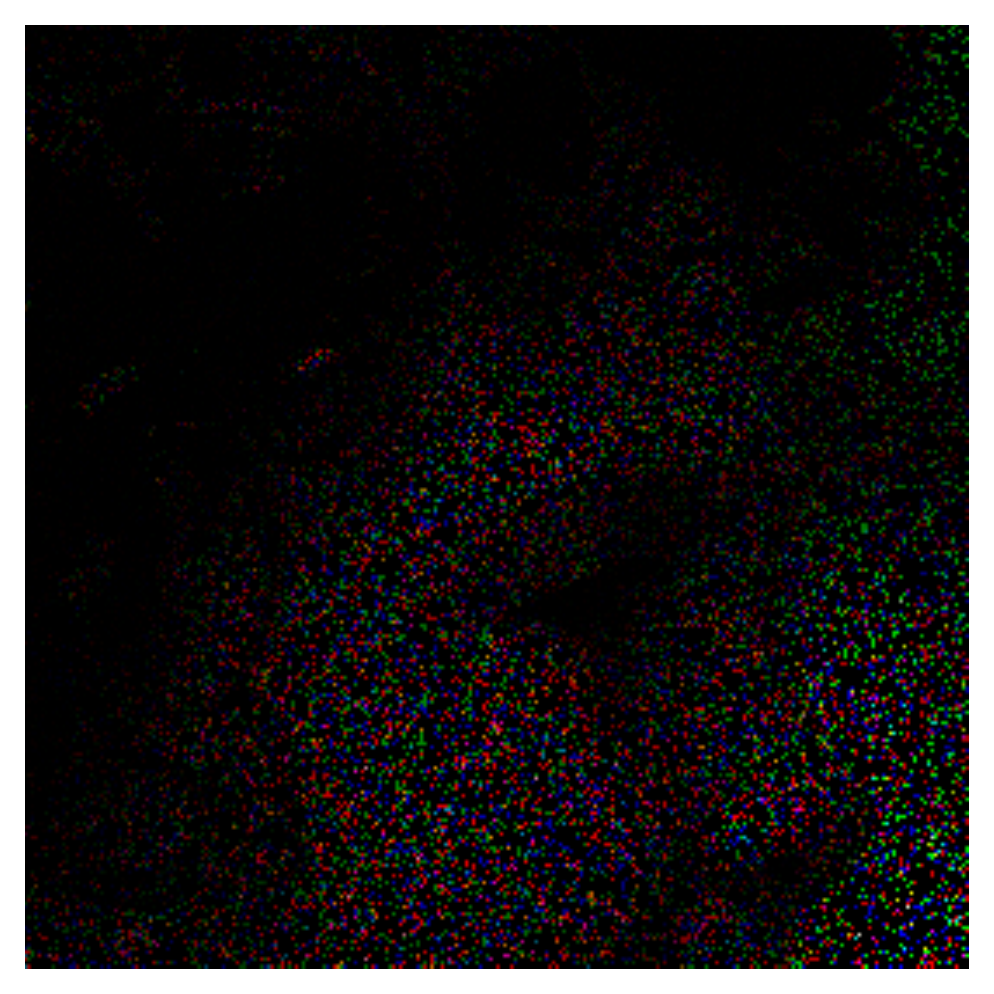}
\includegraphics[width=\fsz]{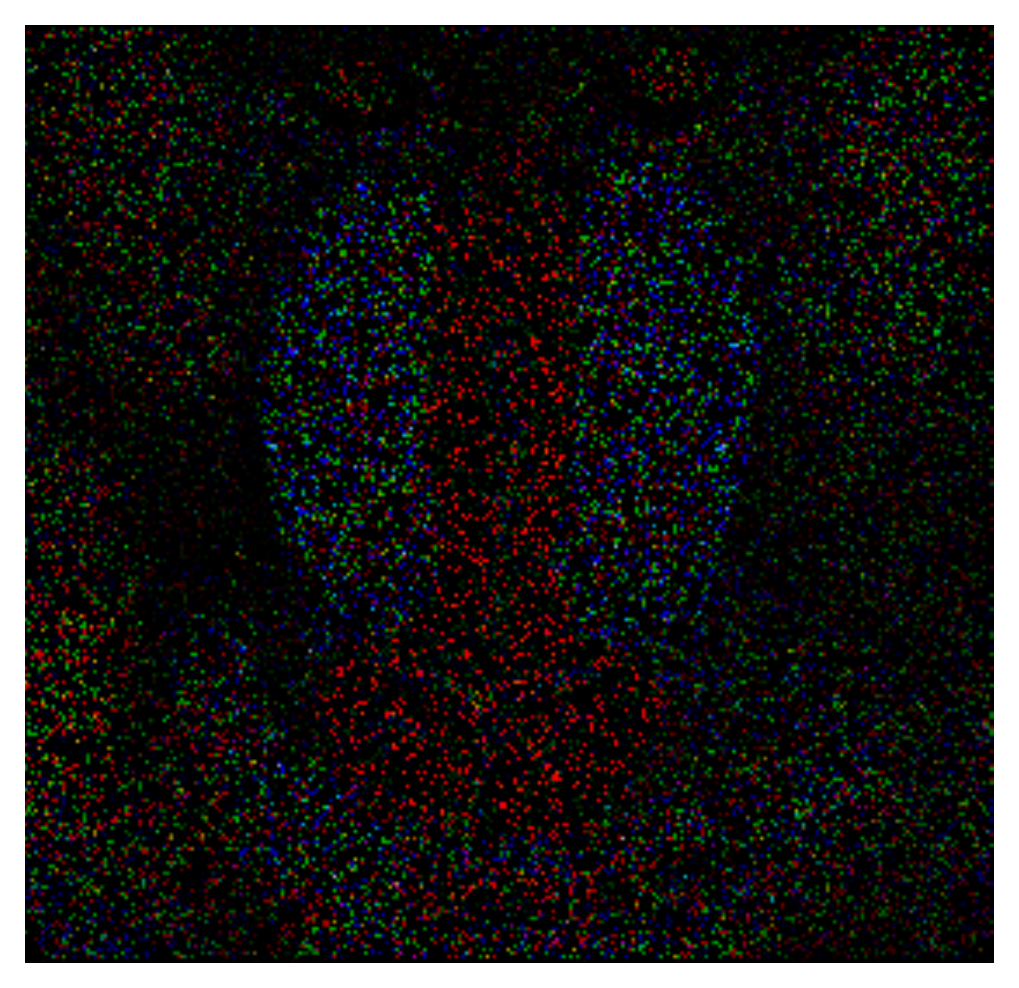} \hfil
\includegraphics[width=\fsz]{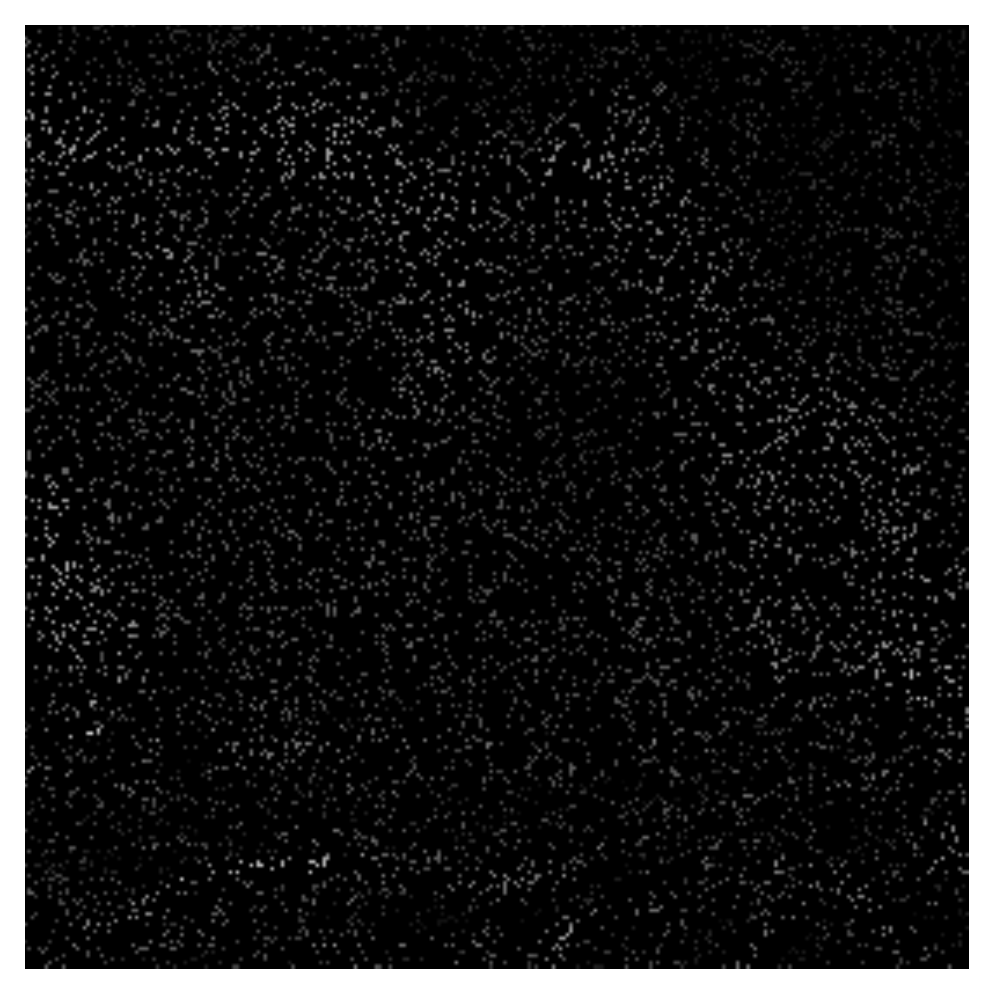} 
\includegraphics[width=\fsz]{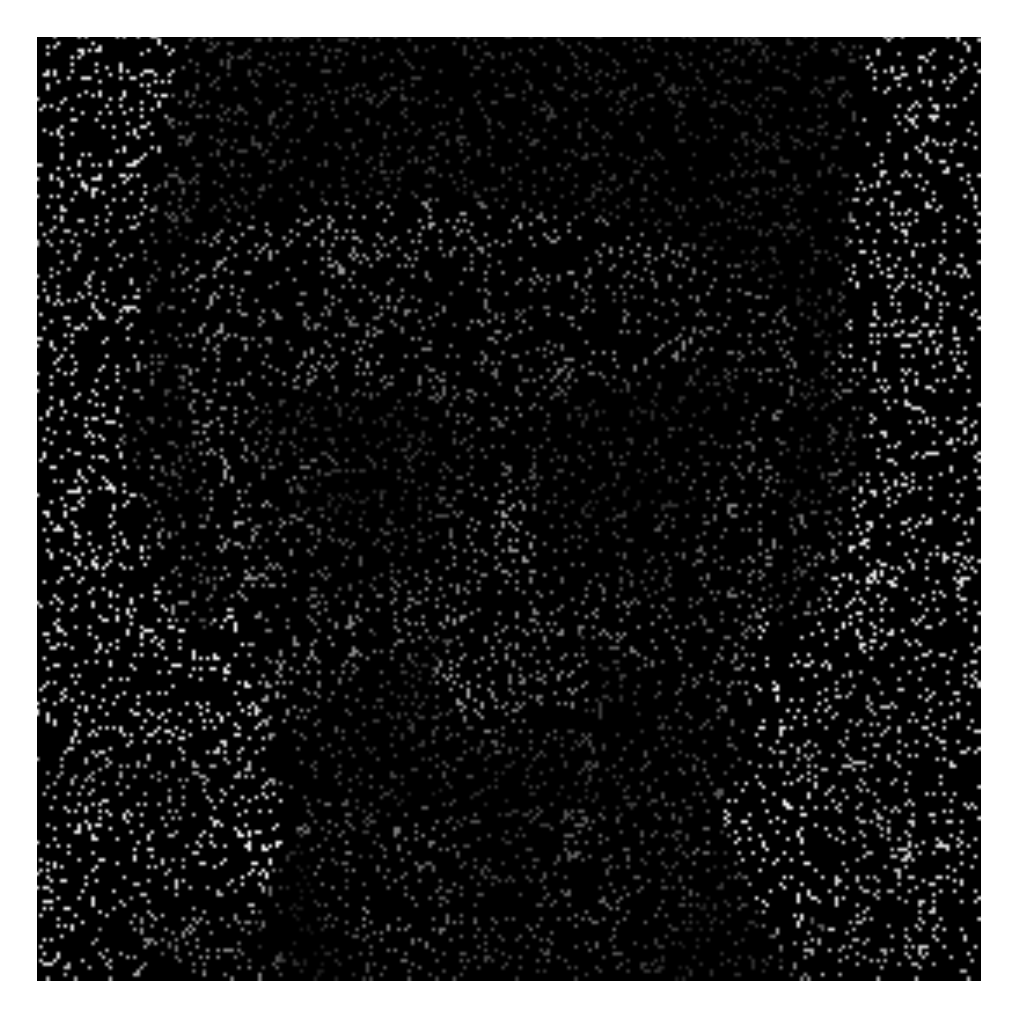}
\includegraphics[width=\fsz]{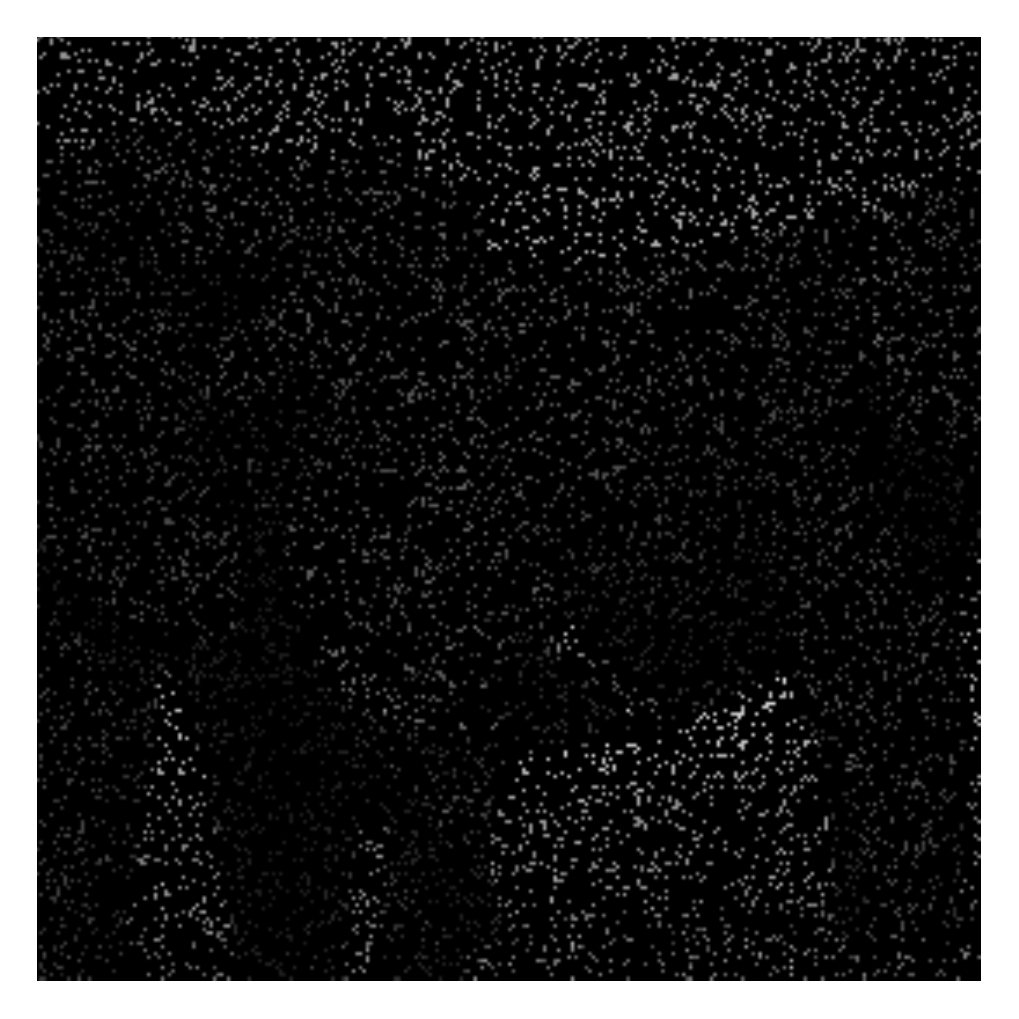} \\
\includegraphics[width=\fsz]{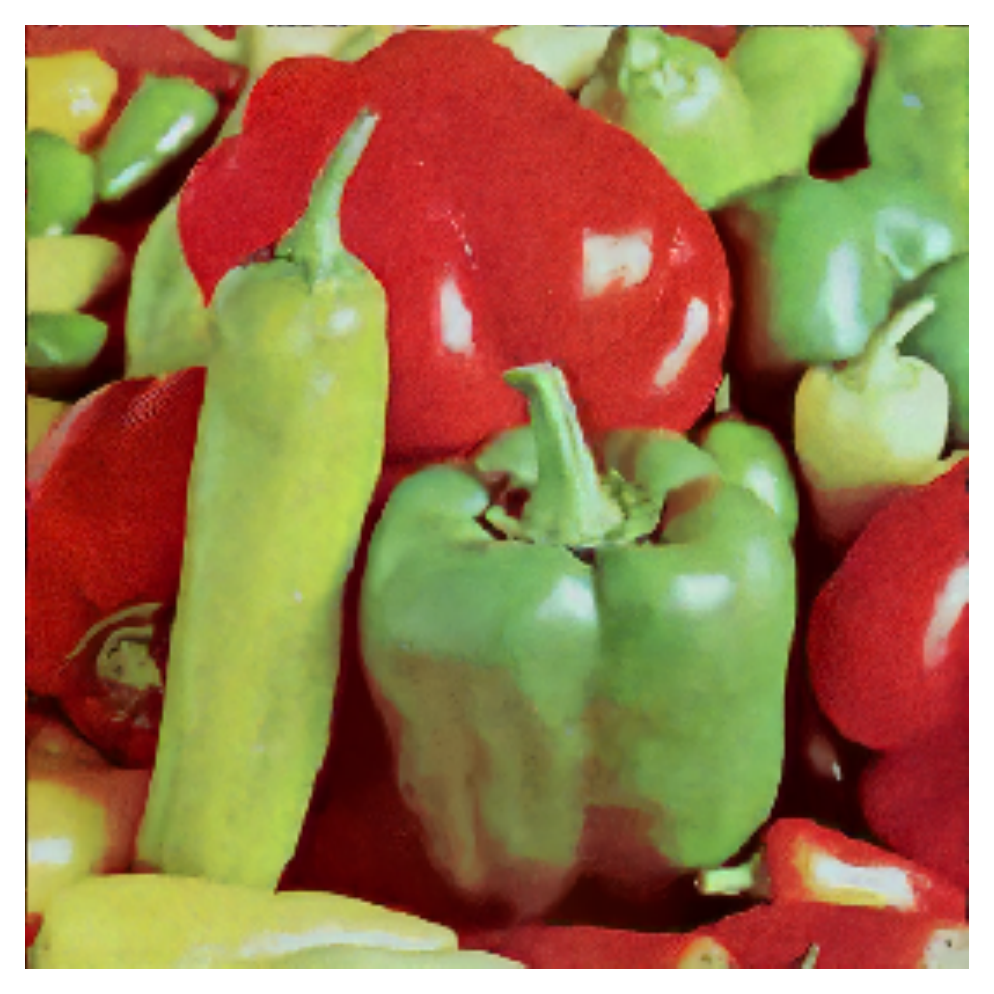} 
\includegraphics[width=\fsz]{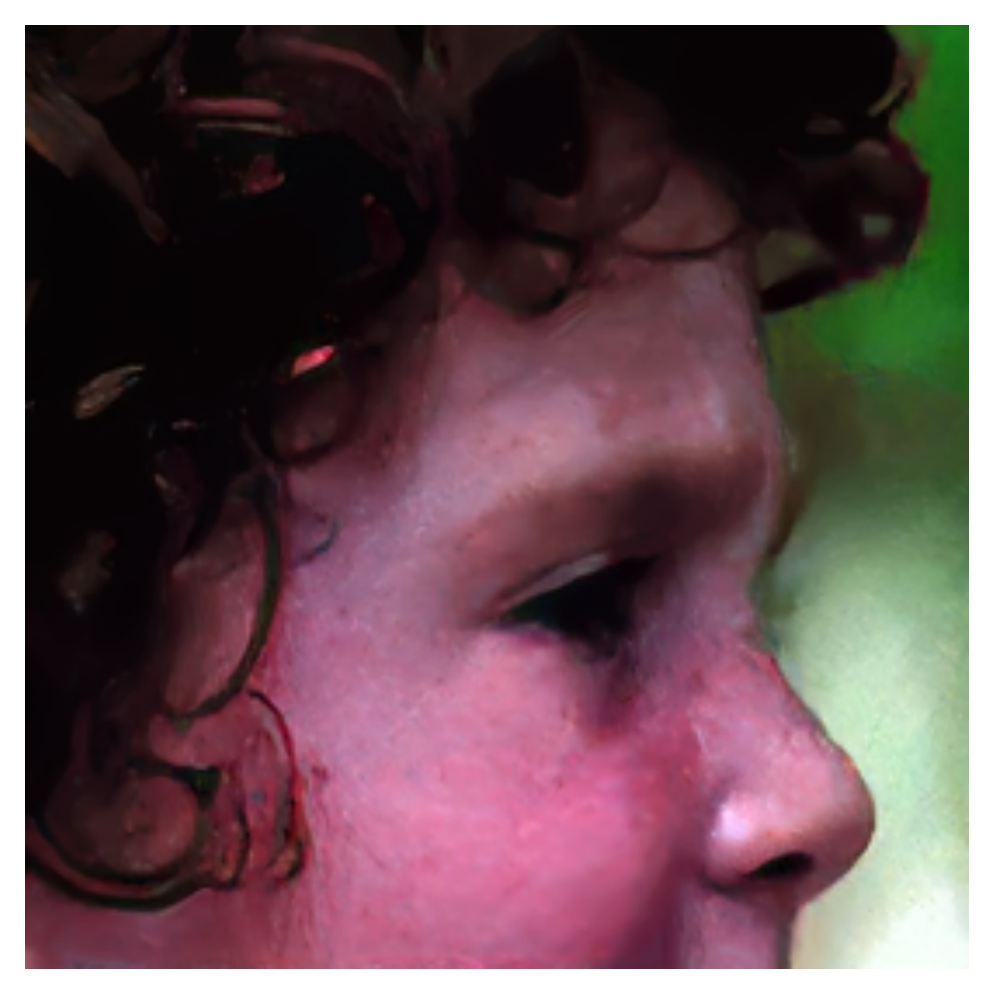}
\includegraphics[width=\fsz]{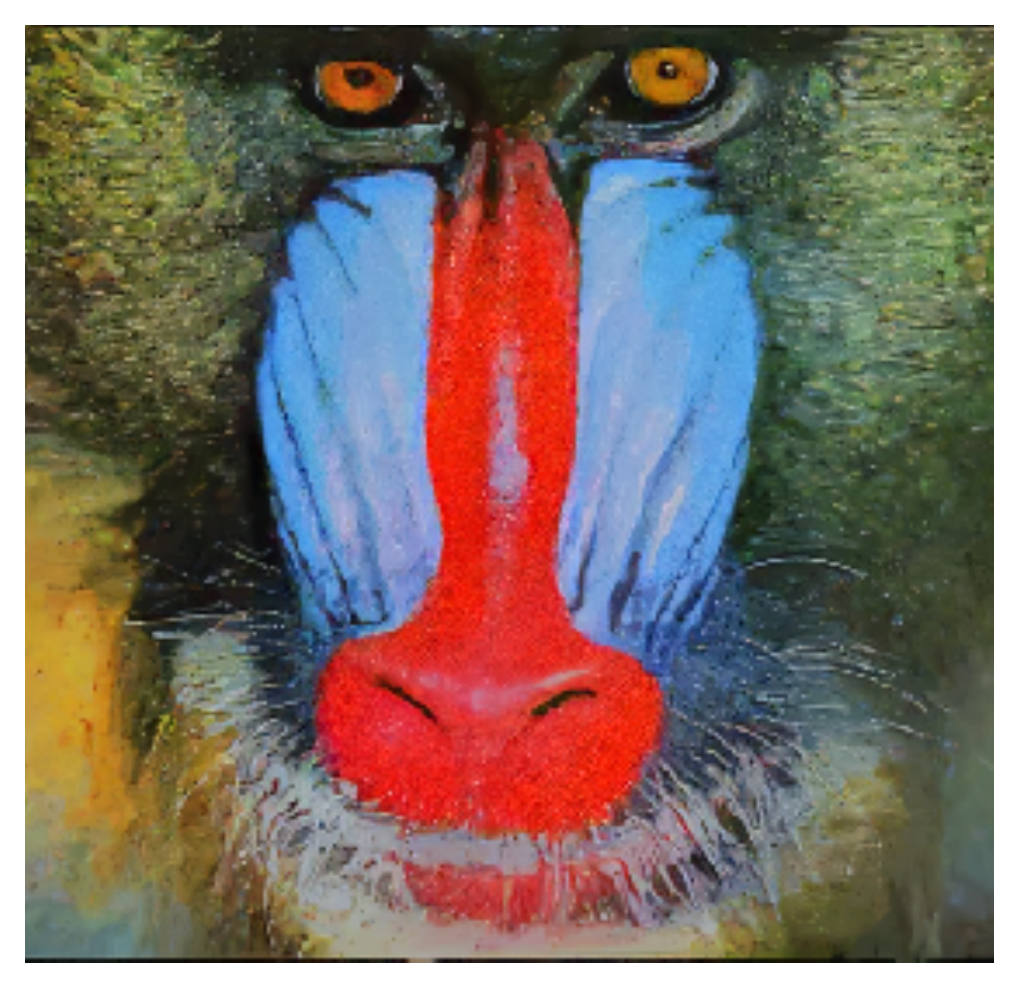} \hfil
\includegraphics[width=\fsz]{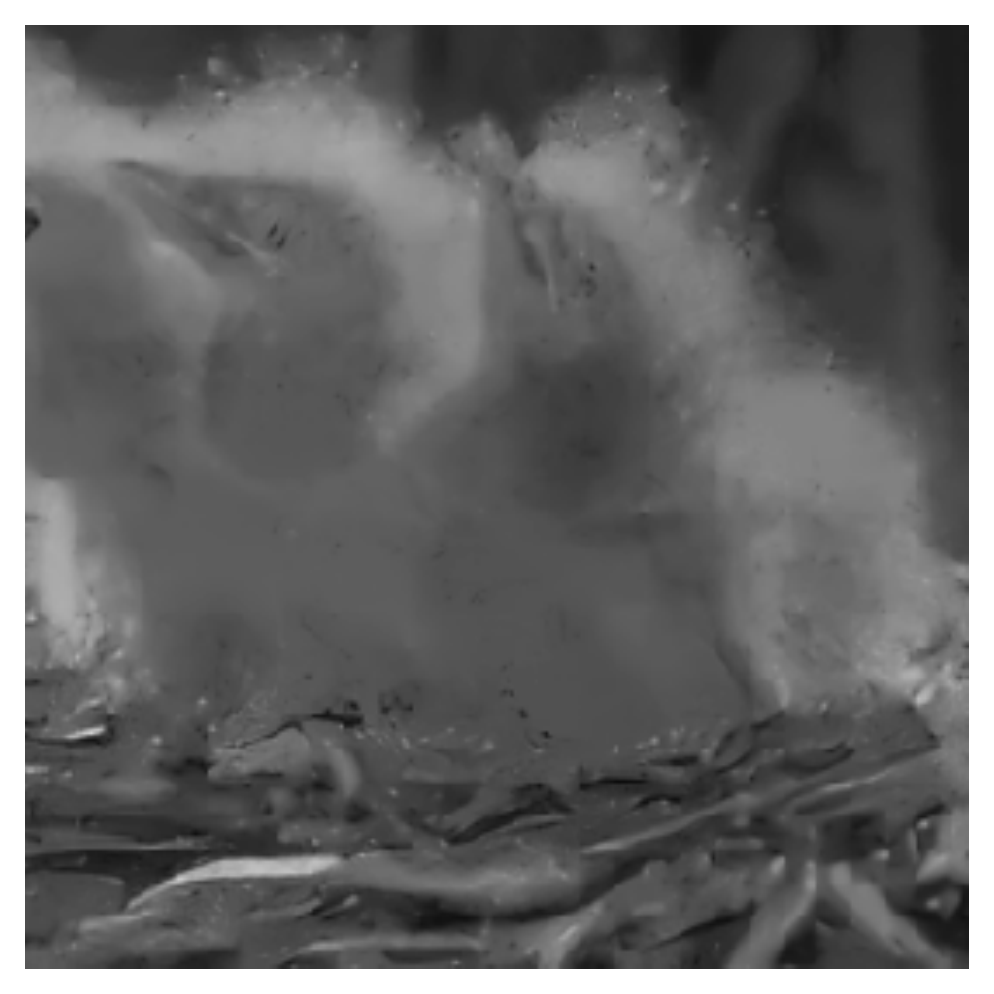} 
\includegraphics[width=\fsz]{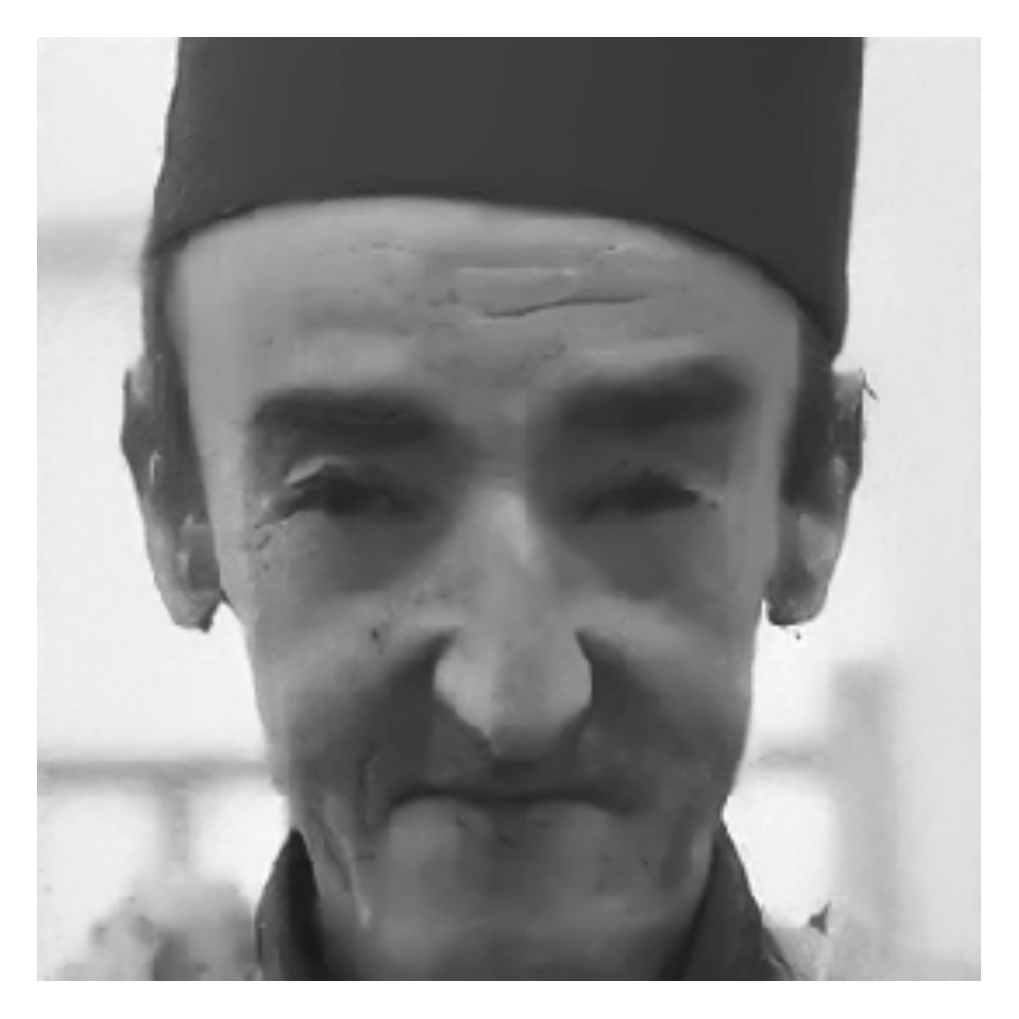} 
\includegraphics[width=\fsz]{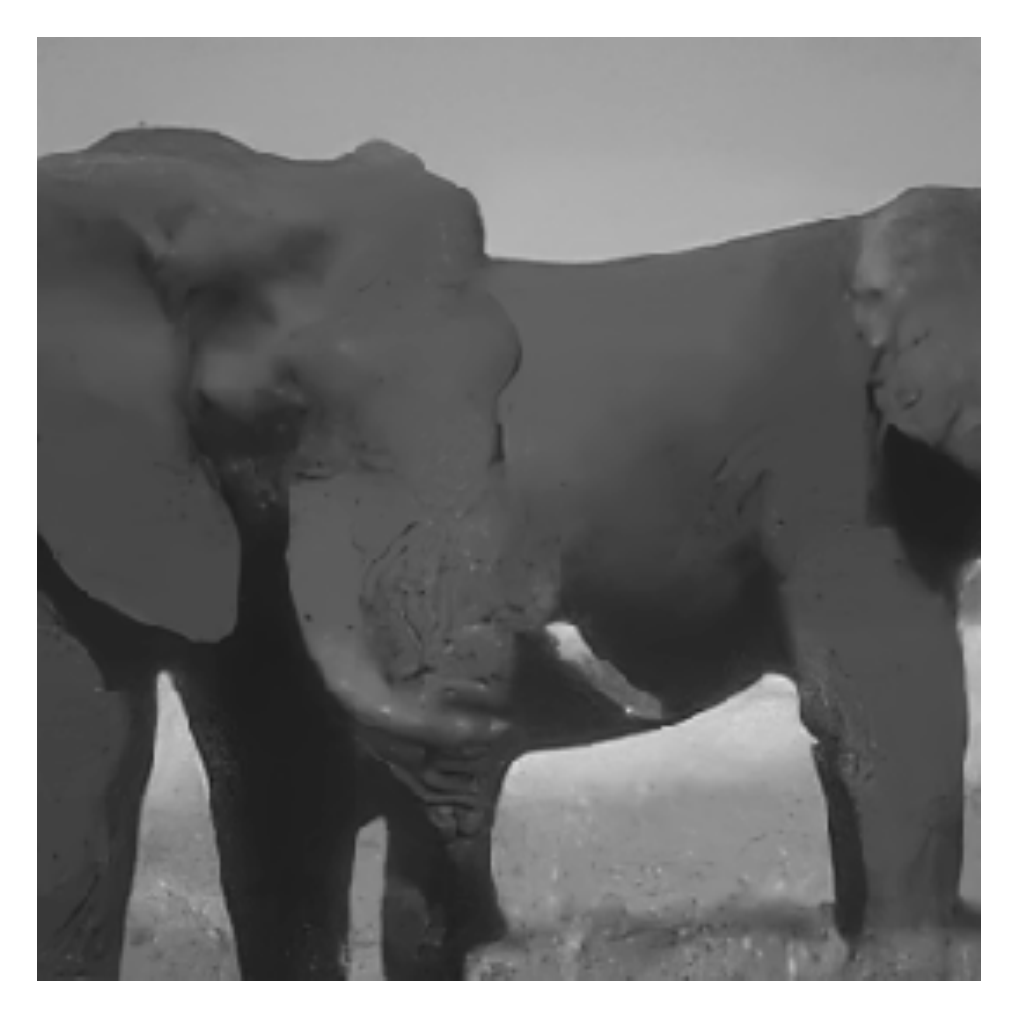} \\[-0.5ex]
\caption{\comment{Top row: original images, $x$. Middle: projection onto measurements, $MM^Tx$, Bottom: images recovered using our algorithm.} 
{Recovery of randomly selected missing pixels.} $10\%$ of dimensions retained.} 
\label{fig:missing_pixels}
\end{figure}  

\textbf{Random missing pixels.}
Consider a measurement process that discards a random subset of pixels. $M$ is a low rank matrix whose columns consist of a subset of the identity matrix corresponding to the randomly chosen set of preserved pixels. Figure \ref{fig:missing_pixels} shows examples with $10\%$ of pixels retained. Despite the significant number of missing pixels, the recovered images are remarkably similar to the originals.

\begin{figure}
\centering
\def\fsz{0.15\linewidth} \def\nsp{\hspace*{-.02\linewidth}}
\begin{tabular}{cccccc}
  \includegraphics[width=\fsz]{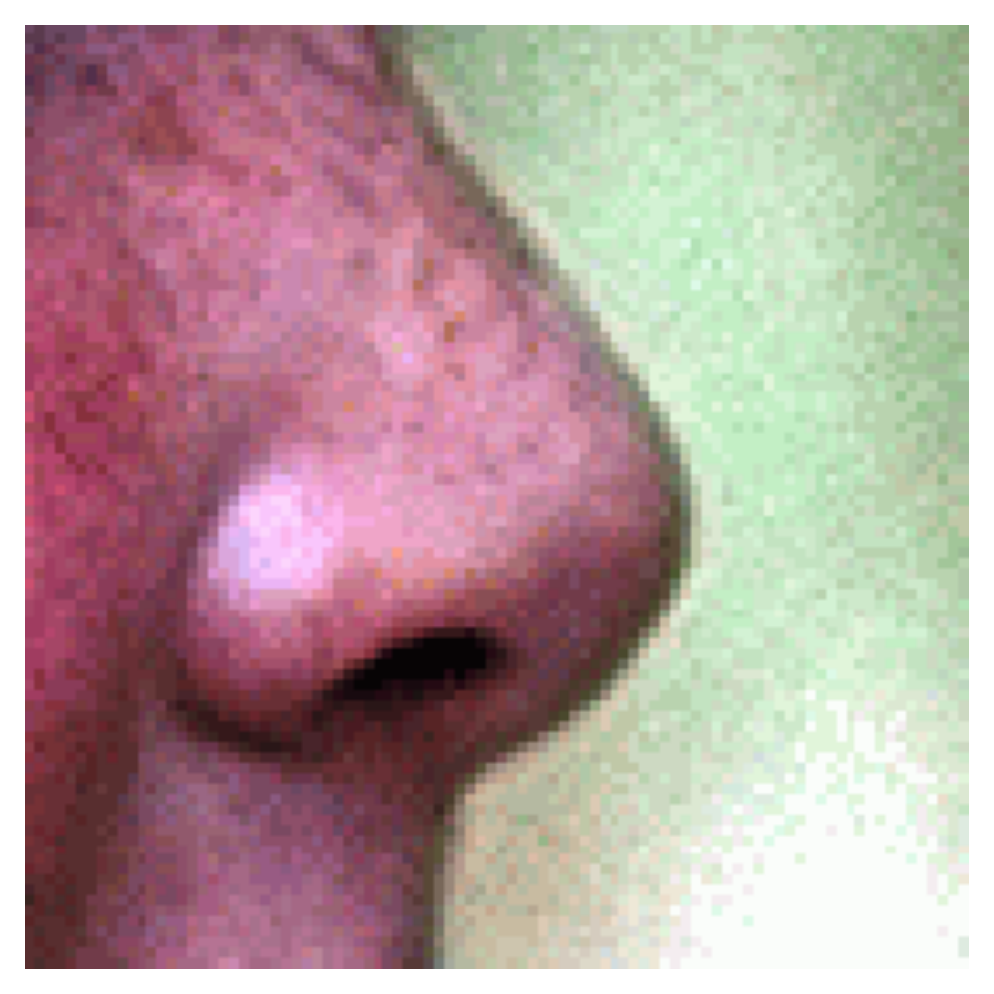}\nsp &
  \includegraphics[width=\fsz]{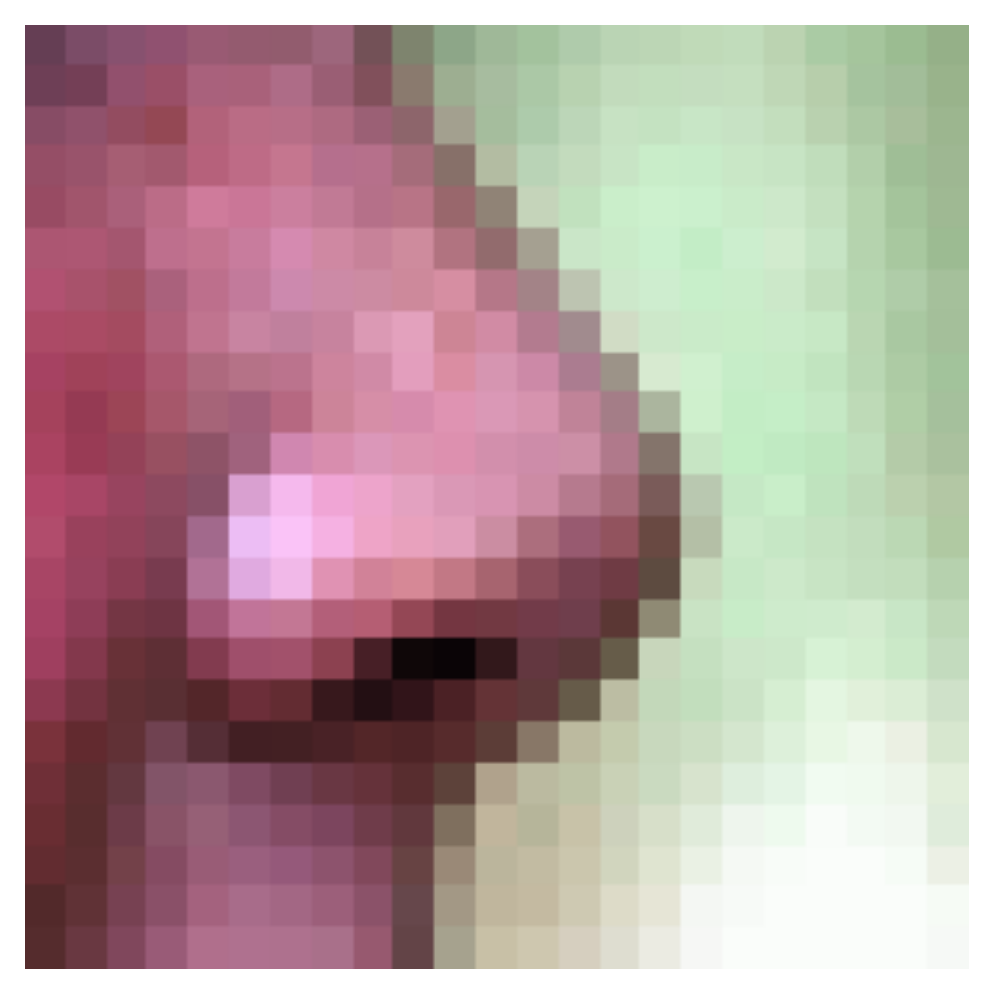}\nsp &
  \includegraphics[width=\fsz]{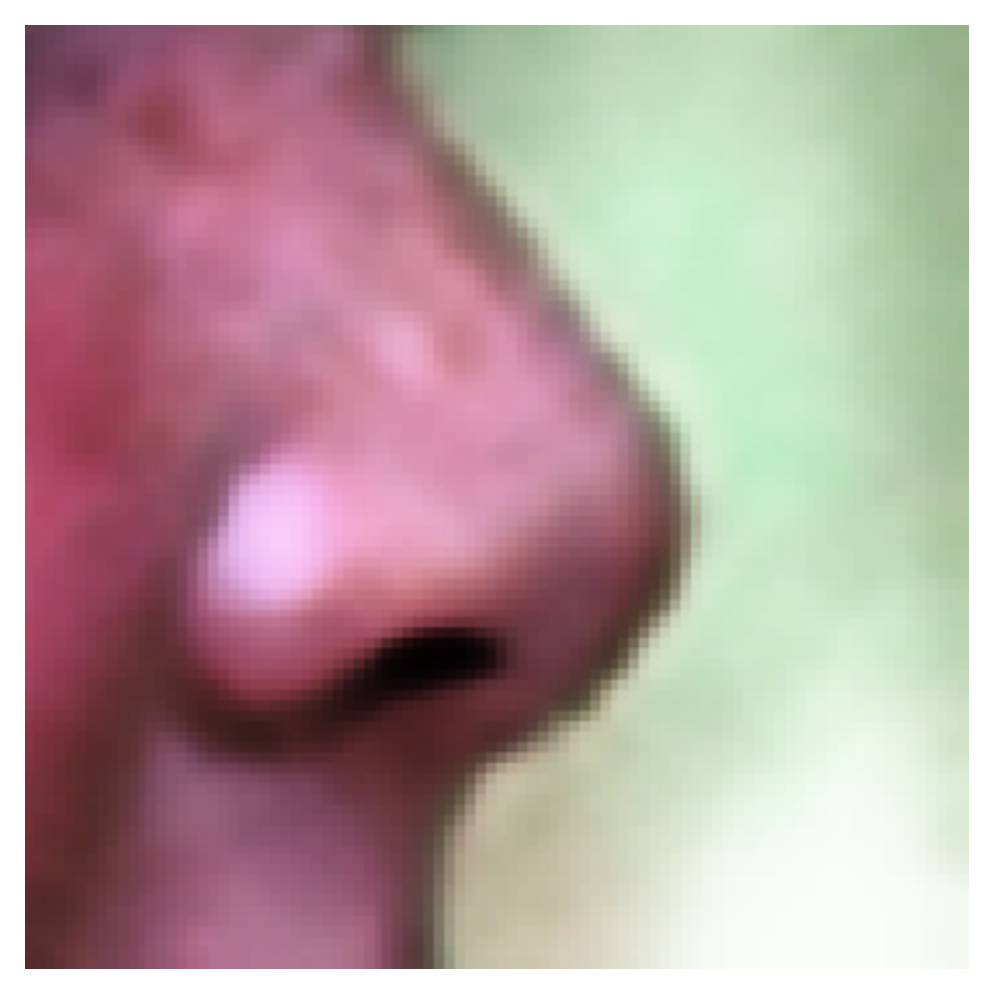}\nsp&
  \includegraphics[width=\fsz]{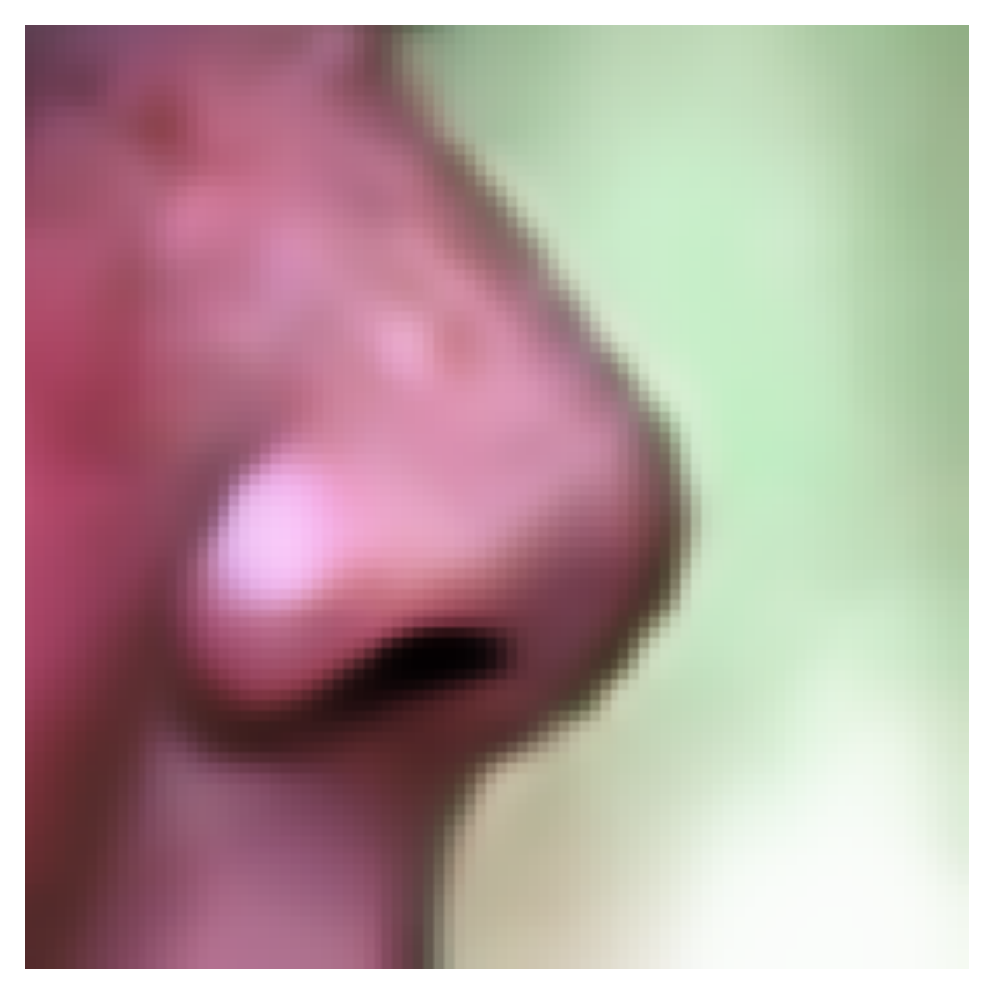}\nsp& 
  \includegraphics[width=\fsz]{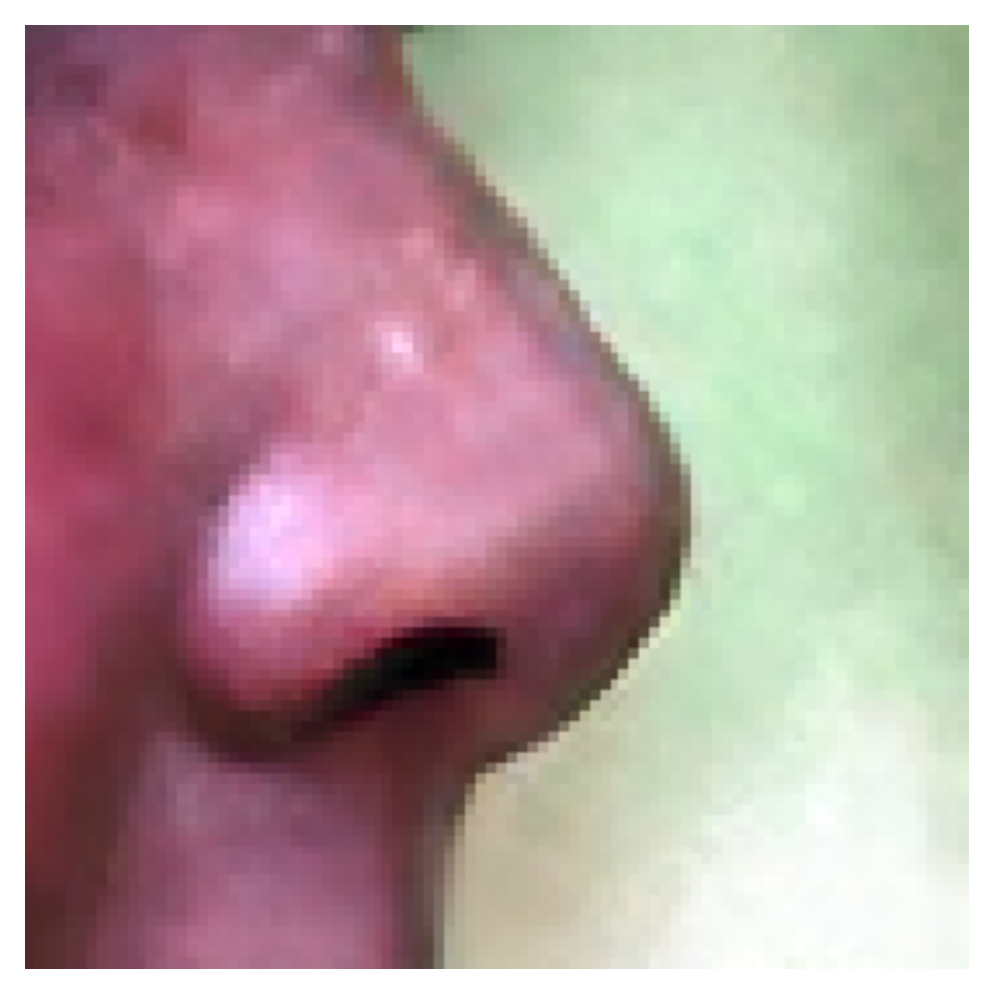}\nsp&
    \includegraphics[width=\fsz]{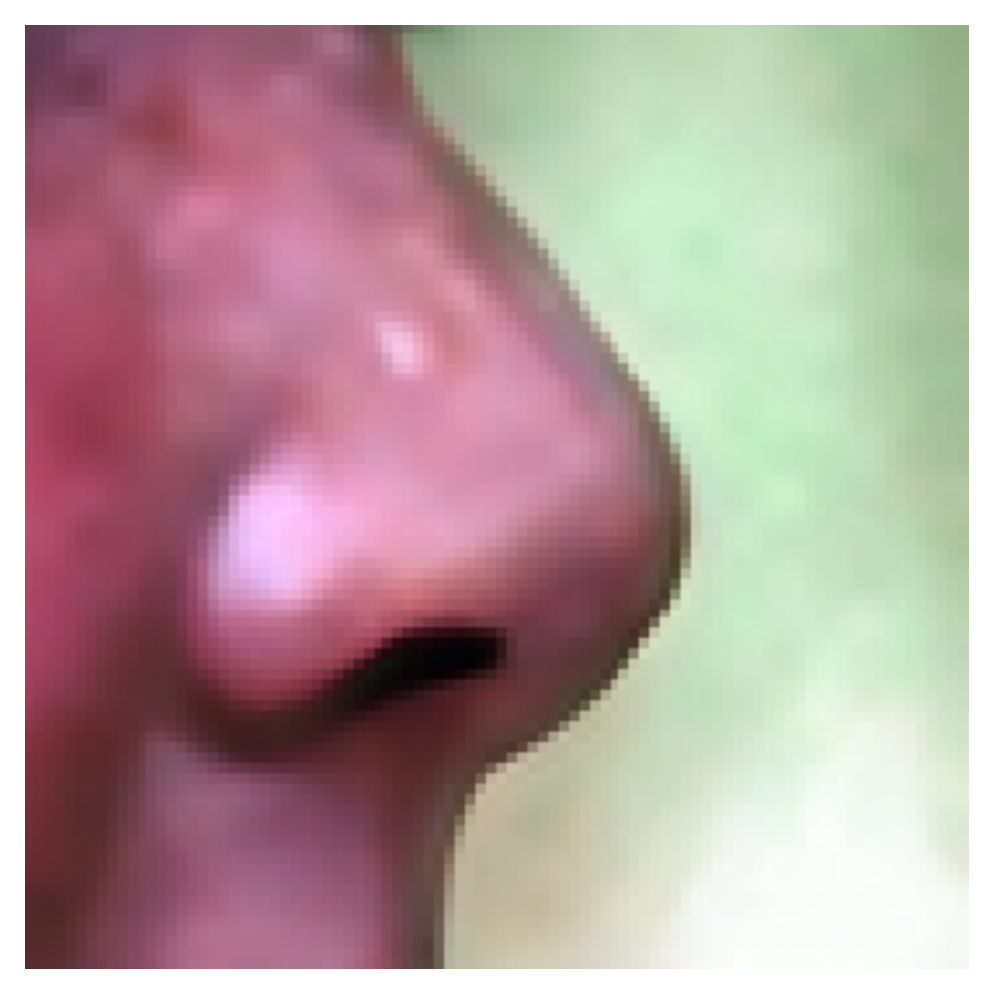}\\
  \includegraphics[width=\fsz]{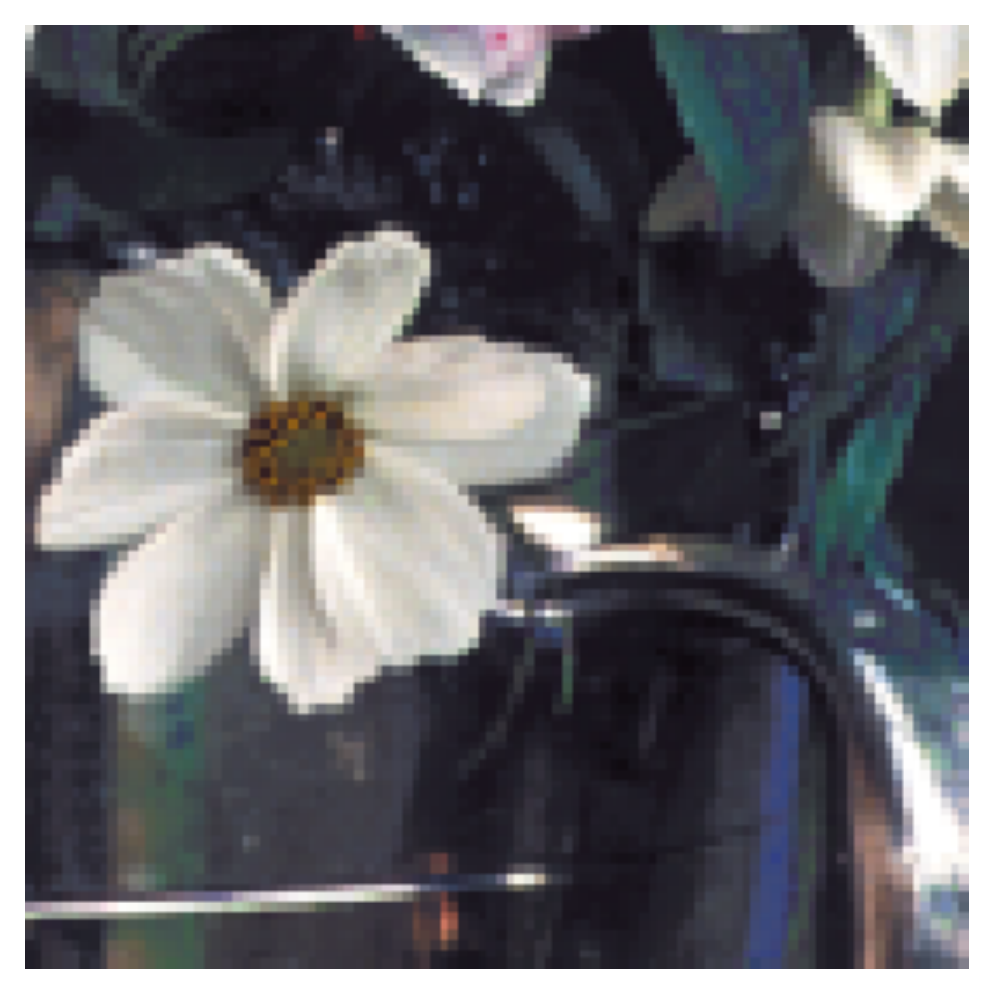}\nsp& 
  \includegraphics[width=\fsz]{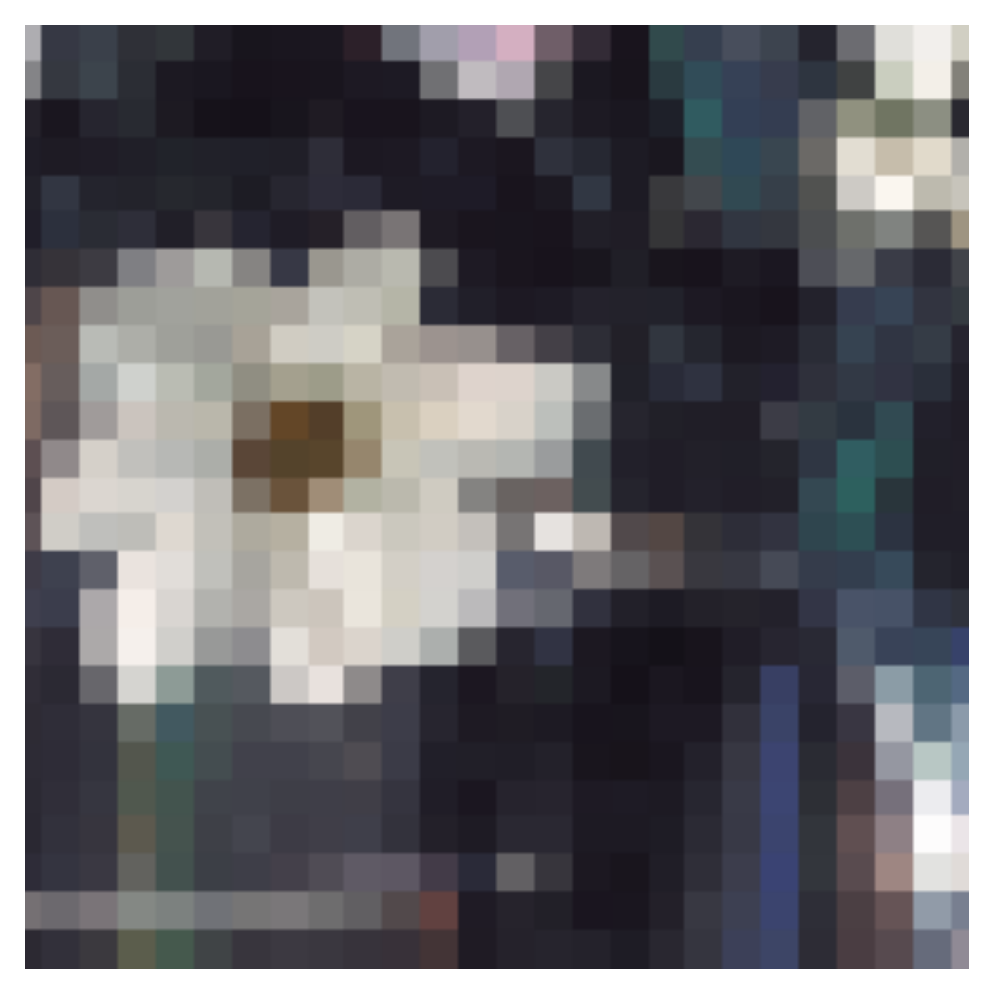}\nsp&
  \includegraphics[width=\fsz]{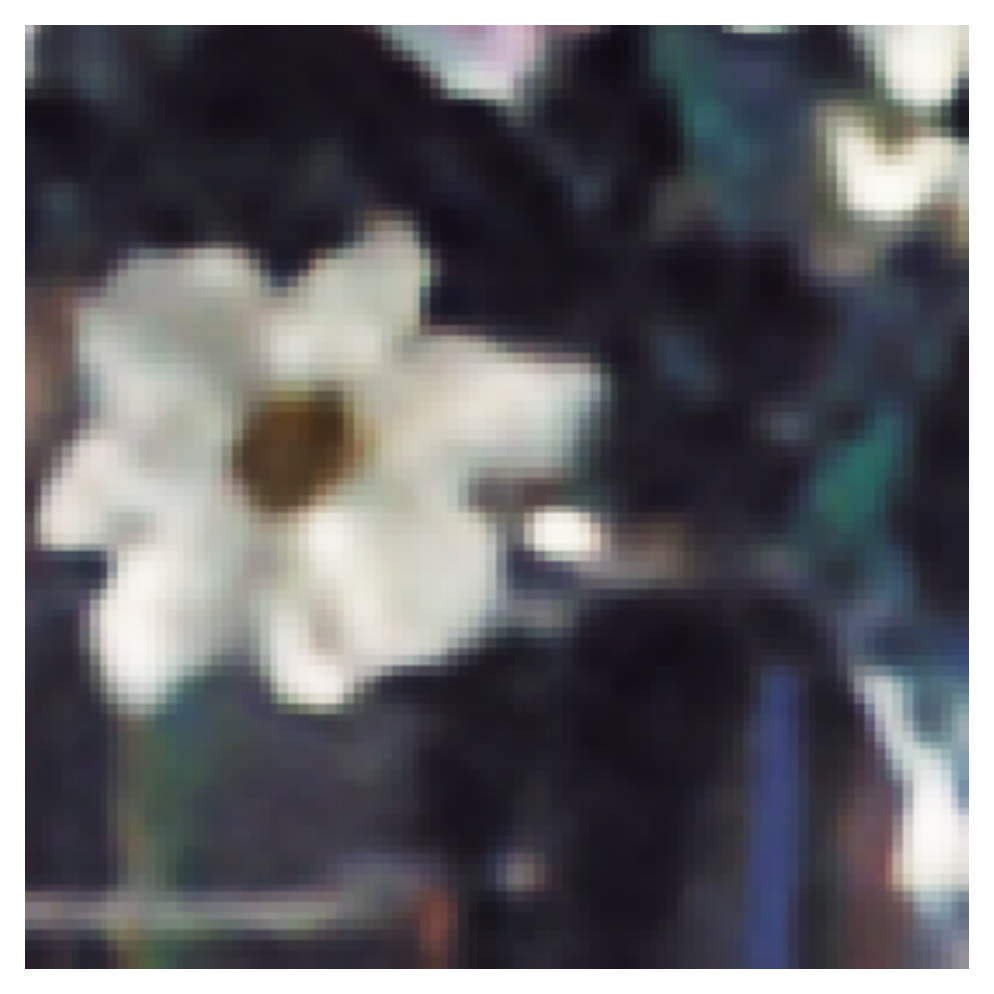}\nsp &
  \includegraphics[width=\fsz]{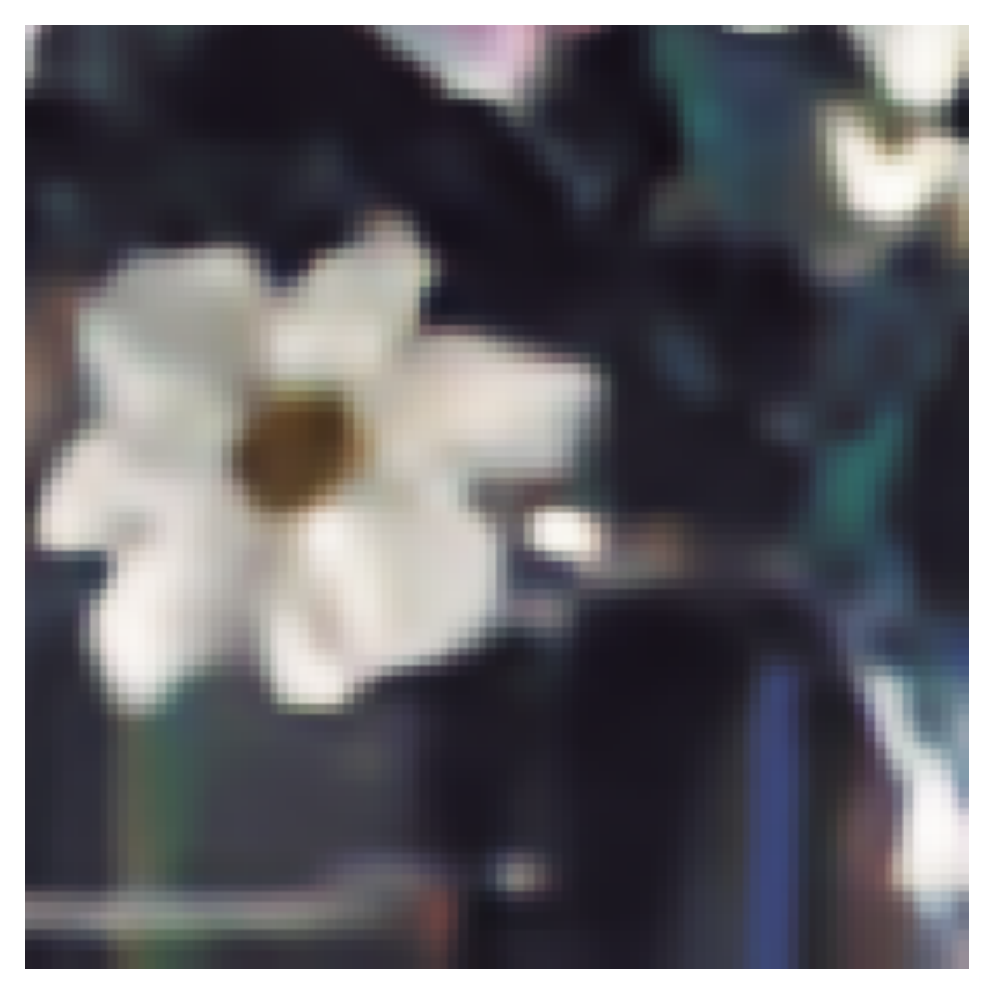}\nsp&
    \includegraphics[width=\fsz]{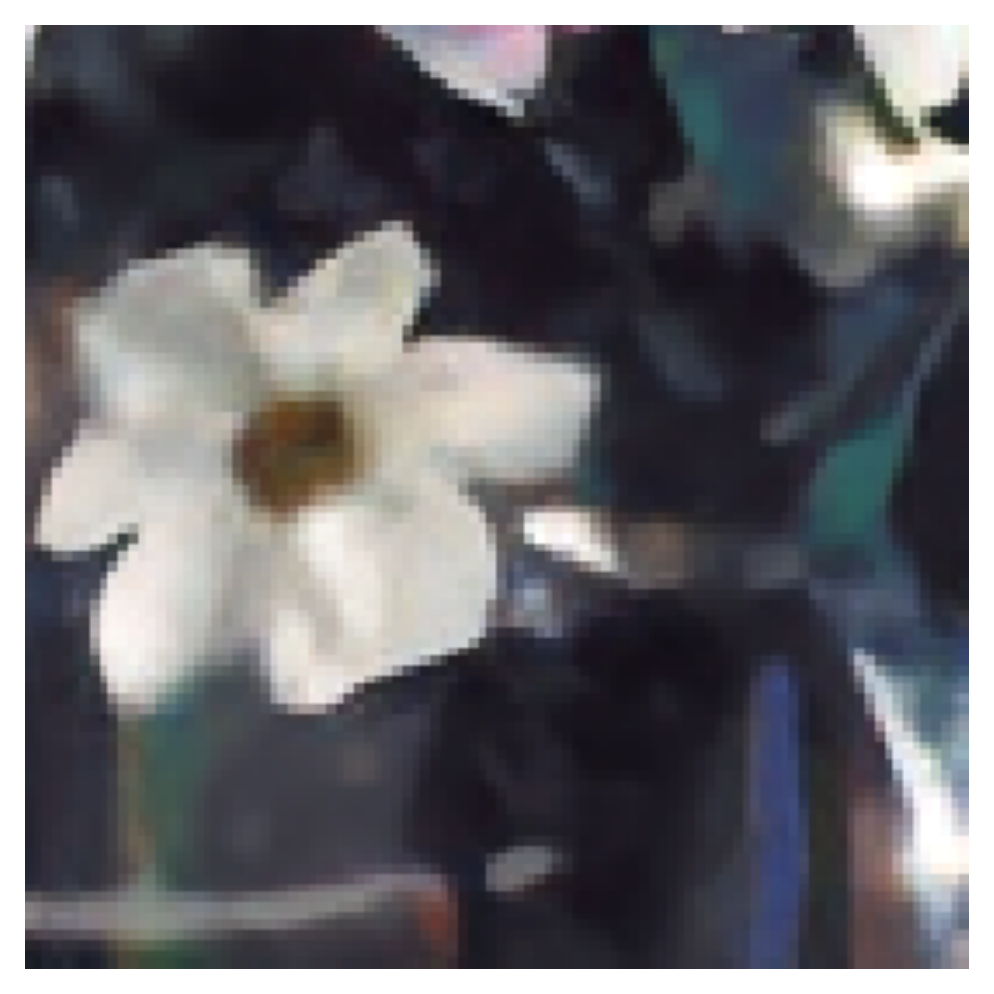}\nsp&
        \includegraphics[width=\fsz]{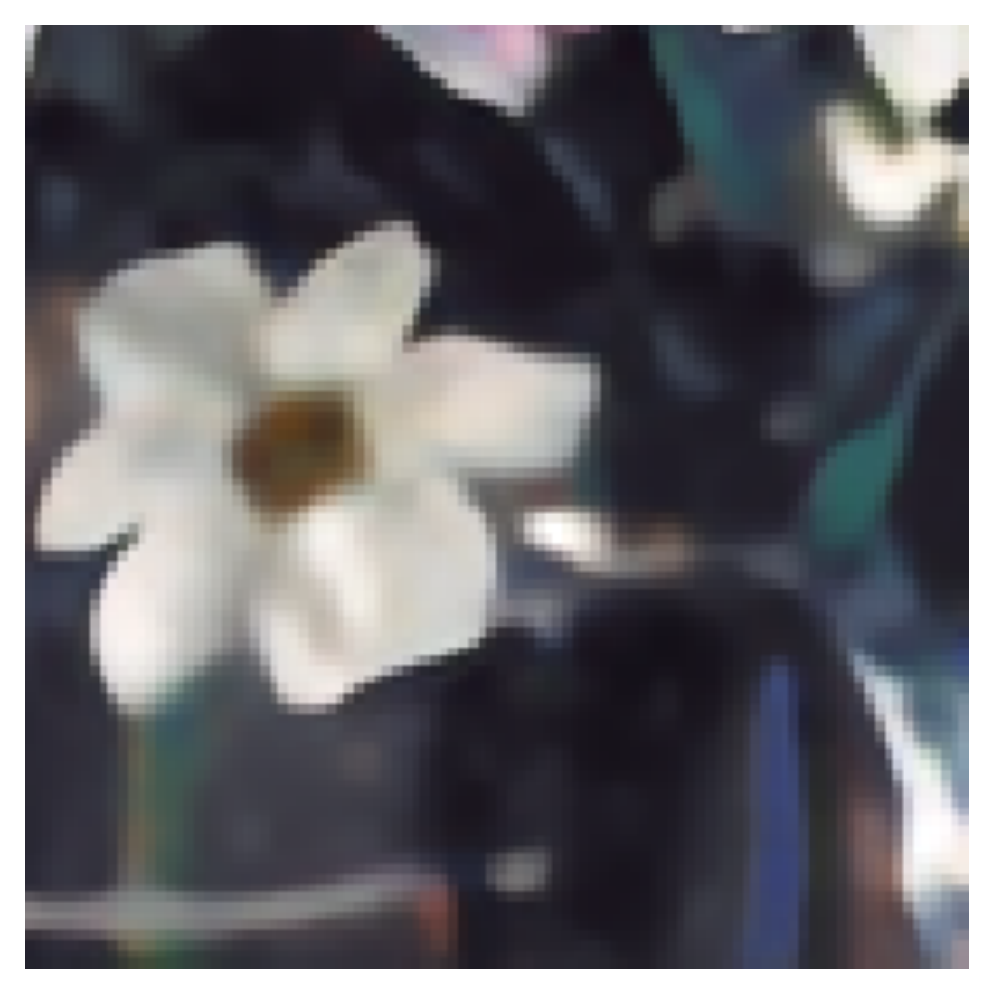}\\
  \includegraphics[width=\fsz]{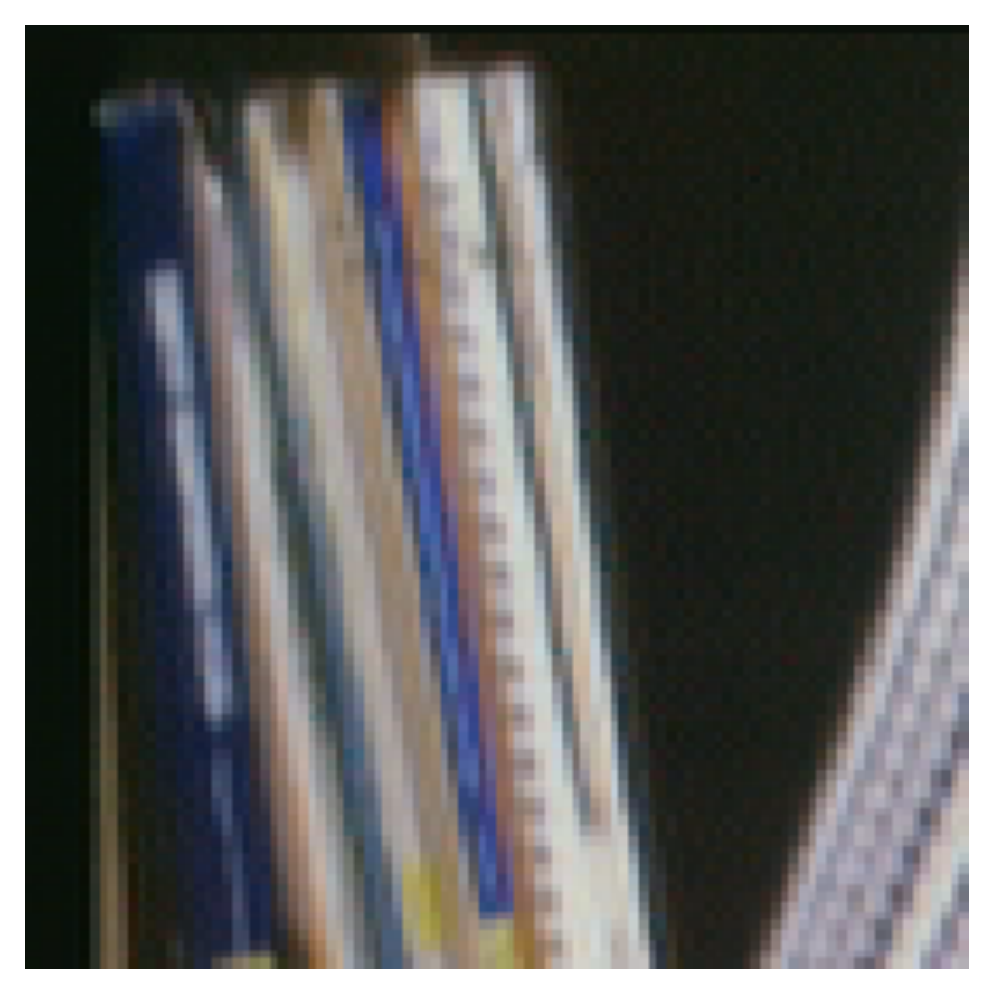}\nsp &
  \includegraphics[width=\fsz]{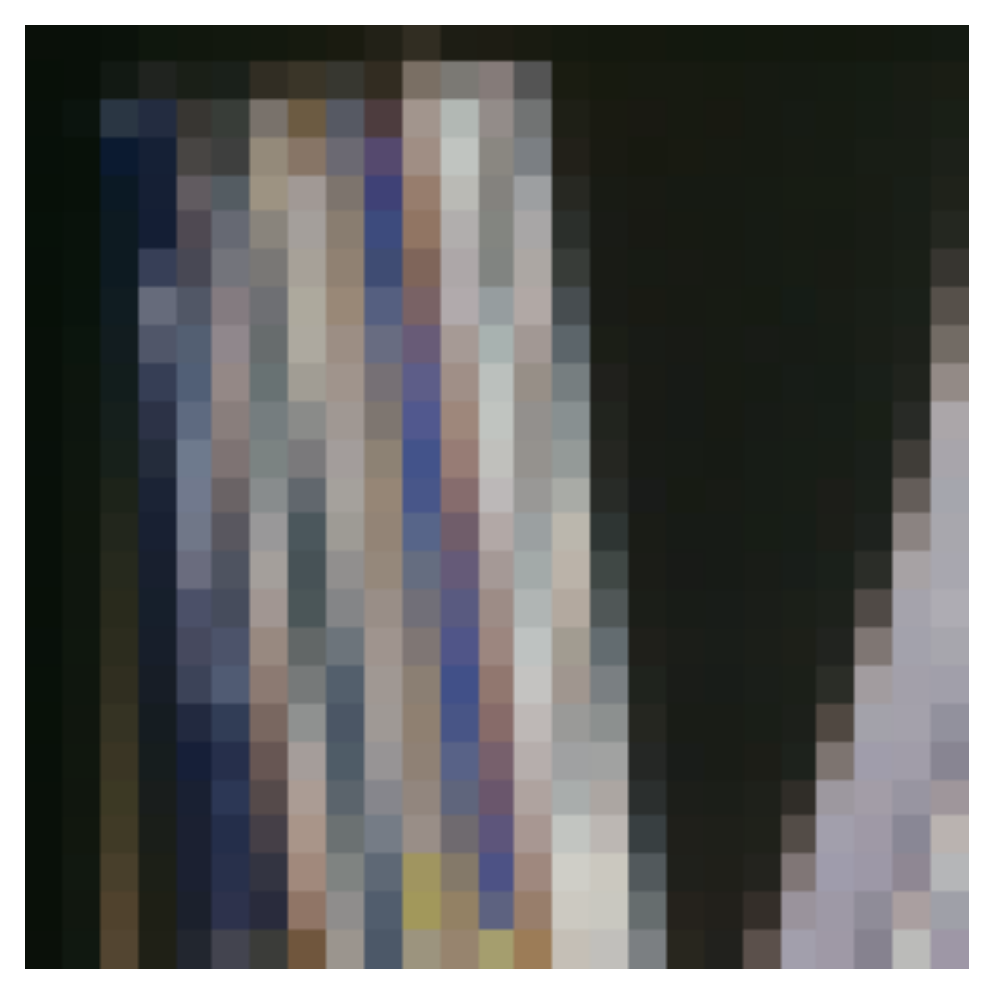}\nsp&
  \includegraphics[width=\fsz]{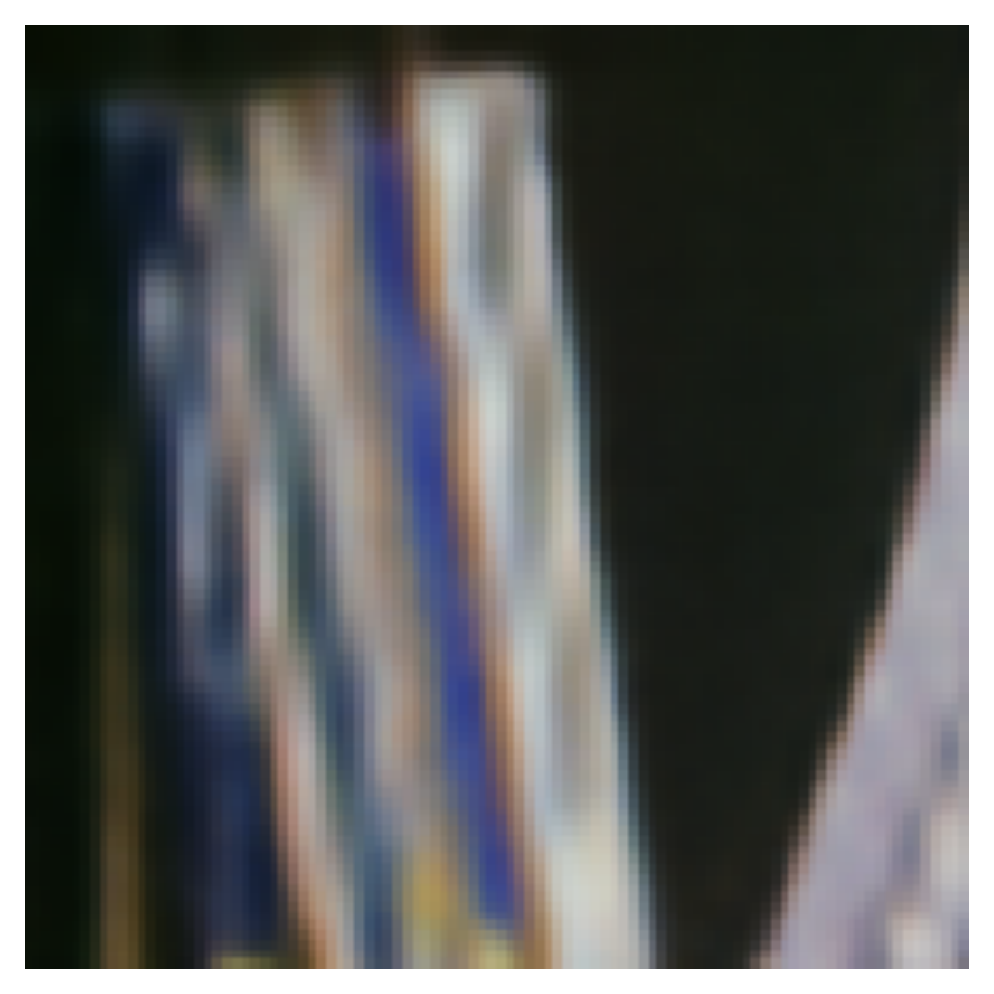}\nsp &
  \includegraphics[width=\fsz]{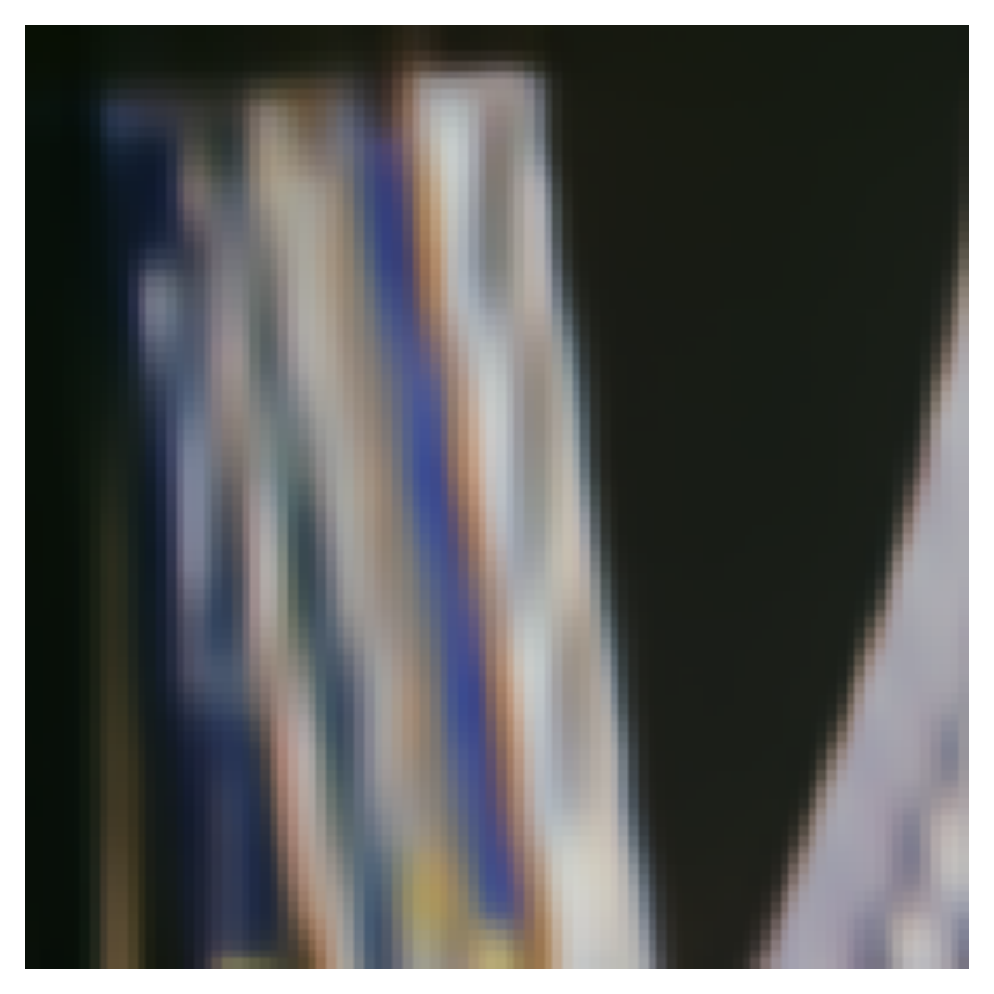}\nsp&
    \includegraphics[width=\fsz]{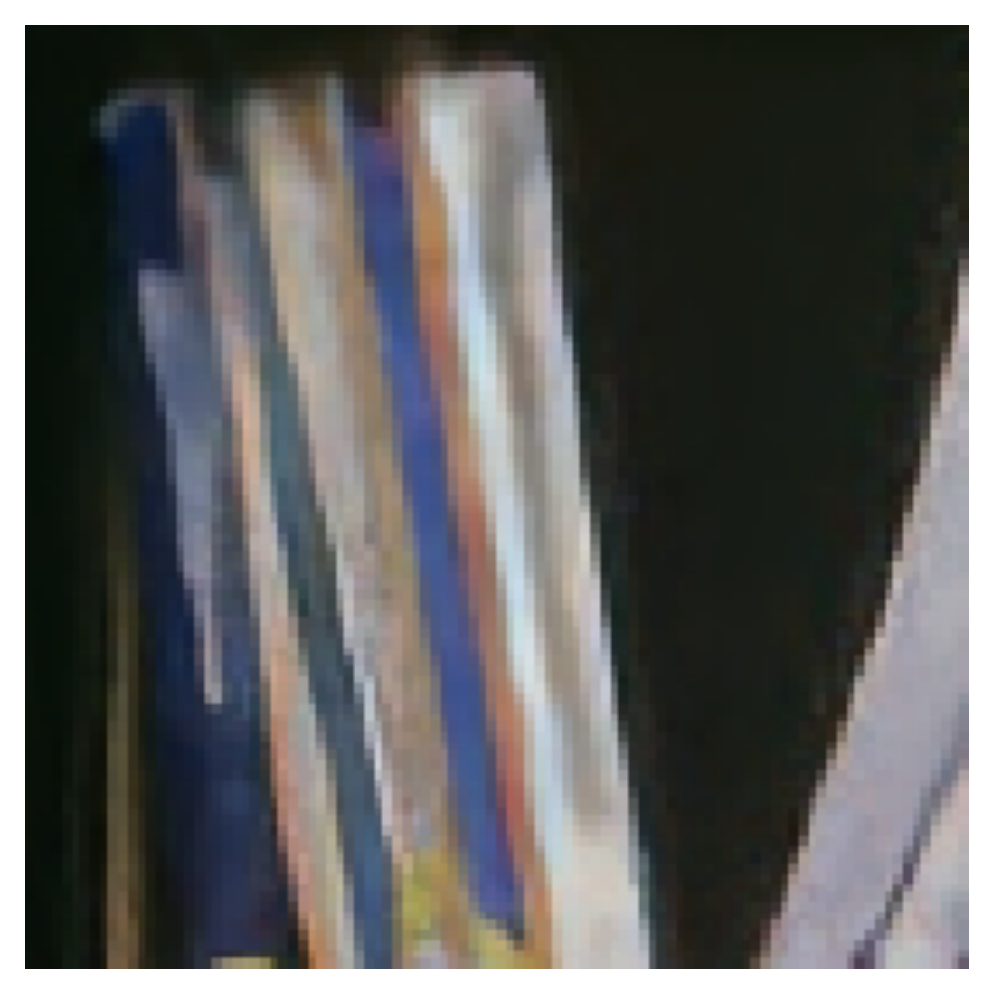}\nsp&
      \includegraphics[width=\fsz]{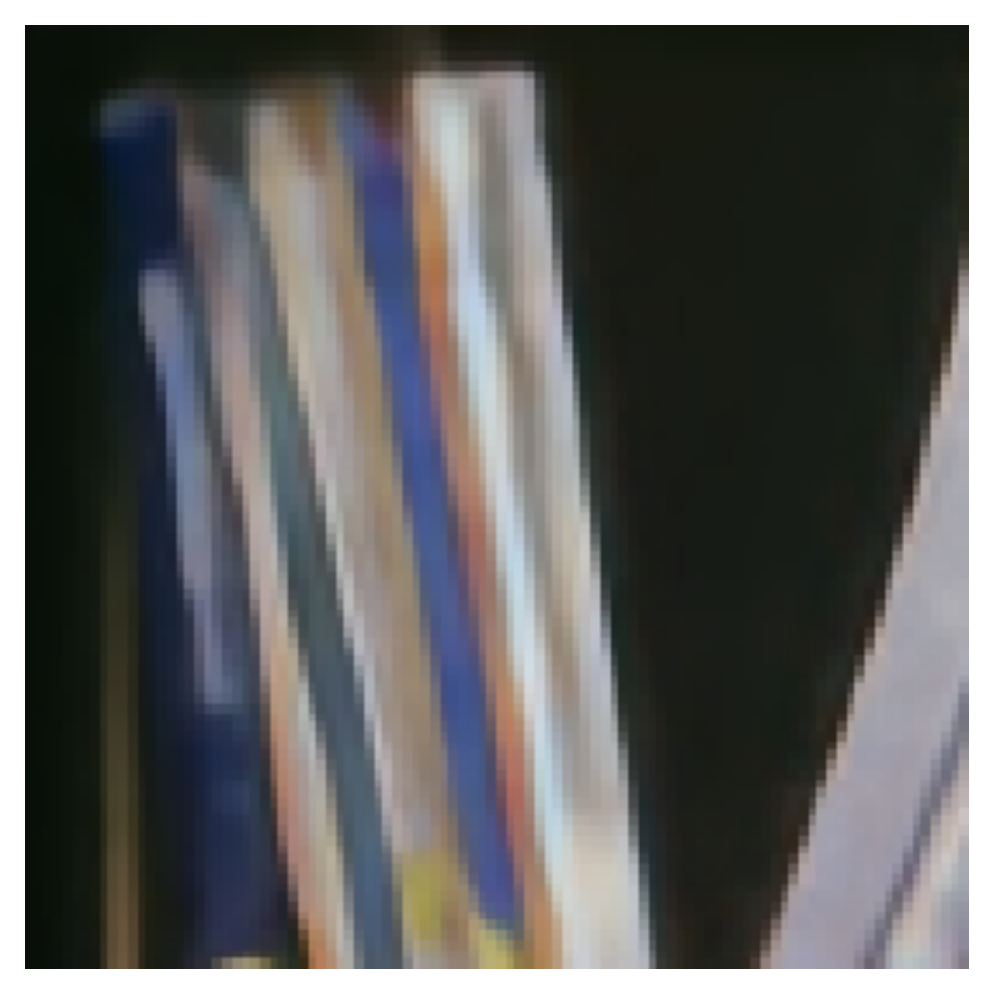}\\[-0.5ex]
   \nsp\footnotesize{cropped}\nsp&
   \nsp\footnotesize{low res}\nsp&
   \nsp\footnotesize{DIP}\nsp&
   \nsp\footnotesize{DeepRED}\nsp&
   \nsp\footnotesize{Ours}\nsp&
   \nsp\footnotesize{Ours - avg}\nsp\\
\end{tabular}
\caption{{Spatial super-resolution.} First column show three original images (cropped portion).  Second column shows cropped portion with resolution reduced by averaging over 4x4 blocks (dimensionality reduction to $6.25\%$).
Next three columns show reconstruction results obtained using 
DIP \cite{Ulyanov2020DeepImagePrior},
DeepRED \cite{mataev2019deepred},
and our method.  Last column shows an average over 10 samples from our method, which is blurrier but with better PSNR (see Tables \ref{tab:sr_set5} and \ref{tab:sr_set14}).
} 
\label{fig:SR4}
\end{figure}

\begin{table}
  \centering
  \begin{tabular}{llllll}
    \toprule
    \multicolumn{6}{c}{Algorithm}                   \\
    \cmidrule(r){2-6}
    Factor   & $MM^Tx$ &  DIP     & DeepRED & Ours  & Ours - avg\\
    \midrule
    4:1 & 26.35 (0.826) & 30.04 (0.902) & 30.22 (0.904) & 29.47 (0.894) & {\bf 31.20} ({\bf 0.913}) \\
    8:1 & 23.02 (0.673) & 24.98 (0.760) & 24.95 (0.760) & 25.07 (0.767) & {\bf 25.64} ({\bf 0.792}) \\
    \bottomrule \\
  \end{tabular}
   \caption{Spatial super-resolution performance averaged over Set5 images. Values indicate YCbCr-PSNR (SSIM).}
  \label{tab:sr_set5}
  \end{table}

\begin{table}
  \centering
  \begin{tabular}{llllll}
    \toprule
    \multicolumn{6}{c}{Algorithm}      \\
    \cmidrule(r){2-6}
    Factor  & $MM^Tx$ & DIP     & DeepRED & Ours  & Ours - avg\\
    \midrule
    4:1 & 24.65 (0.765)  & 26.88 (0.815) & 27.01 (0.817) & 26.56 (0.808) & {\bf 27.14} ({\bf 0.826}) \\
    8:1 & 22.06 (0.628) & 23.33 (0.685) & 23.34 (0.685) & 23.32 (0.681) & {\bf 23.78} ({\bf 0.703}) \\
    \bottomrule \\
  \end{tabular}
  \caption{Spatial super-resolution performance averaged over Set14 images. Values indicate YCbCr-PSNR (SSIM).}
 \label{tab:sr_set14}
  \end{table}

\begin{table}
  \centering
  \begin{tabular}{llll}
    \toprule
    \multicolumn{4}{c}{Algorithm}\\
    \cmidrule(r){2-4}
    Factor  & DIP &  DeepRED & Ours  \\
    \midrule
    4:1 &  1,190  & 1,584 & 9   \\
    \bottomrule\\
  \end{tabular}
  \label{tab:runtime}
  \caption{Run time (in seconds) for super-resolution algorithm, executed on an NVIDIA DGX GPU, averaged over images in Set14.}
\end{table}

\textbf{Spatial super-resolution.}
In this problem, the goal is to construct a high resolution image from a low resolution (i.e. downsampled) image. Downsampling is typically performed after lowpass filtering, and the combination of downsampling factor and filter kernel determines the measurement model, $M$. Here, we use a $4\times 4$ block-averaging filter, and $4 \times 4$ downsampling (i.e., measurements are averages over non-overlapping blocks). For comparison, we also reconstructed high resolution images using Deep image Prior (DIP) \cite{Ulyanov2020DeepImagePrior} and DeepRED \cite{mataev2019deepred}. DIP chooses a random input vector, and adjusts the weights of a CNN to minimize the mean square error between the output and the corrupted image. The authors interpret the denoising capability of this algorithm as arising from inductive biases imposed by the CNN architecture that favor clean natural images over those corrupted by artifacts or noise. By stopping the optimization early, the authors obtain surprisingly good solutions to the linear inverse problem. 
Regularization by denoising (RED) develops a MAP solution, by using a least squares denoiser as a regularizer \citep{Romano17}.  
DeepRED \cite{mataev2019deepred} combines DIP and RED, obtaining better performance than either method alone. 
Results for three example images are shown in Figure \ref{fig:SR4}. 
In all three cases, our method produces an image that is sharper with less noticeable artifacts than the others.  Despite this, the PSNR and SSIM values are slightly worse (see Tables \ref{tab:sr_set5} and \ref{tab:sr_set14}).  These can be improved by averaging over realizations (i.e., running the algorithm with different random initializations), at the expense of some blurring (see last column of Figure \ref{fig:SR4}). We can interpret this in the context of the prior embedded in our denoiser: if each super-resolution reconstruction corresponds to a point on the (curved) manifold of natural images, then the average (a convex combination of those points) will lie off the manifold. This illustrates the point made earlier that comparison to ground truth (e.g. PSNR, SSIM) is not particularly meaningful when the measurement matrix is very low-rank.
Finally, our method is more than two orders of magnitude faster than either DIP or DeepRED, as can be seen from average execution times provided in Table \ref{tab:runtime}.

\begin{figure}
\centering
\def\fsz{0.163\linewidth} \def\nsp{\hspace*{-.015\linewidth}}
\begin{tabular}{cccccc}
  \nsp\includegraphics[width=\fsz]{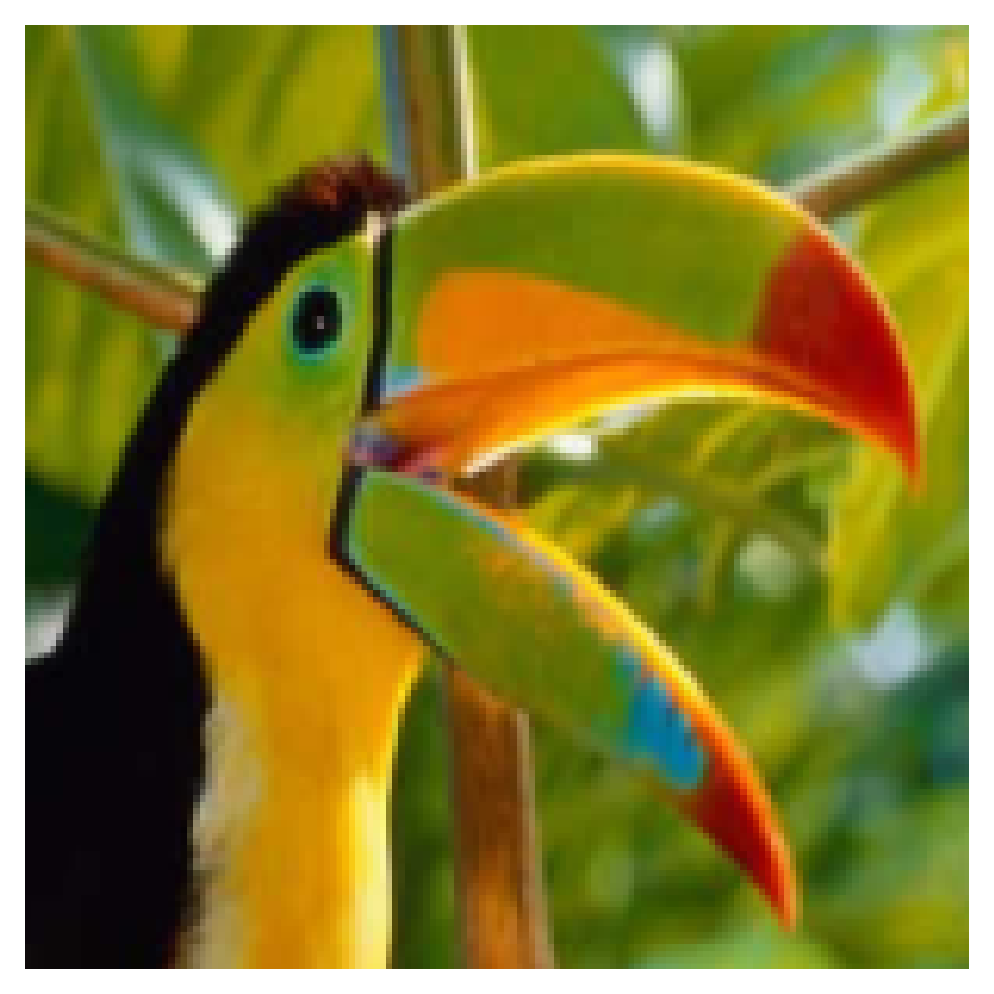}\nsp &
  \nsp\includegraphics[width=\fsz]{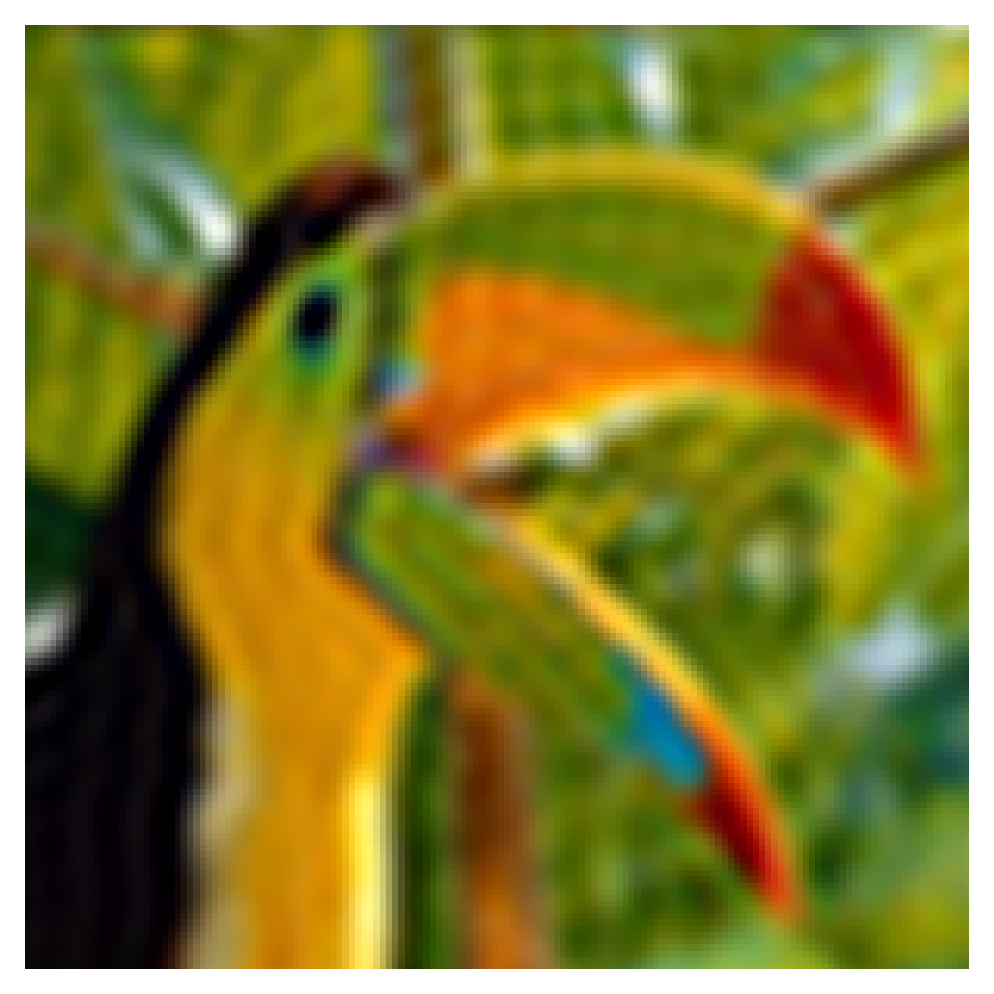}\nsp&
  \nsp\includegraphics[width=\fsz]{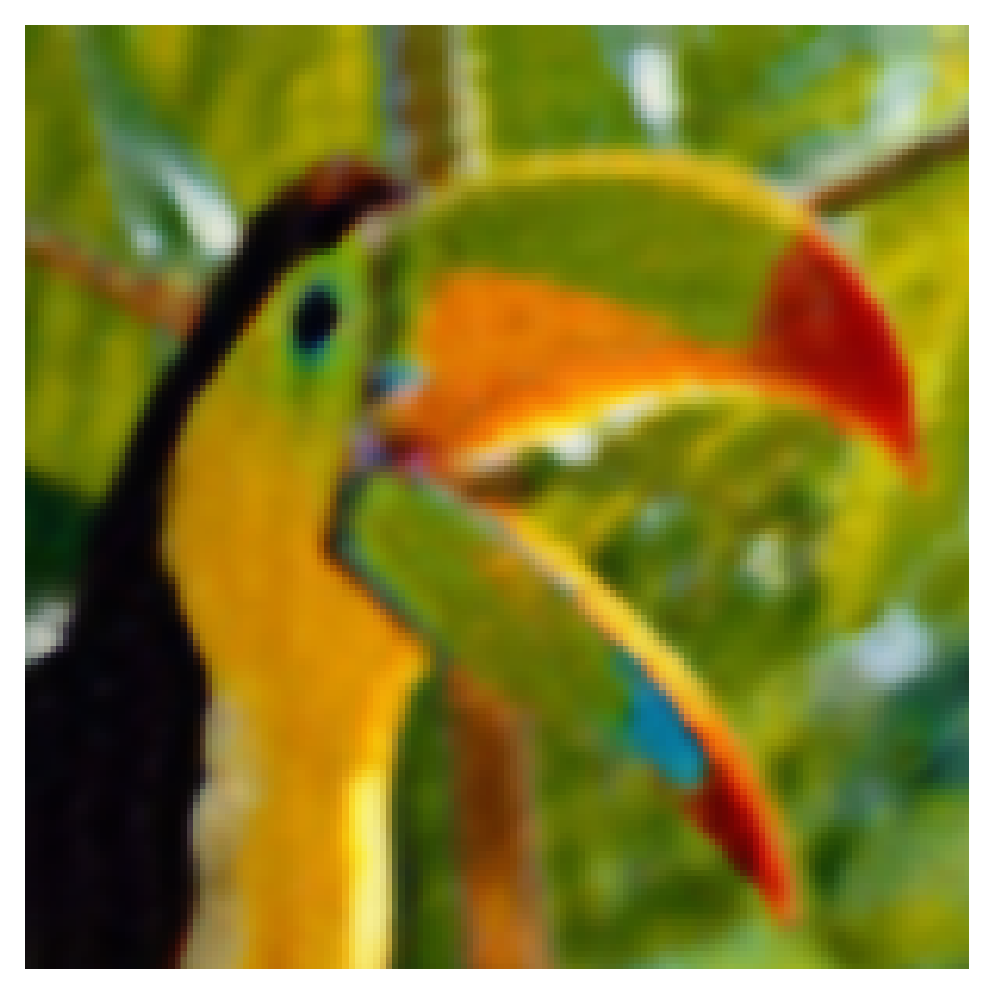}\nsp&
  \nsp\includegraphics[width=\fsz]{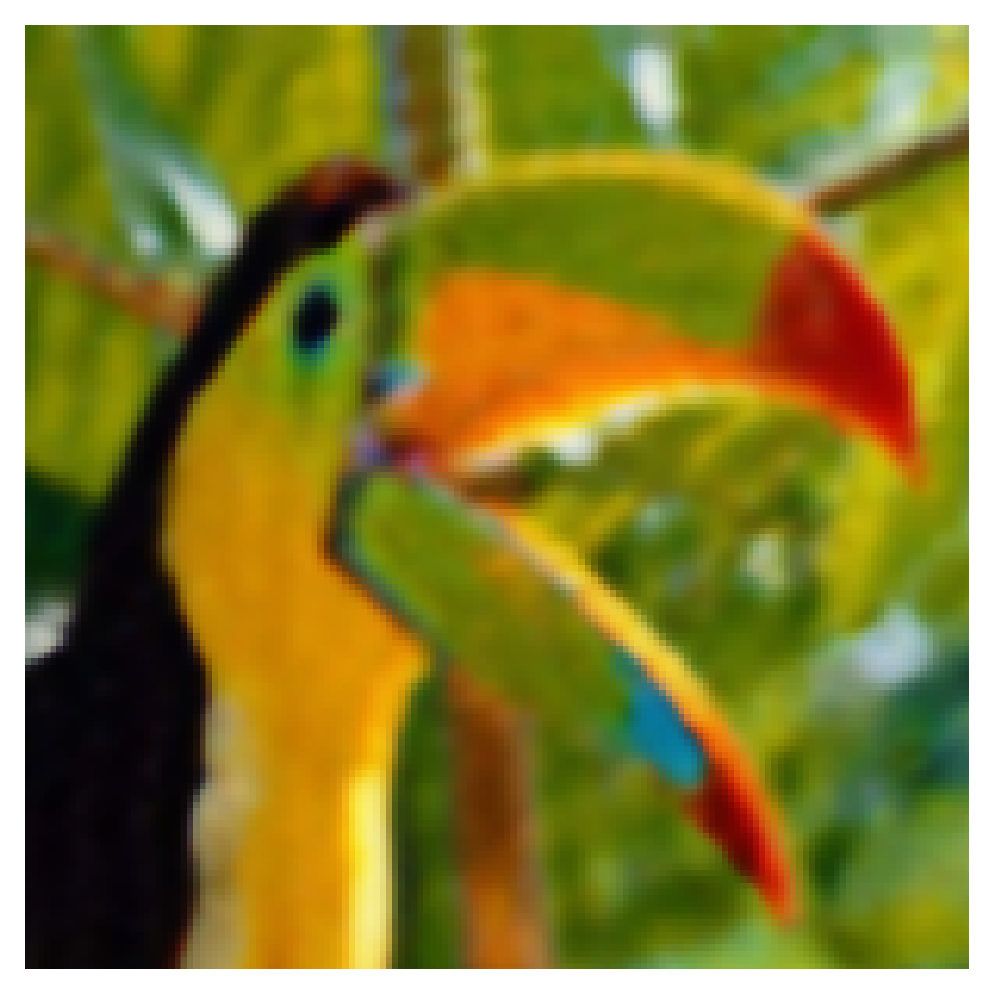}\nsp &
  \nsp\includegraphics[width=\fsz]{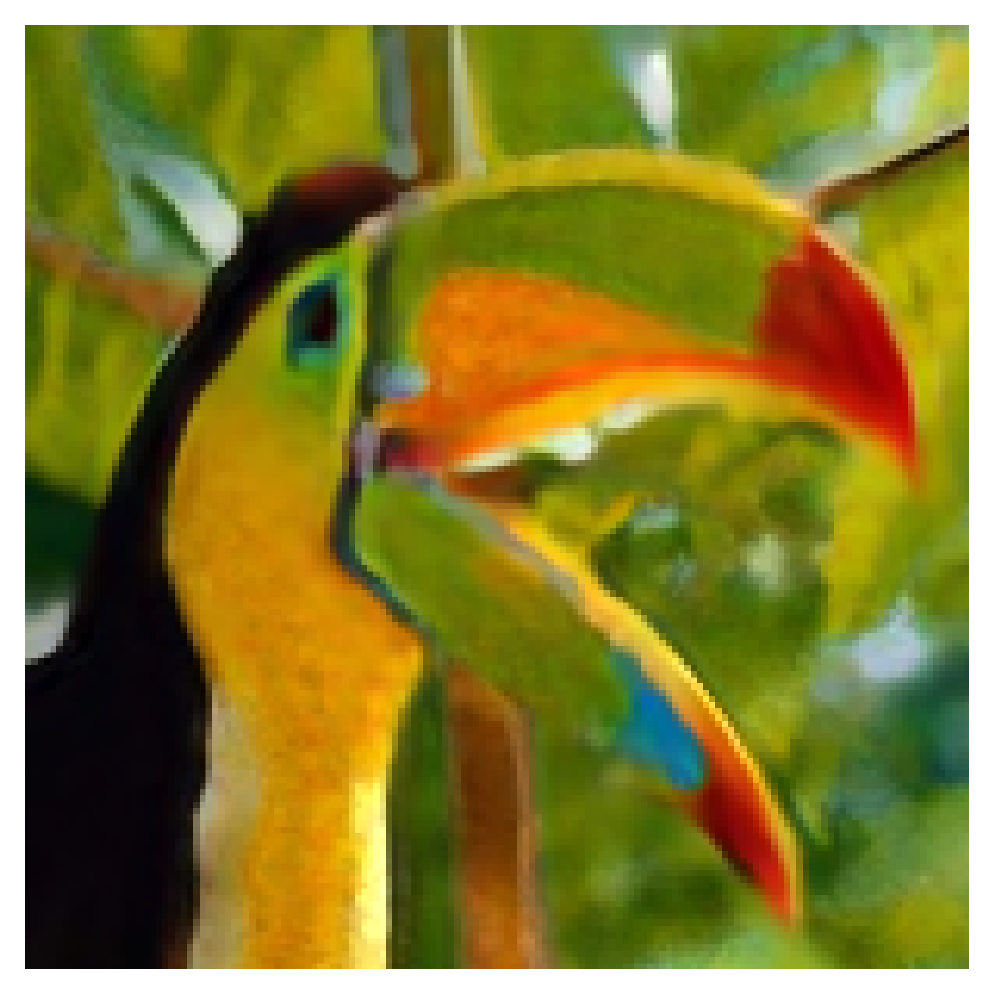}\nsp & 
  \nsp\includegraphics[width=\fsz]{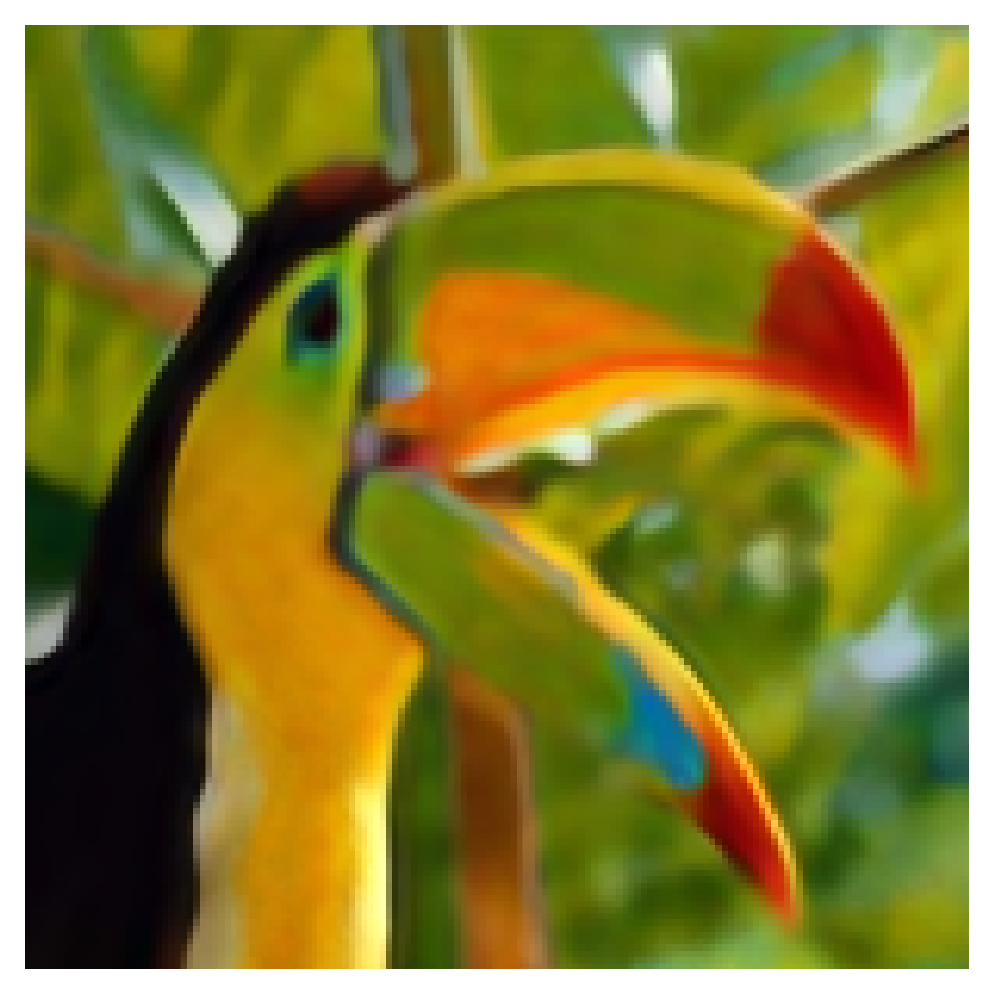}\nsp\\
  \nsp\includegraphics[width=\fsz]{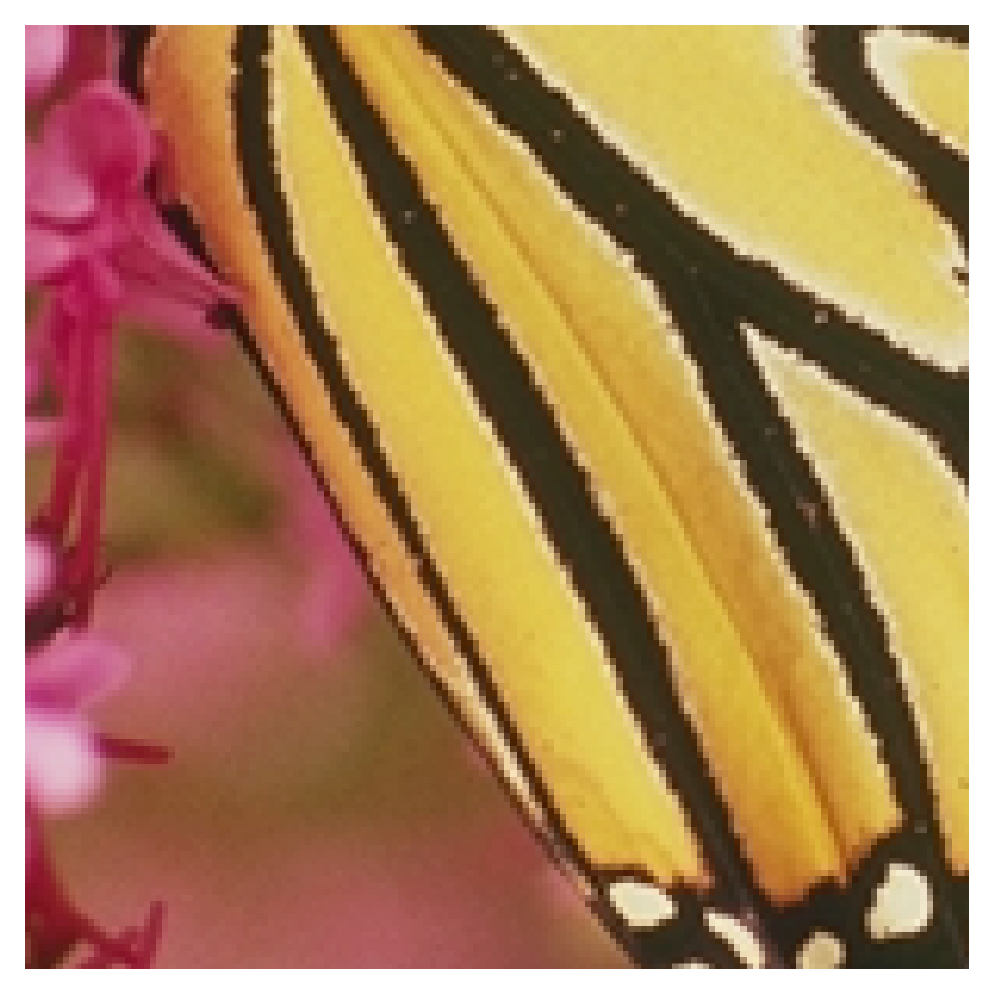}\nsp &  
  \nsp\includegraphics[width=\fsz]{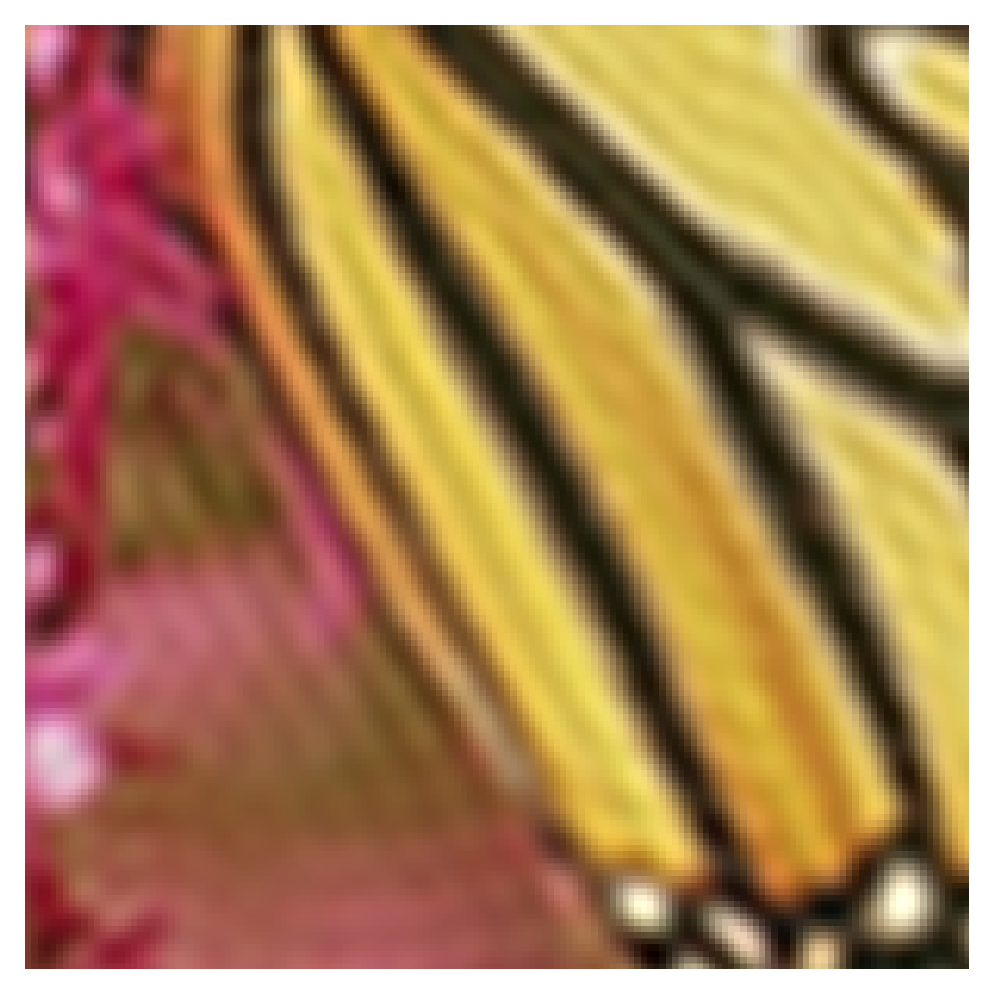}\nsp&    
  \nsp\includegraphics[width=\fsz]{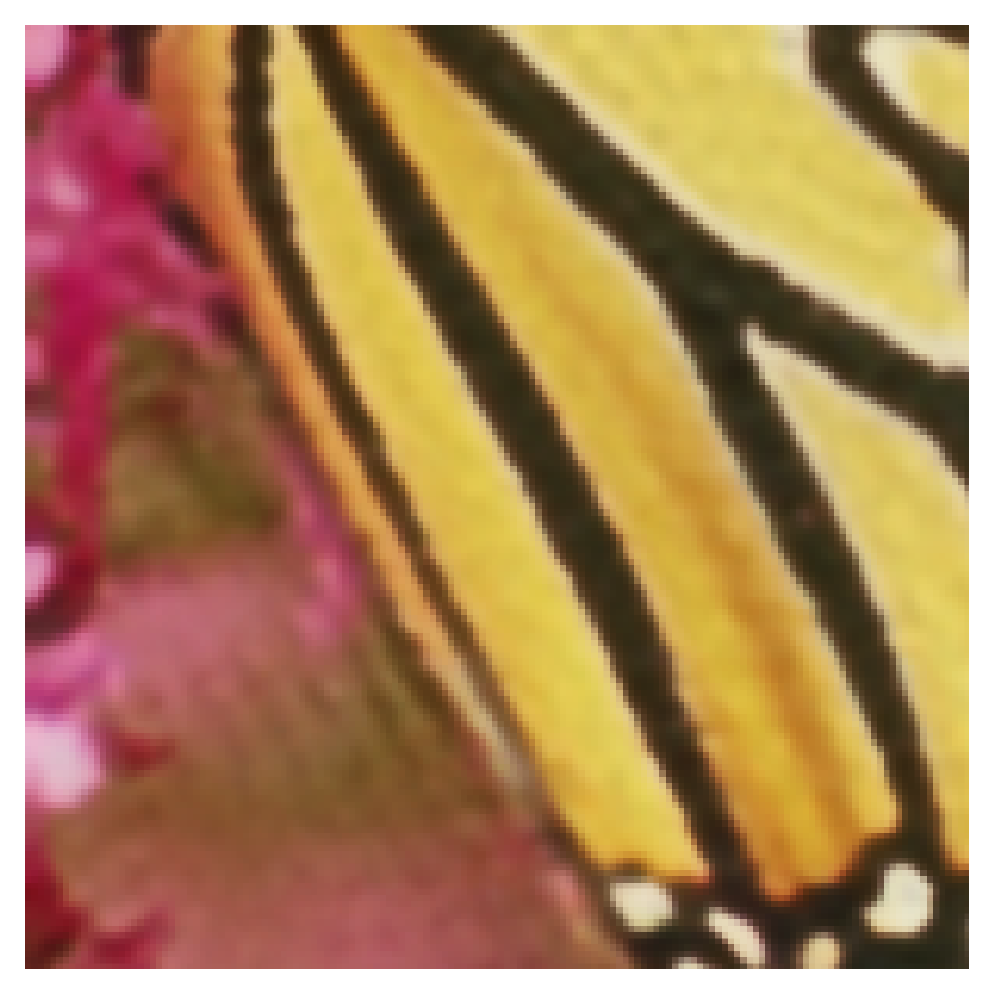}\nsp &
  \nsp\includegraphics[width=\fsz]{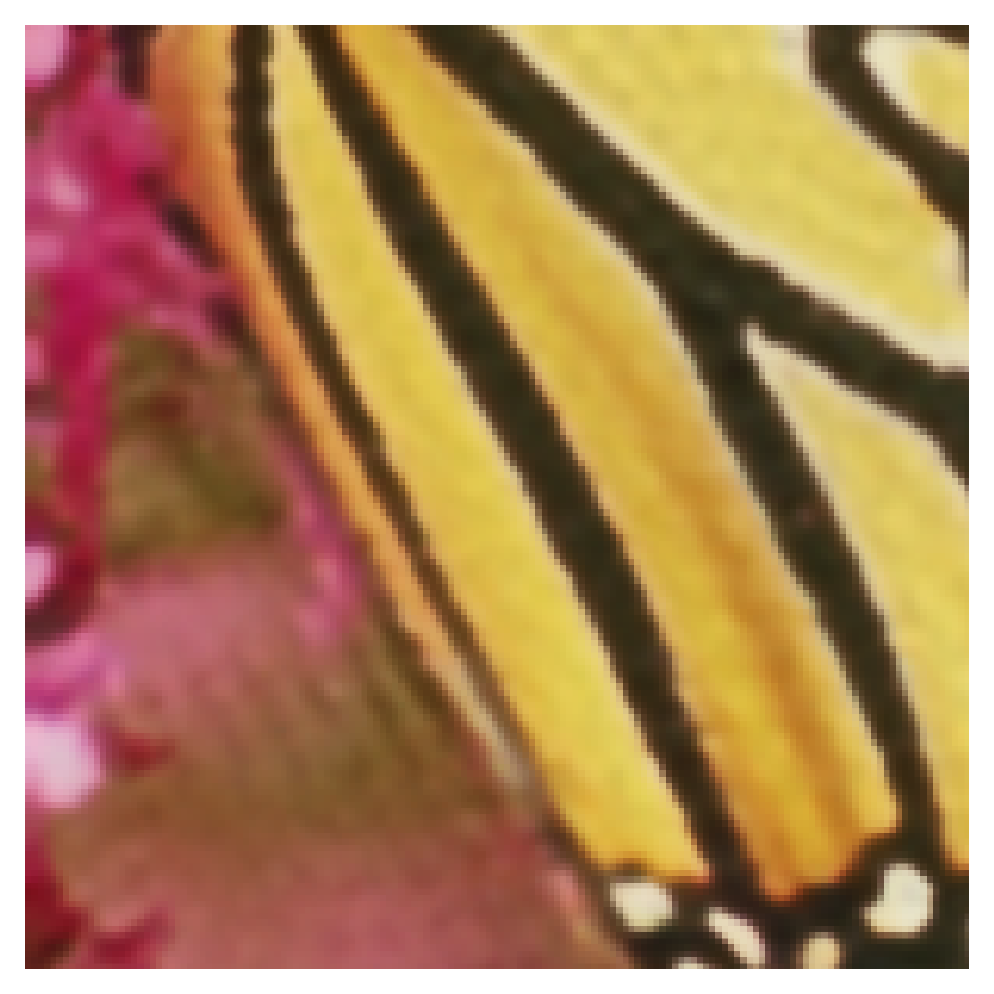}\nsp &
  \nsp\includegraphics[width=\fsz]{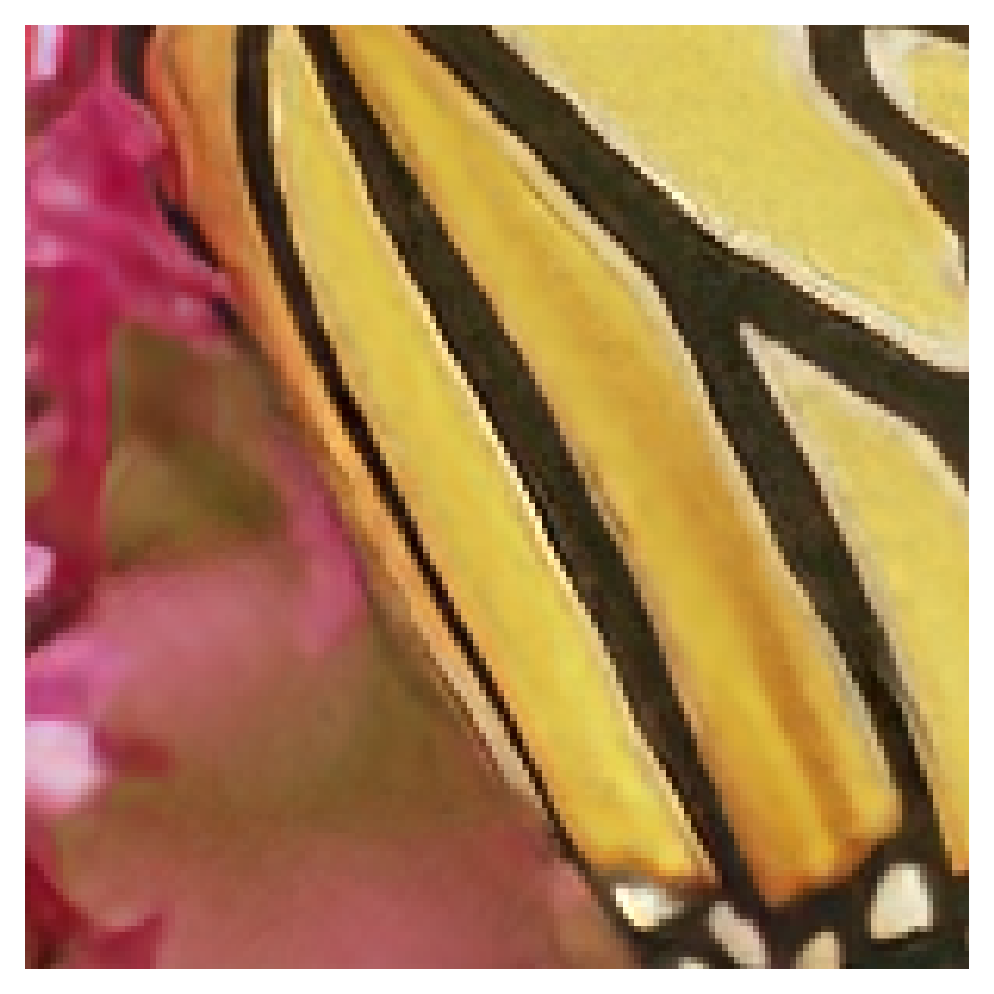}\nsp &  \nsp\includegraphics[width=\fsz]{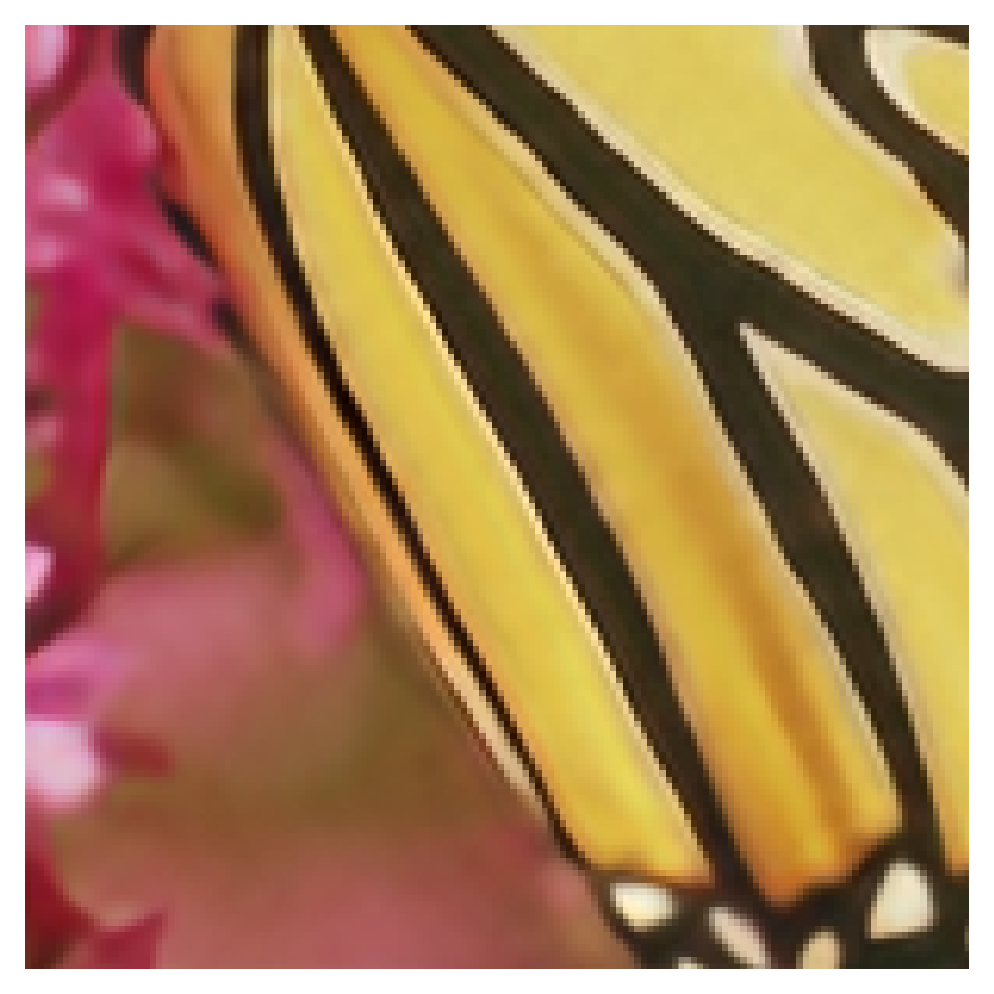}\nsp\\
  \nsp\includegraphics[width=\fsz]{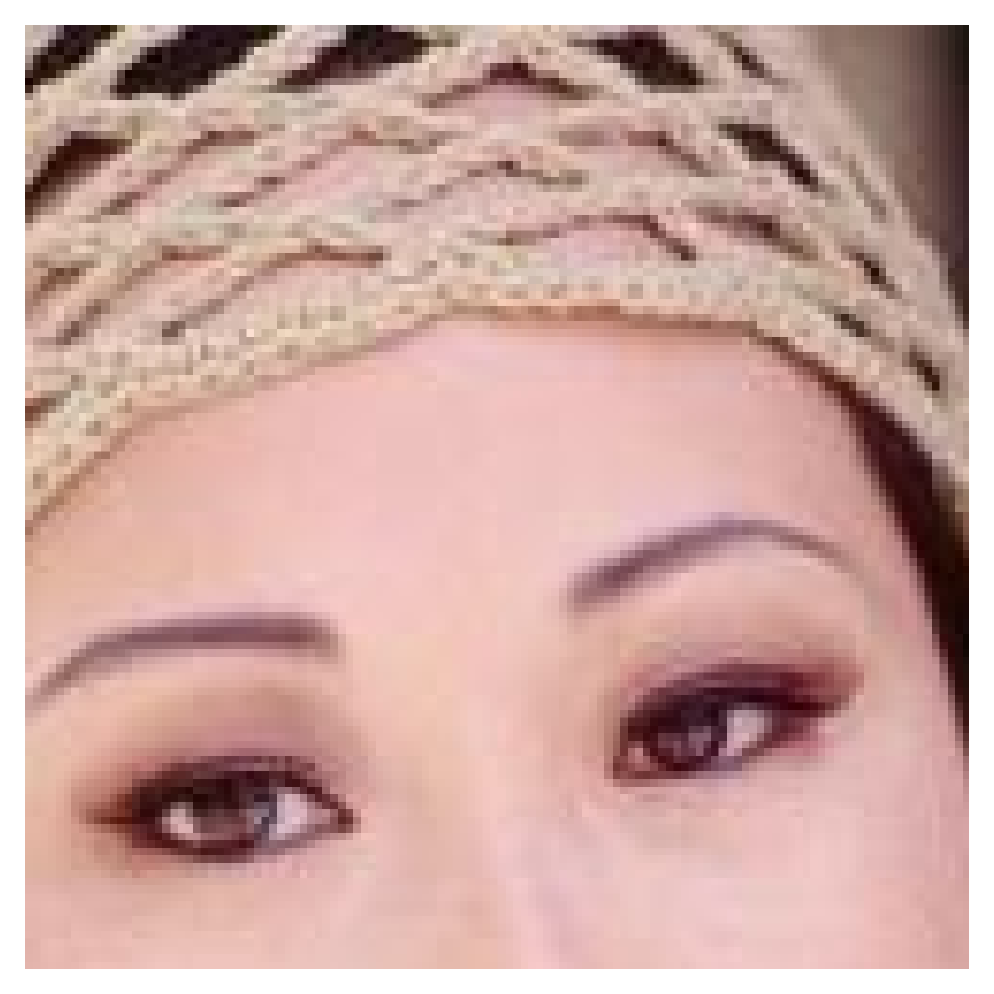}\nsp & 
  \nsp\includegraphics[width=\fsz]{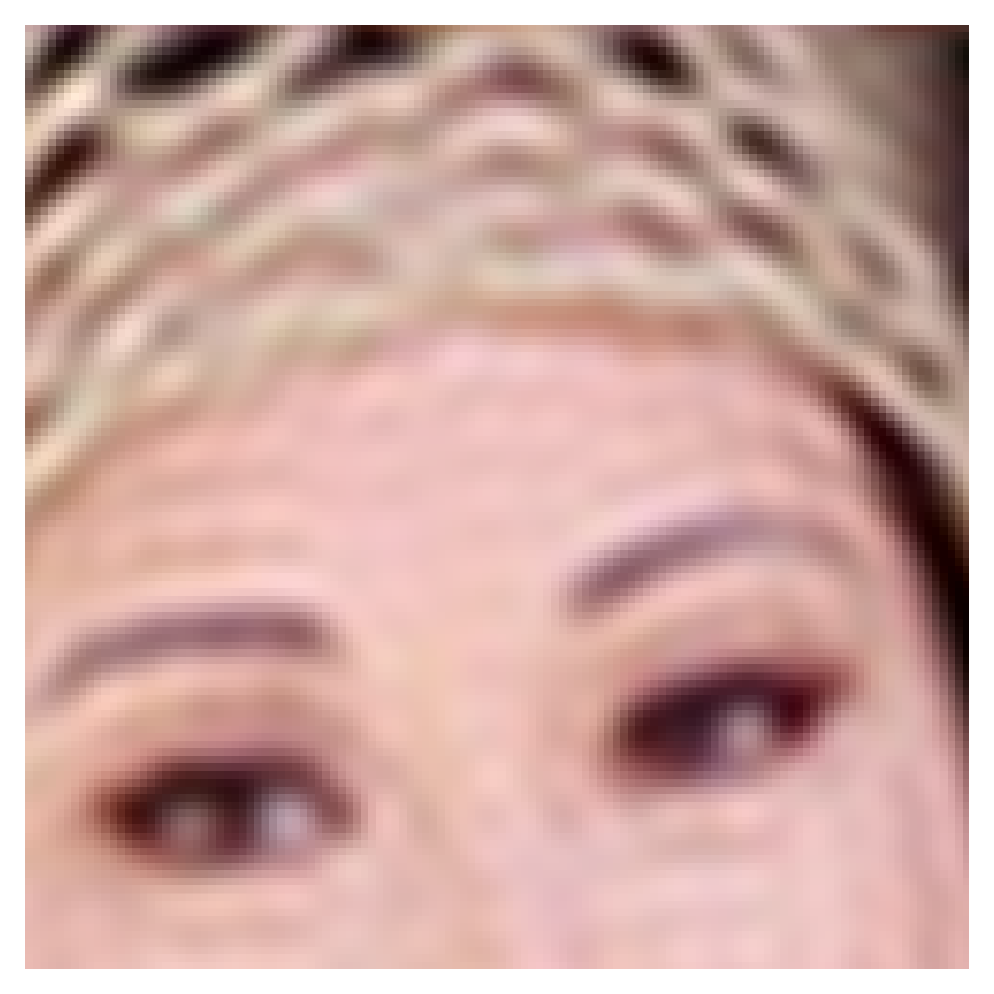}\nsp &  
  \nsp\includegraphics[width=\fsz]{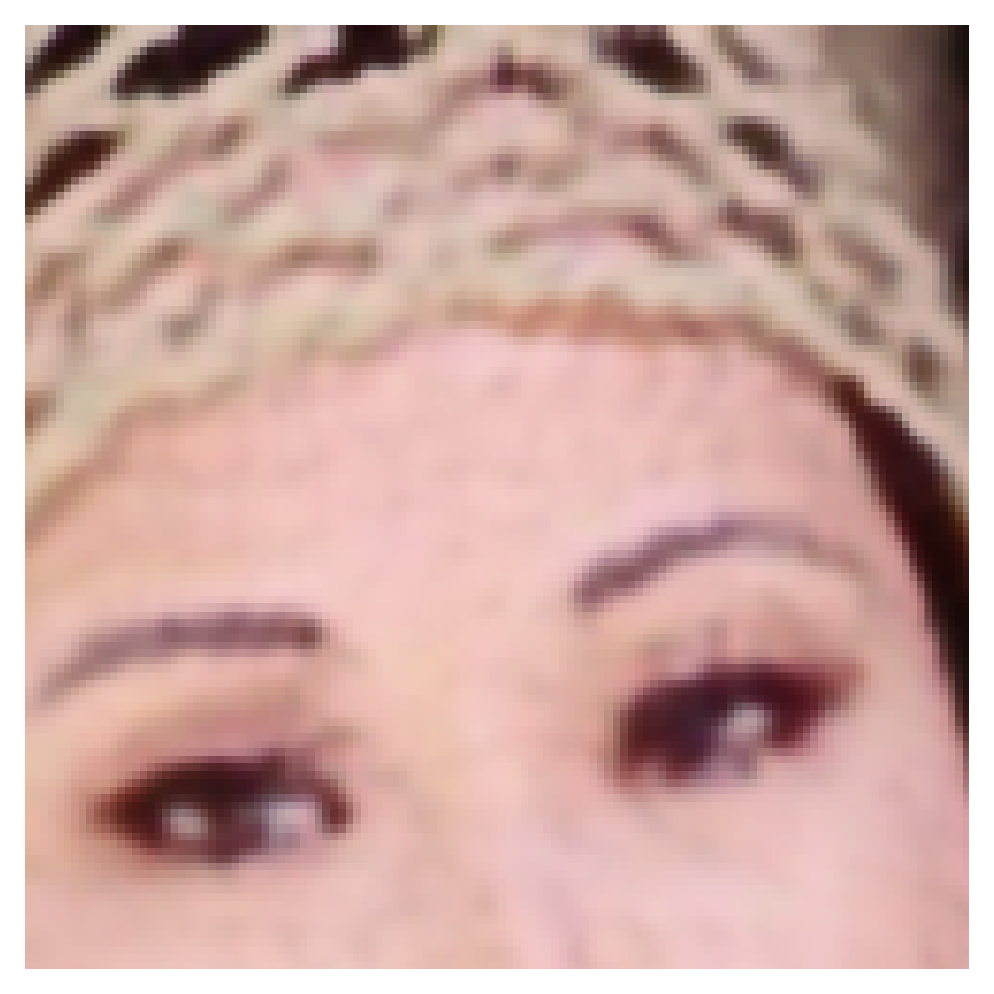}\nsp&
  \nsp\includegraphics[width=\fsz]{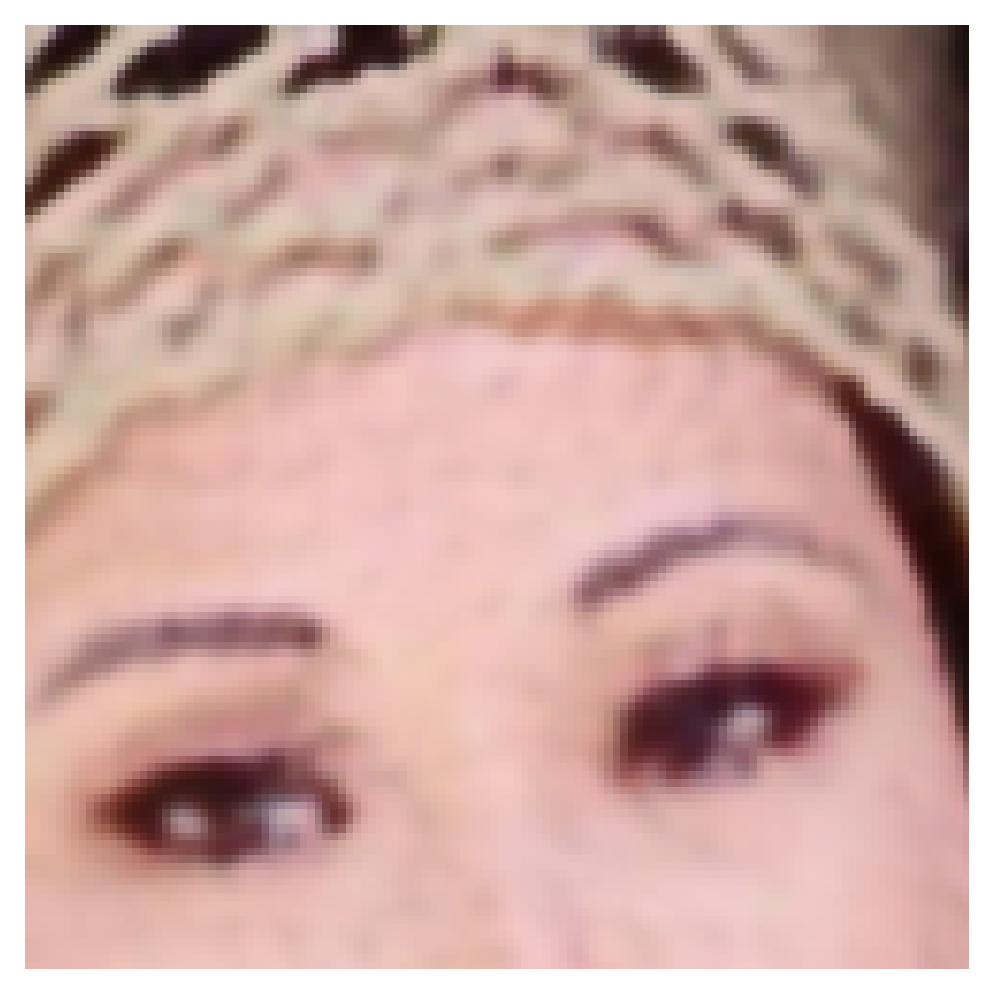}\nsp& 
  \nsp\includegraphics[width=\fsz]{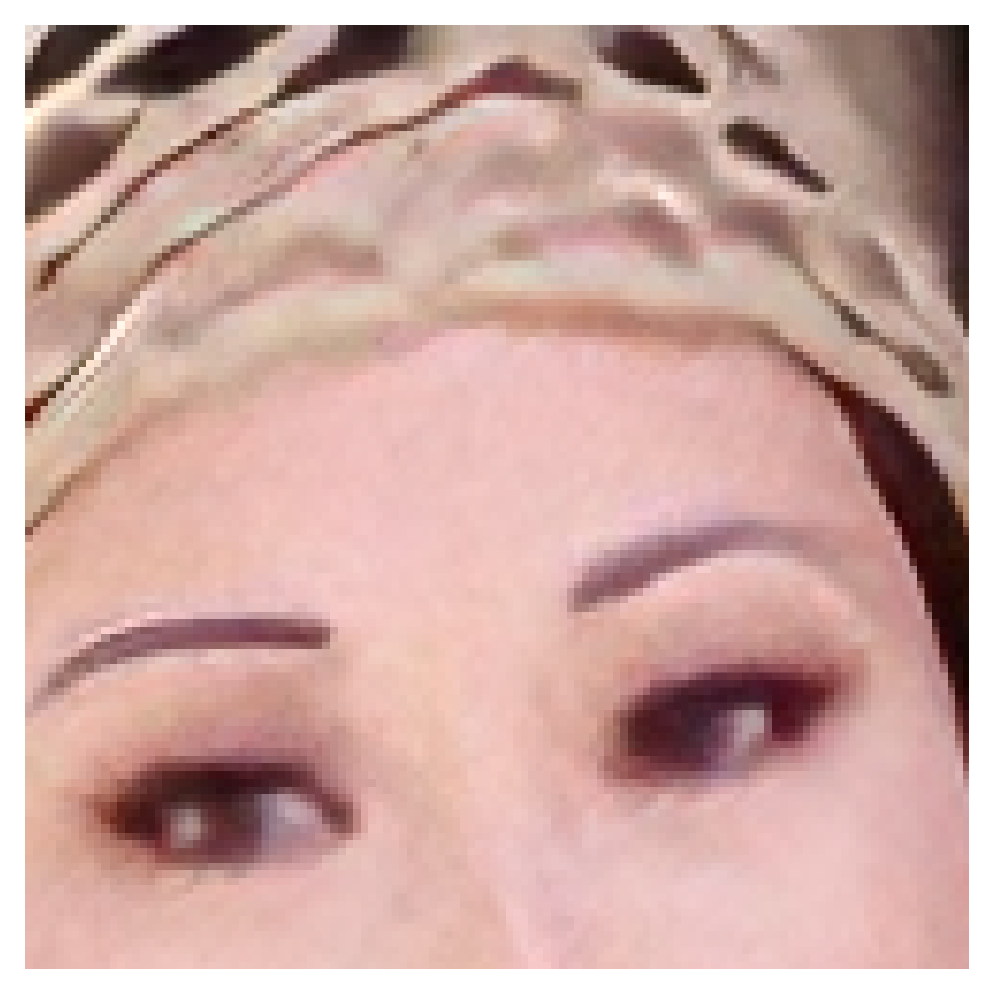}\nsp&  
  \nsp\includegraphics[width=\fsz]{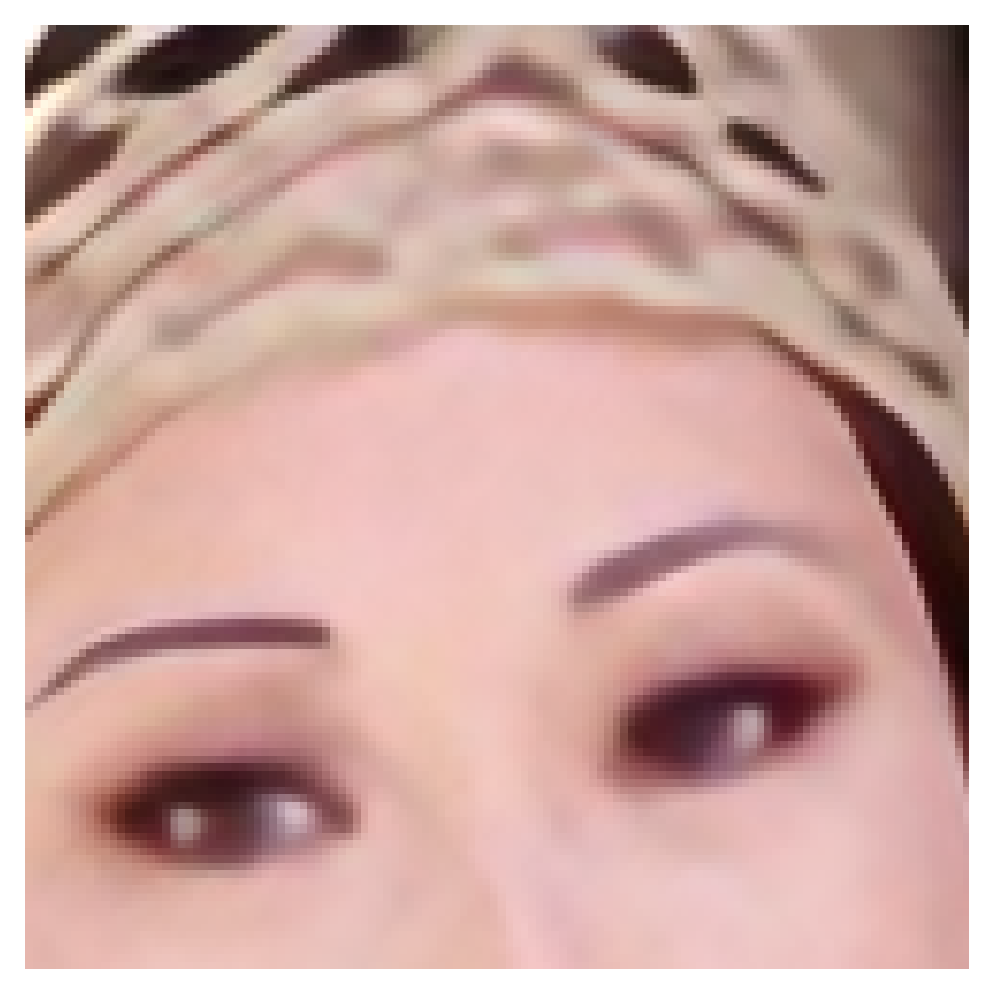}\nsp  
   \\
   \nsp\footnotesize{cropped}\nsp&
   \nsp\footnotesize{low res}\nsp&
   \nsp\footnotesize{DIP}\nsp&
   \nsp\footnotesize{DeepRED}\nsp&
   \nsp\footnotesize{Ours}\nsp&
   \nsp\footnotesize{Ours - avg}\nsp\\   
\end{tabular}
\caption{{Deblurring (spectral super-resolution)}. Images blurred by retaining only $5\%$ of low frequencies.} 
\label{fig:deblur}
\end{figure}  

\begin{table}
  \centering
  \begin{tabular}{llllll}
    \toprule
    \multicolumn{6}{c}{Algorithm}                   \\
    \cmidrule(r){2-6}
    Ratio   & $MM^Tx$ &  DIP     & DeepRED & Ours  & Ours - avg\\
    \midrule
    10\% & 30.2 (0.91)  & 32.54 (0.93) &  32.63 (0.93) &31.82 (0.93)  & {\bf 32.78} ({\bf 0.94})\\
    5\% & 27.77 (0.85) & 29.88 (0.89) & 29.91 (0.89) & 29.22 (0.89) & {\bf 30.07} ({\bf 0.90}) \\
    \bottomrule\\
  \end{tabular}
 \caption{Spectral super-resolution (deblurring) performance averaged over Set5 images. Values indicate YCbCr-PSNR (SSIM).}
  \label{tab:deblur_set5}
 \end{table}

\textbf{Deblurring (spectral super-resolution).}
The applications described above are based on partial measurements in the pixel domain. Here, we consider a blurring operator that operates by retaining a set of low-frequency coefficient in the Fourier domain, discarding the rest. This is equivalent to blurring the image with a sinc kenerl. In this case, $M$ consists of the preserved low-frequency columns of the discrete Fourier transform, and $MM^Tx$ is a blurred version of $x$. Note that, unlike Gaussian deblurring, this problem does not have a trivial solution. Examples are shown in Figure \ref{fig:deblur}.

\begin{table}[t]
  \centering
  \begin{tabular}{llllll}
    \toprule
    \multicolumn{6}{c}{Algorithm}      \\
    \cmidrule(r){2-6}
     Ratio  &  TVAL3  &  ISTA-Net     &  DIP  & BNN  & Ours \\
    \midrule
    25\% & 26.48 (0.77)  & 29.07 (0.84) & 27.78 (0.80) & 28.63 (0.84) & {\bf 29.16} ({\bf 0.88})\\
    10\%  & 22.49 (0.58) & 25.23 (0.69) & 24.82 (0.69)  & 25.24 (0.71) & {\bf 25.47} ({\bf 0.78})\\
    4\%  & 19.10 (0.42) &  22.02 (0.54) &  22.51 (0.58)  & {\bf 22.52} (0.58) & 22.07 ({\bf 0.68})\\
    \bottomrule\\
  \end{tabular}
\caption{Compressive sensing performance averaged over Set68 images \cite{MartinFTM01}. Values indicate PSNR (SSIM).}
  \label{tab:cs_set68}
\end{table}

\begin{figure}[ht]
\centering
\def\fsz{0.23\linewidth} \def\nsp{\hspace*{-.01\linewidth}}
\begin{tabular}{cccc}
  \includegraphics[width=\fsz]{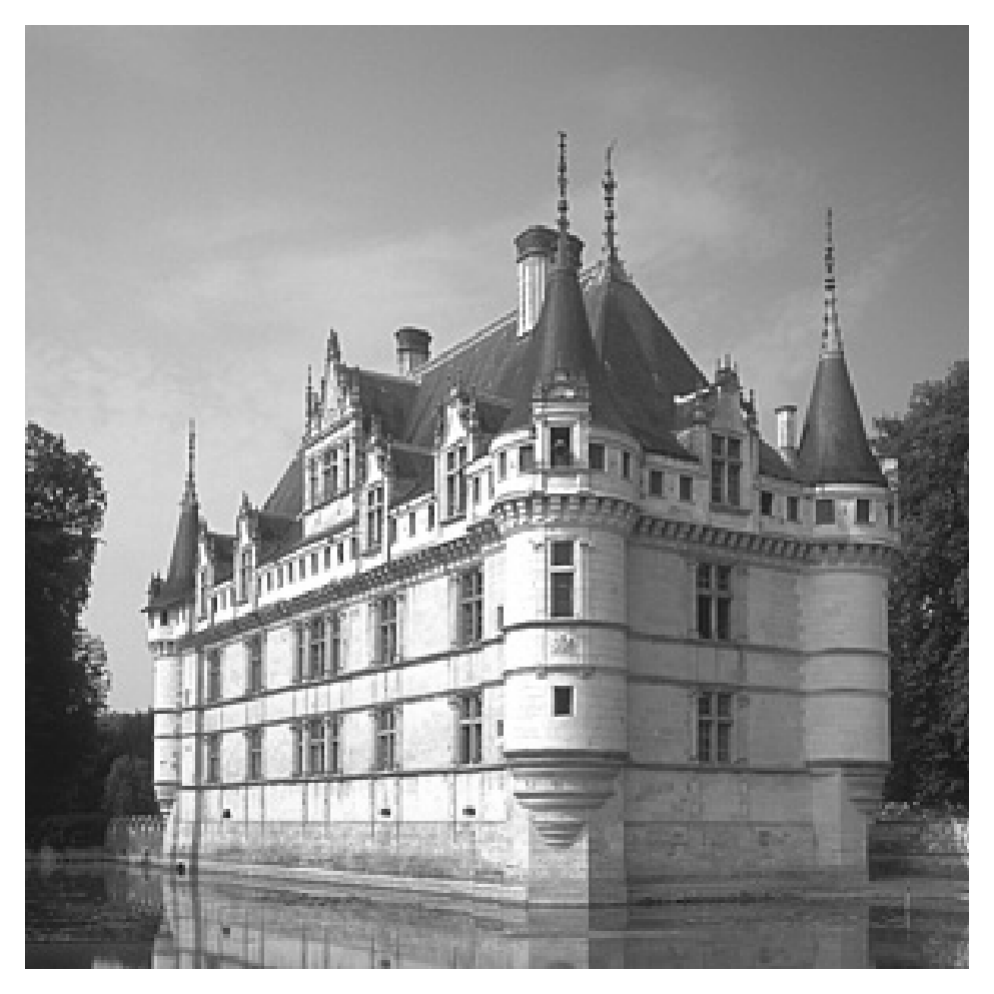}\nsp &
  \includegraphics[width=\fsz]{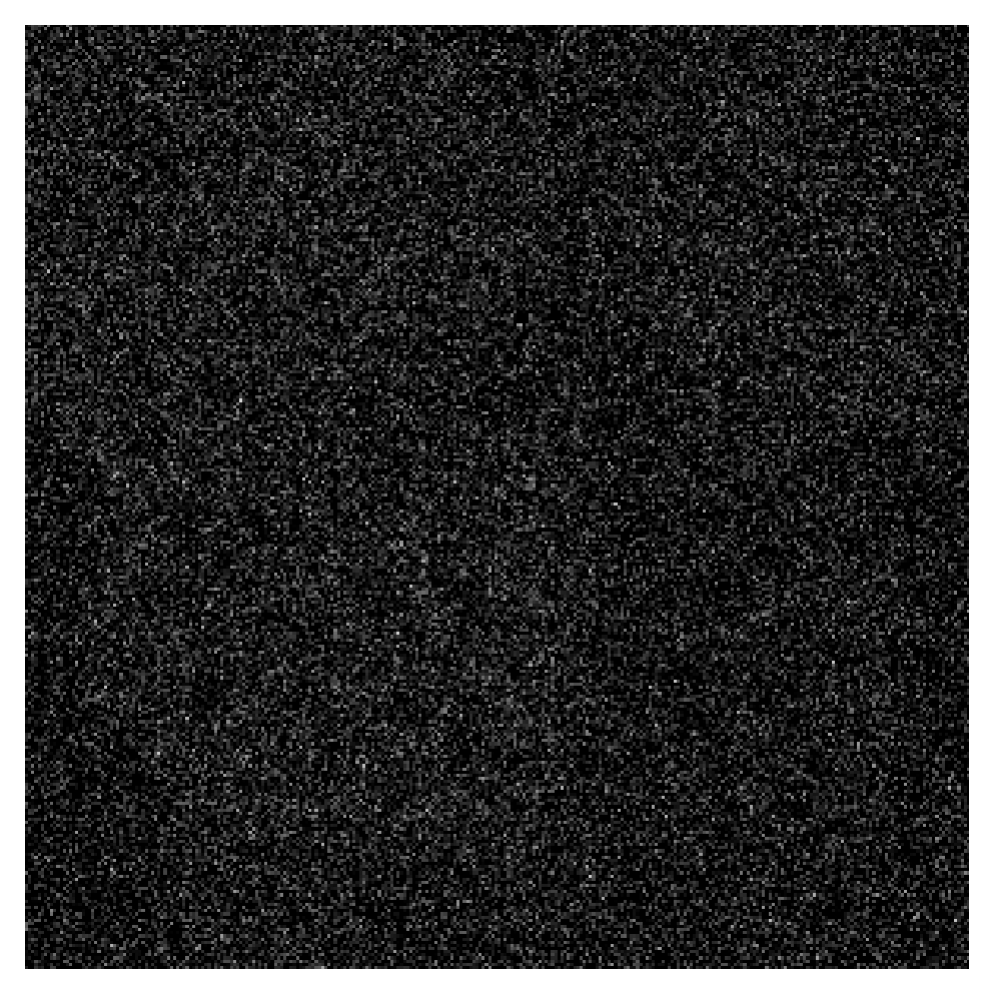}\nsp &
  \includegraphics[width=\fsz]{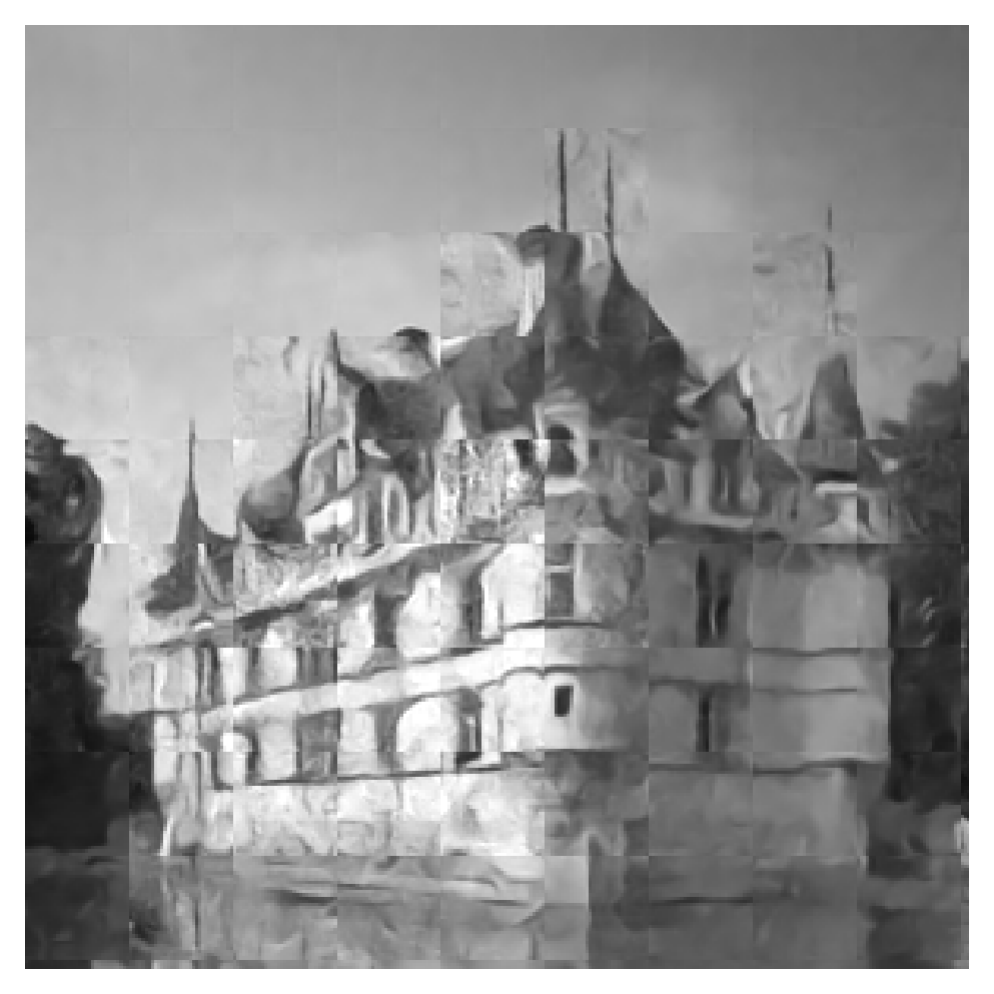}\nsp &
  \includegraphics[width=\fsz]{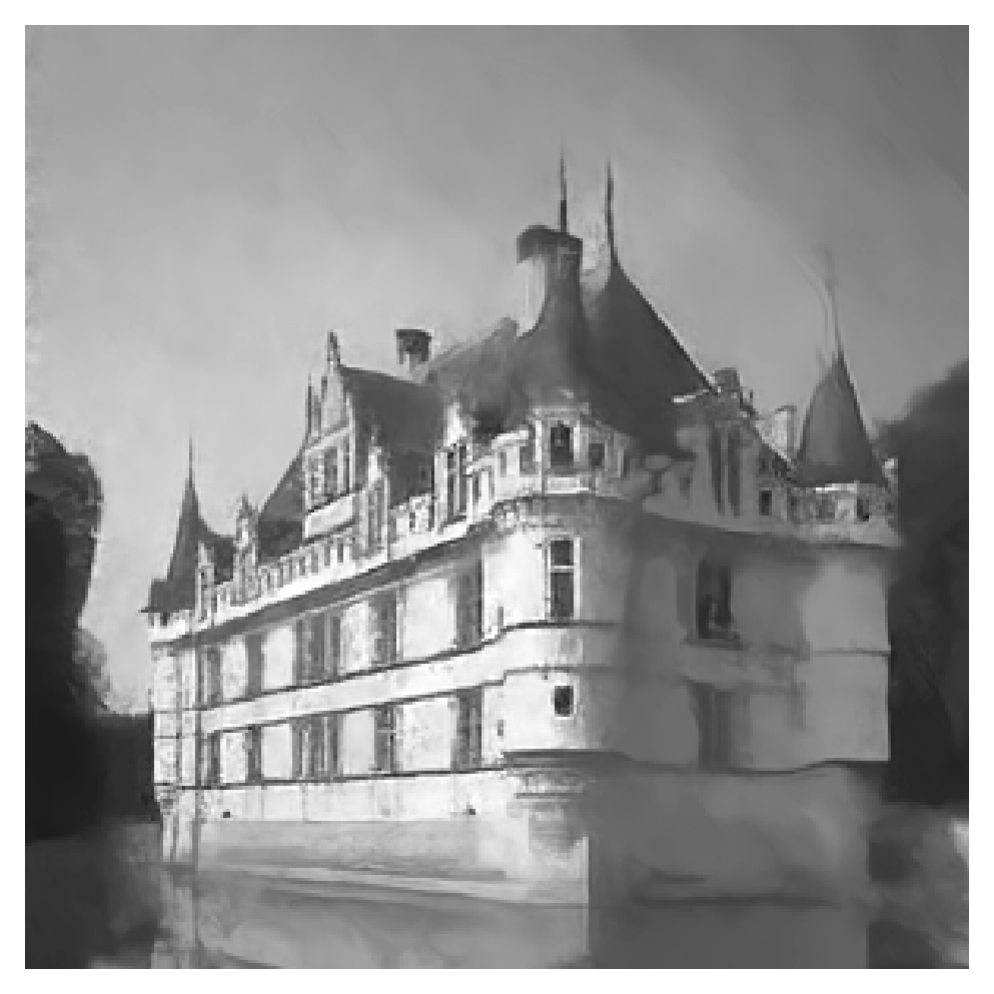}\nsp \\
  \includegraphics[width=\fsz]{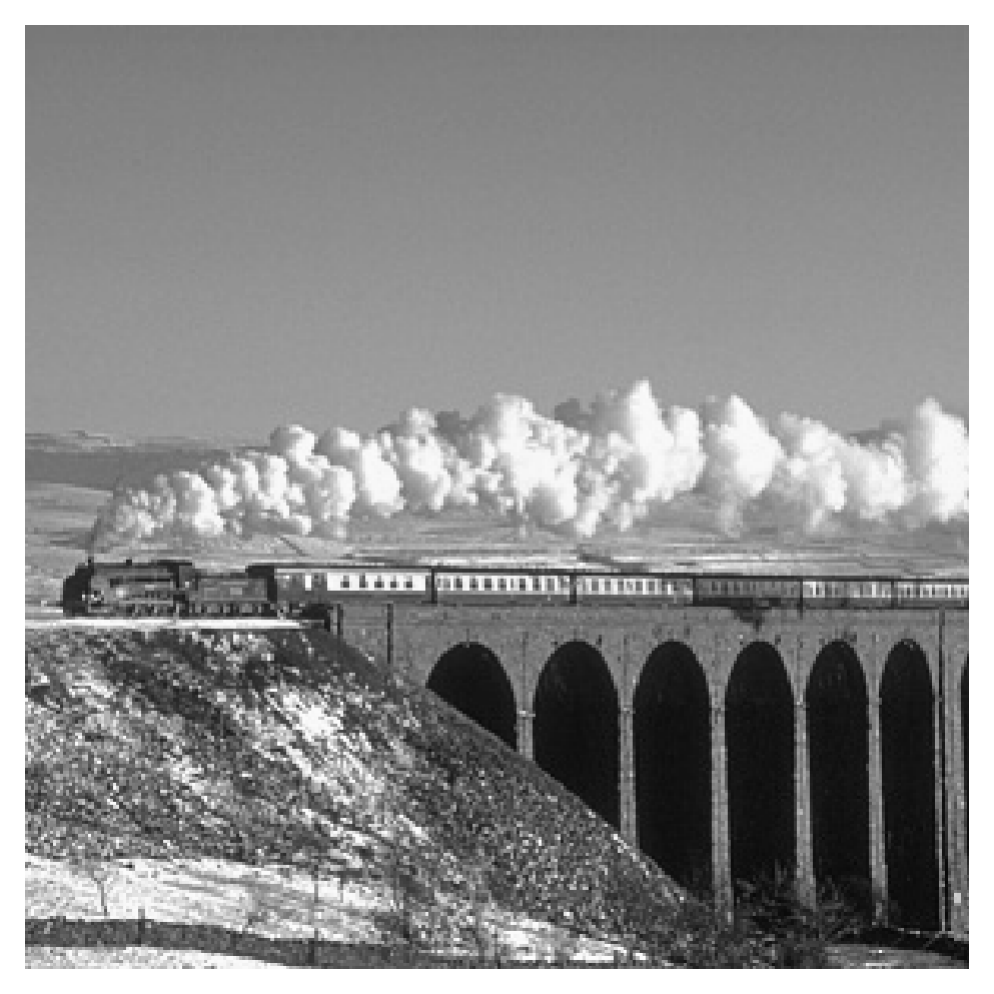}\nsp & 
  \includegraphics[width=\fsz]{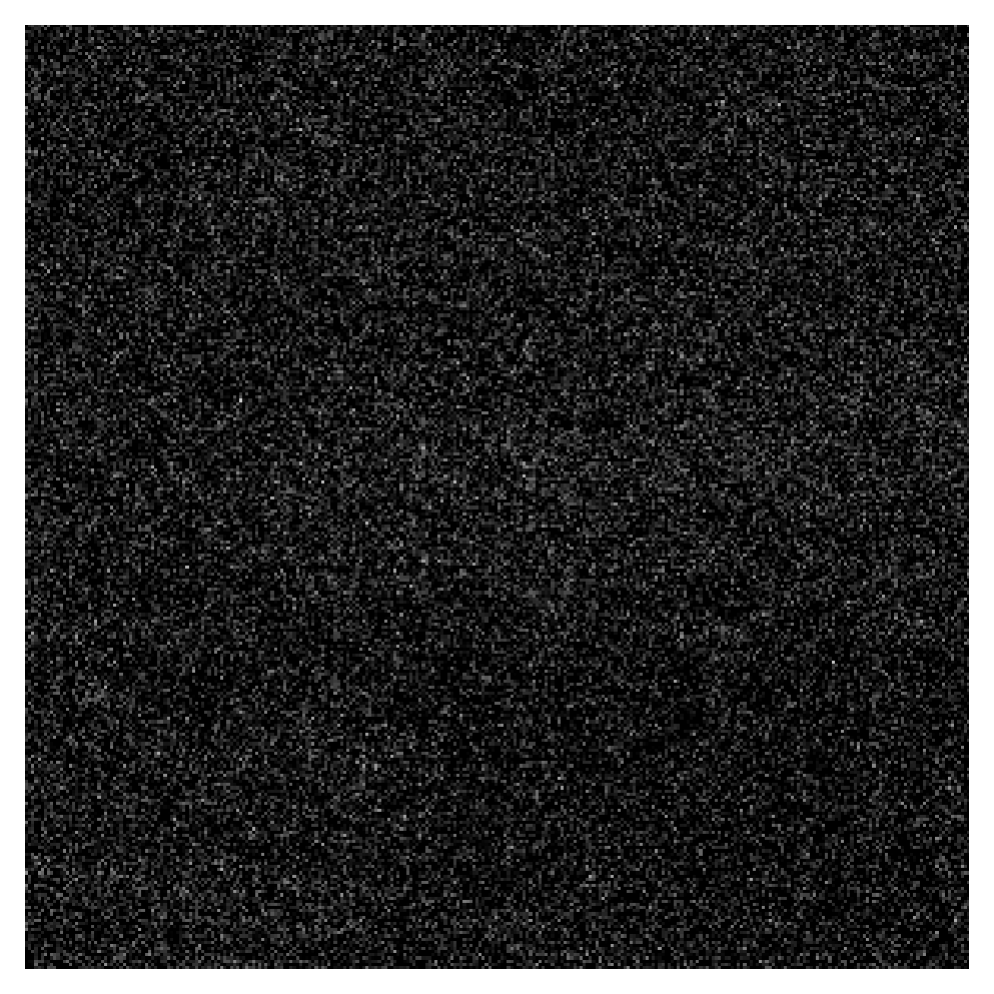}\nsp &
    \includegraphics[width=\fsz]{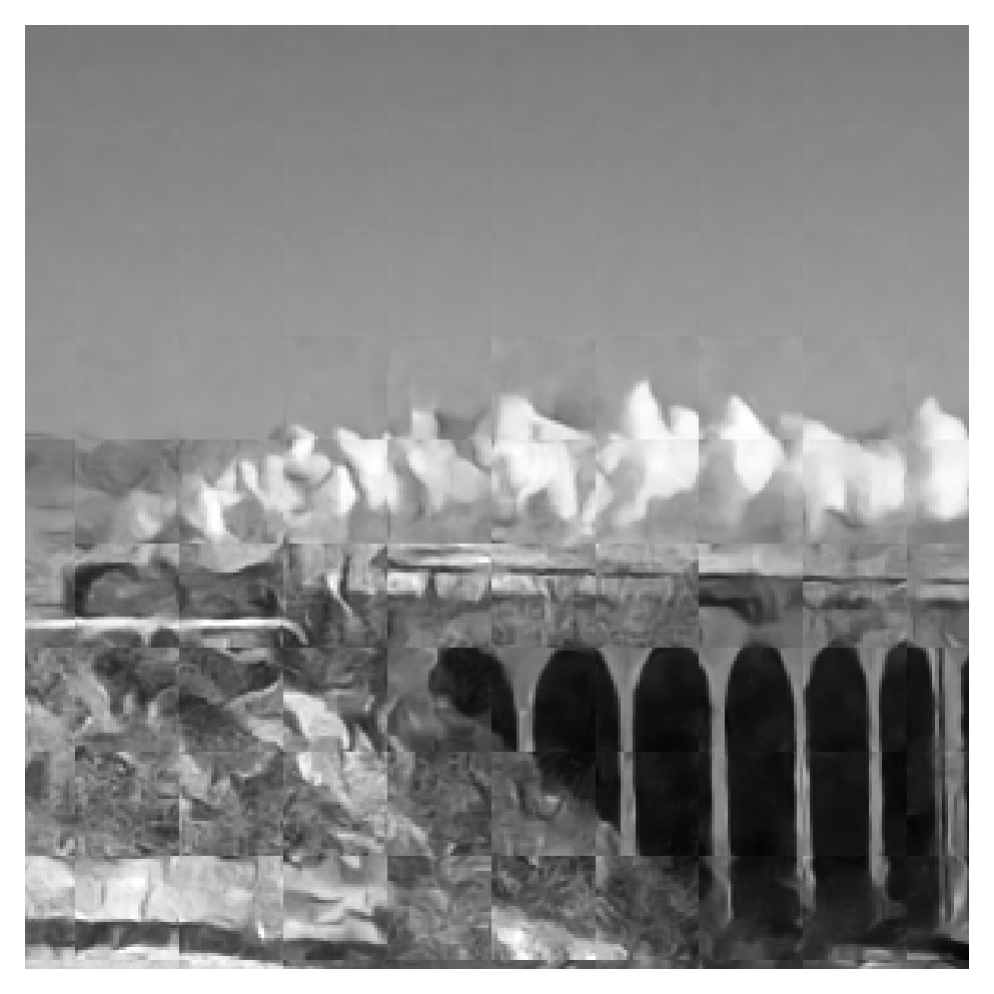}\nsp &
  \includegraphics[width=\fsz]{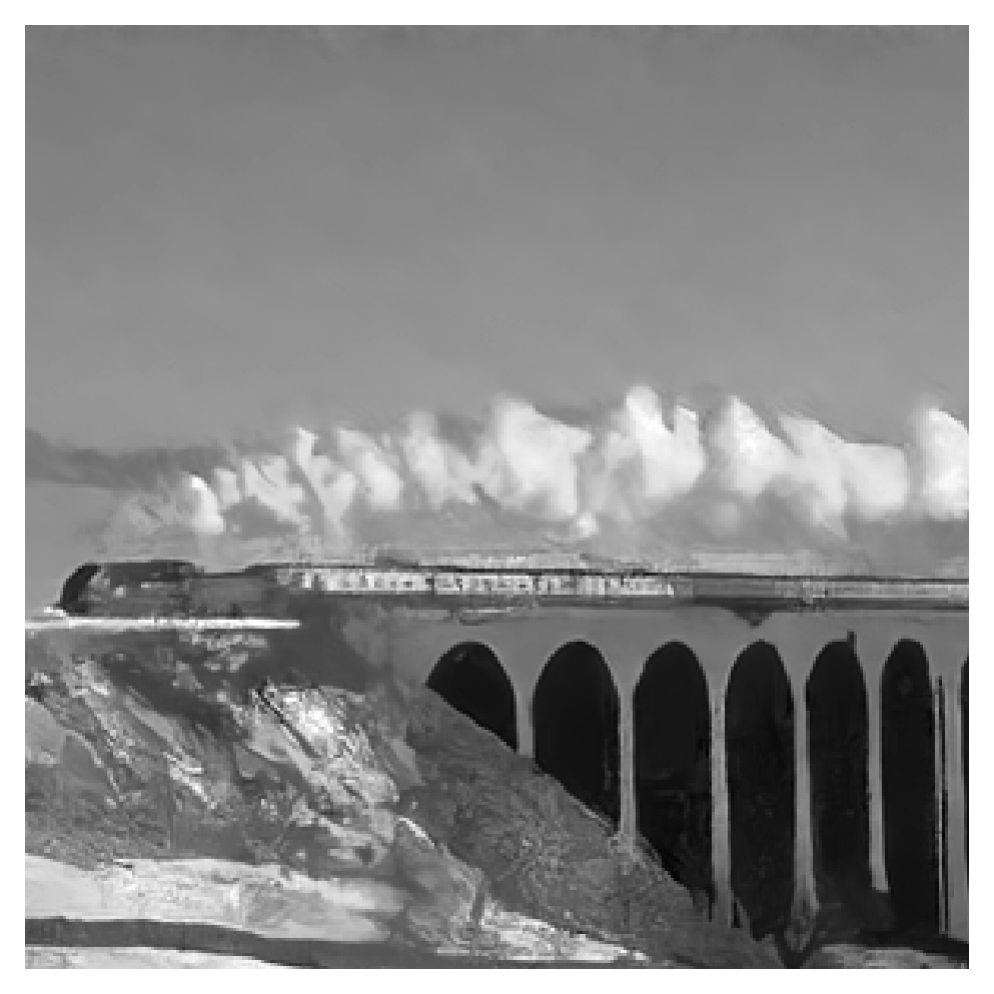}\nsp \\
  \includegraphics[width=\fsz]{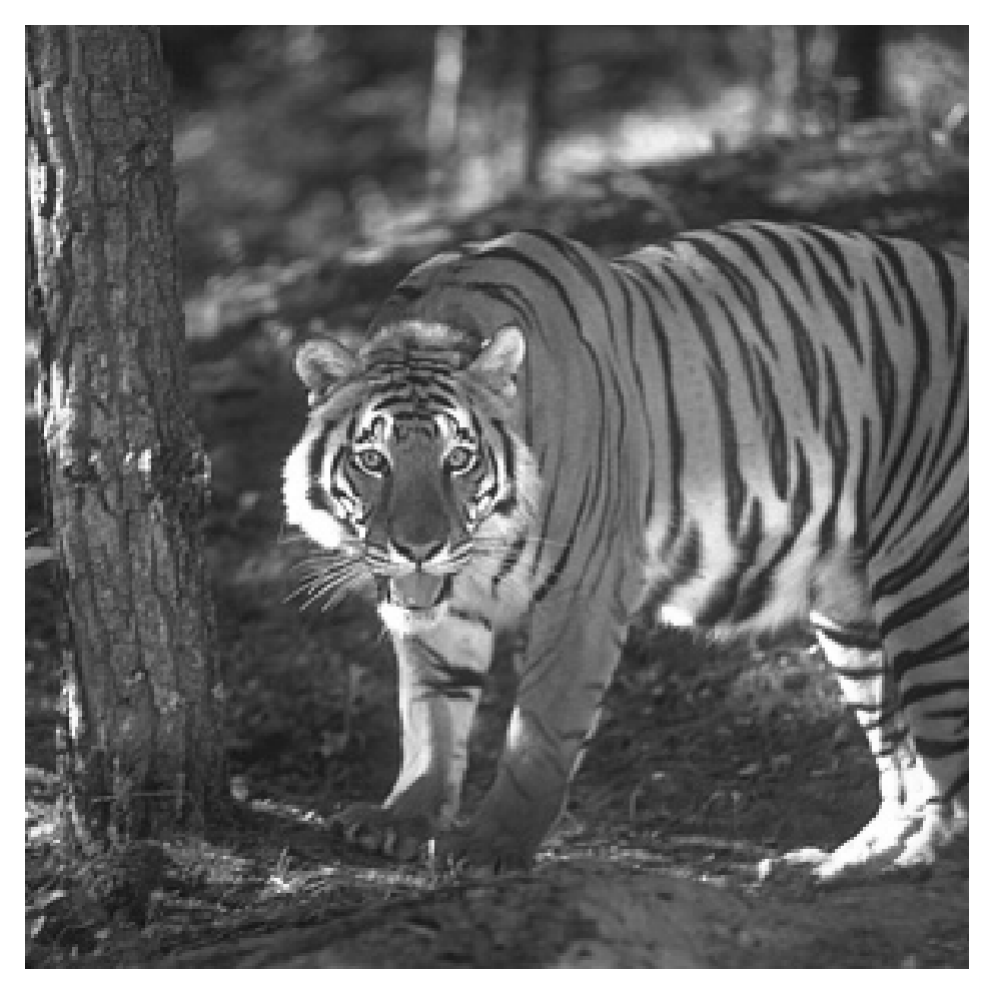}\nsp &
  \includegraphics[width=\fsz]{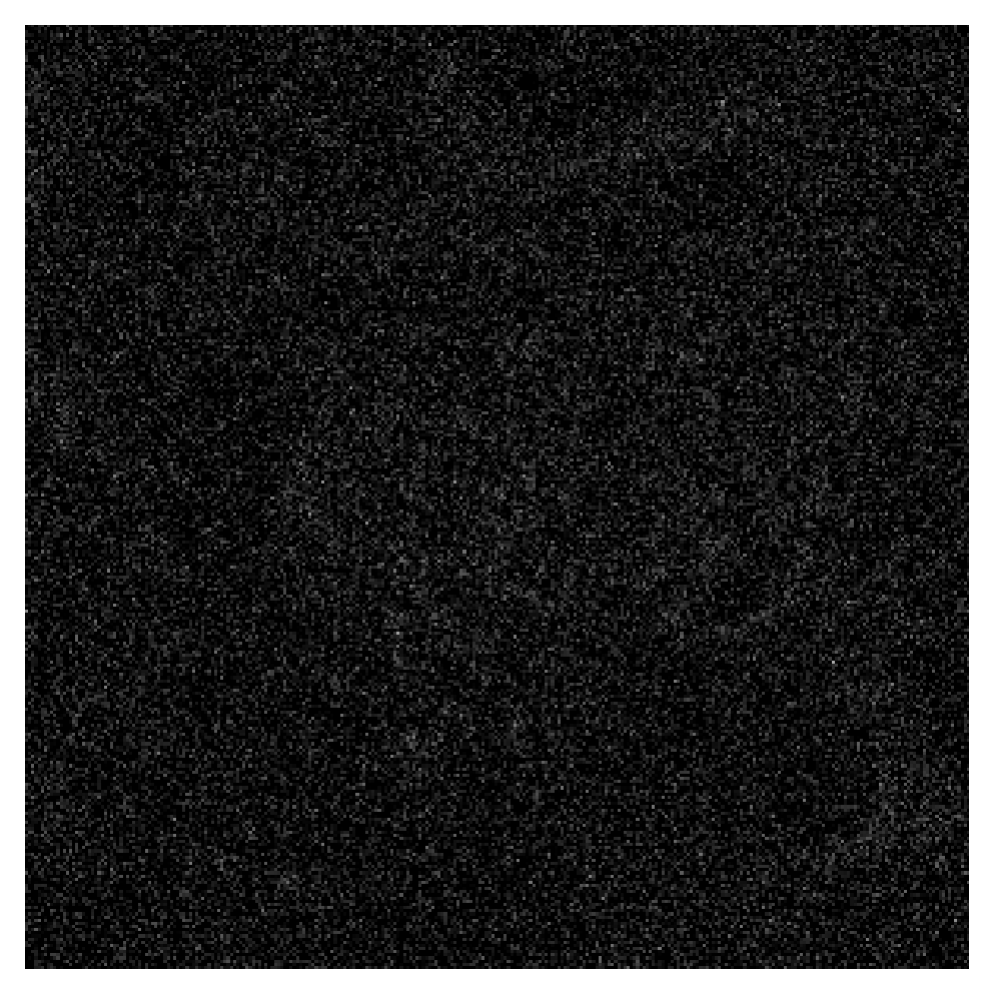}\nsp &
  \includegraphics[width=\fsz]{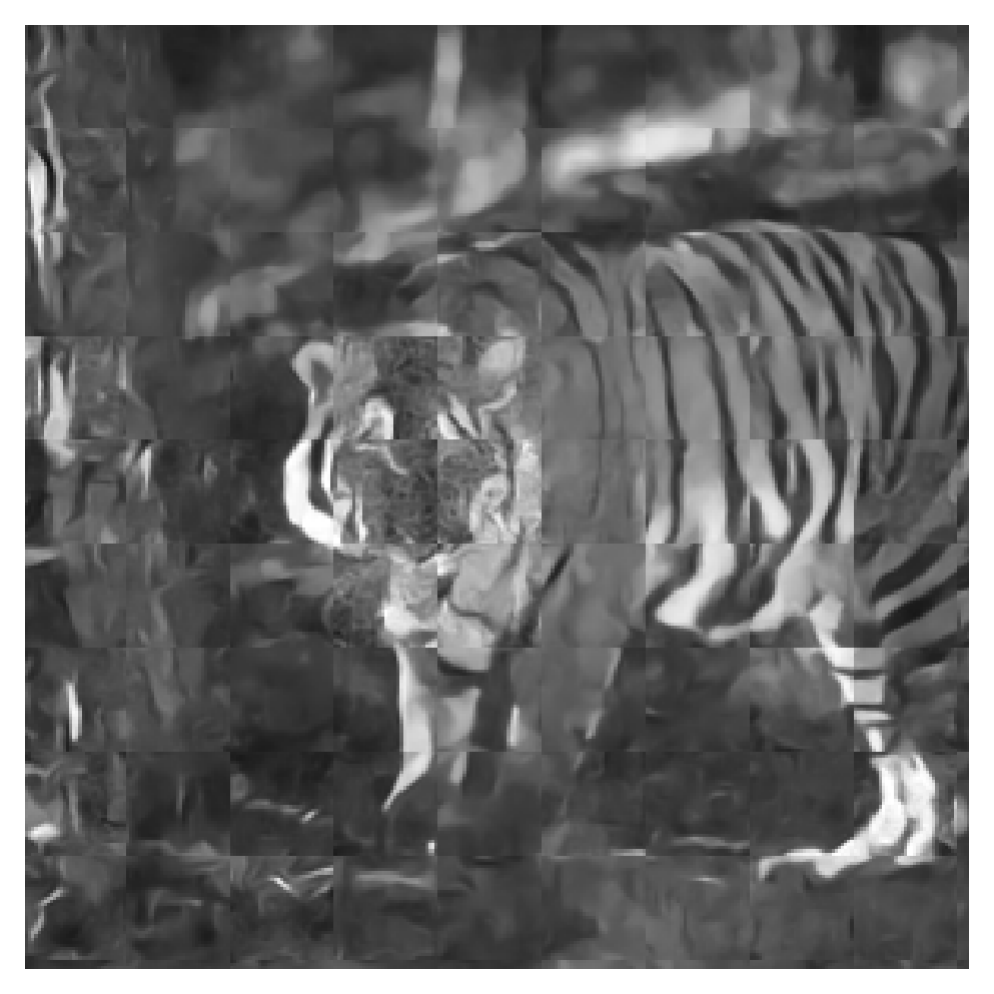}\nsp &
  \includegraphics[width=\fsz]{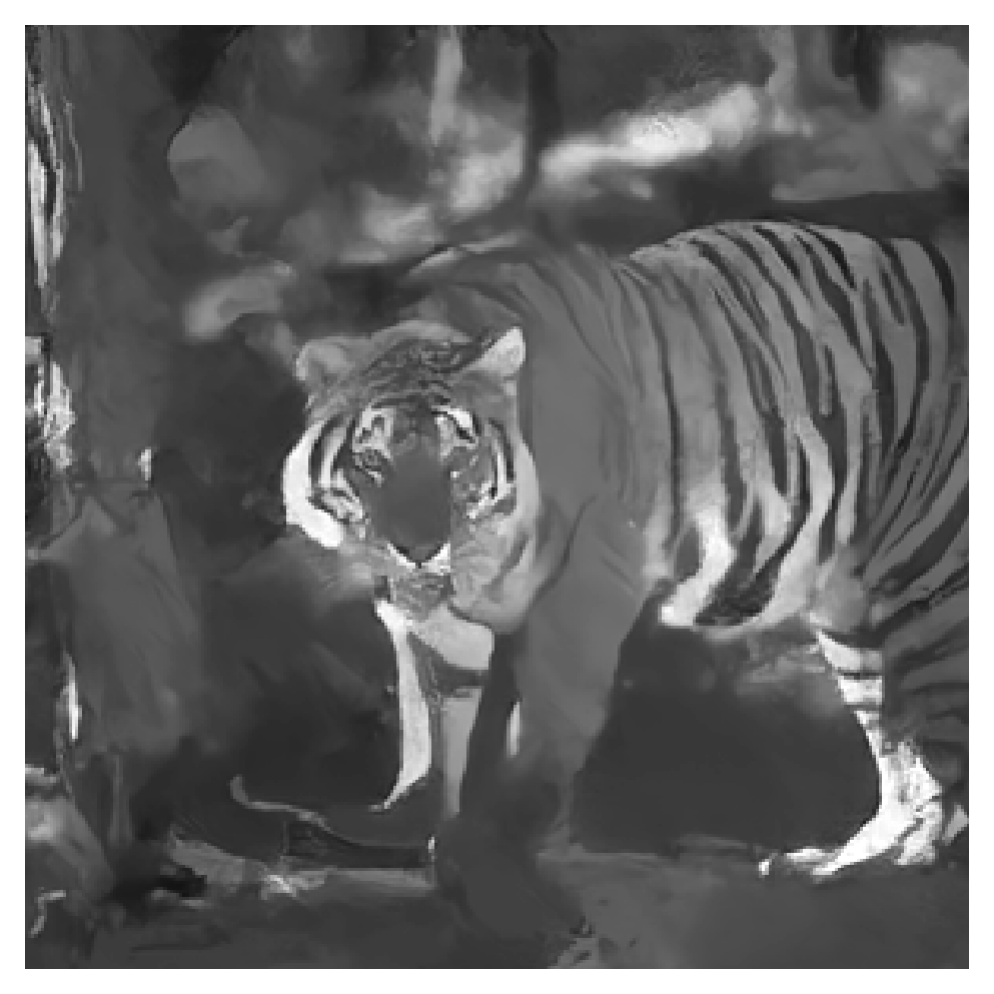}\nsp \\
  \nsp\footnotesize{original}\nsp&
  \nsp\footnotesize{measured}\nsp&
  \nsp\footnotesize{ISTA-Net}\nsp&
  \nsp\footnotesize{Ours}\nsp\\
\end{tabular}
\caption{
{Compressive sensing}. 
Measurement matrix $M$ contains random, orthogonal unit vectors, with dimensionality reduced to $10\%$ of the number of image pixels. First column shows original images ($x$), second column is linear (pseudo-inverse) reconstruction ($MM^Tx$).  Third column: images recovered using ISTA-Net (Supervised and optimized for one random measurement matrix). Fourth column: Our method. 
} 
\label{fig:CS}
\end{figure}  

\textbf{Compressive sensing.} Compressive sensing~\citep{Candes06,Donoho06} provides a set of theoretical results regarding recovery of sparse signals from a small number of linear measurements. Specifically, if one assumes that signals can be represented with at most $k<<N$ non-zero coefficients in a known basis, they can be recovered from a measurements obtained by projecting onto a surprisingly small number of axes (approaching $k \log (N/k)$), far fewer than expected from traditional Shannon-Whitaker sampling theory. The theory relies on the sparsity property, which corresponds to a ``union of subspaces'' prior \citep{Blumensath09}, and on the measurement axes being incoherent (essentially, weakly correlated) with the sparse basis. 
Typically, one chooses a sensing matrix \comment{, $M \in R^{N \times n}$,} containing a set of $n<<N$ random orthogonal axes.
\comment{Specifically, if $X$ is $s$-sparse, then we need the minimum number of measurements to satisfy $n \geq 4s$. Also $\phi$ and $\psi$ should be incoherent. Coherence of two orthogonal matrices is defined as $ \mu(\phi, \psi) = \sqrt{N} \max_{j,k} \phi^T_j \psi_k$, and measures the largest correlation between any two elements of $\phi$
and $\psi$.}  Recovery is achieved by solving the sparse inverse problem, using any of a number of  methods.
\comment{To recover the image, we need to solve an optimization problem, $\min_X ||X||_1$, subject to $M^T \psi X = x^c$, 
Which is a relaxation of  $||X||_0$ minimization. An example of a pair of incoherent $\phi$ and $\psi$ is to choose $\phi$ to be a random orthogonal matrix, and to set $\psi$ to Fourier or wavelet transforms. }

Photographic images are not truly sparse in any fixed linear basis, but they can be reasonably approximated by low-dimensional subsets of Fourier or wavelet basis functions, and compressive sensing results are typically demonstrated using one of these. The manifold prior embedded within our CNN denoiser corresponds to a nonlinear form of sparsity, and analogous to sparse inverse algorithms used in compressed sensing, 
our stochastic coarse-to-fine ascent algorithm can be used to recover an image from a set of linear projections onto a set of random basis functions. Table \ref{tab:cs_set68} shows average numerical performance on Set68 \cite{MartinFTM01}, in terms of both PSNR and SSIM. TVAL3 \citep{li2013efficient} is an optimization-based algorithm with total variation regularization, ISTA-Net \cite{zhang2018ista} is a CNN supervised method trained to reconstruct images from measurements obtained from one particular random matrix. DIP \cite{Ulyanov2020DeepImagePrior} is an unsupervised general linear inverse solver, BNN \cite{pang2020self} is an unsupervised Bayesian method for solving compressive sensing problems. Our method outperforms almost all of the other methods. All the numbers are taken from \cite{pang2020self} except for ISTA-Net which were obtained by running their open-source code. Figure \ref{fig:CS} shows three examples of images recovered from random projections using our denoiser-induced manifold prior, compared with. In all cases, the denoiser-recovered images exhibit fewer artifacts.

\begin{table}
  \caption{Run time, in seconds,  and (number of iterations until convergence) for different applications, averaged over images in Set12, on an NVIDIA  RTX 8000 GPU.}
  \label{tab:runtime}
  \centering
  \begin{tabular}{lllll}
    \toprule
    Inpainting $30\times 30$  & Missing pixels 10\% &  SISR  4:1  &  Deblurring 10\% &  CS 10\%\\
    \midrule
           15.7 (647) & 7.8 (313)&  7.6 (301) &  8.4 (332) &  32.5 (317) \\
    \bottomrule
  \end{tabular}
\end{table}

\section{Related work}
\label{sec:related work}
Our method is conceptually similar to ``plug and play''\cite{venkatakrishnan2013plug}, \comment{briefly described in Section \ref{sec:intro},} in using a denoiser to solve  linear inverse problems. But it differs in a number of important ways: 
(1) 
We derive our method from Miyasawa's explicit relationship between the mapping and the prior, which is exact and direct, and makes the algorithm interpretable. The rationale for using a denoiser as a regularizer in P\&P framework arises from interpreting the proximal operator of the regularizer as a MAP solution of a denoising problem. RED, which is based on P\&P, offers some intuitions as to why the smoothness regularization is a proper choice, but the connection to the prior is is still not explicit; 
(2) 
We sample a high probability image from the implicit prior that is consistent with a linear constraint. This stochastic solution does not minimize MSE, but has high perceptual quality. RED and other P\&P methods are derived as MAP solutions, and although this is not equivalent to minimizing MSE (maximizing PSNR), the results generally have better PSNR than our sampling results, but are visually more blurred (see Figure \ref{fig:SR4} and \ref{fig:deblur}). As shown in Fig. \ref{fig:deblur} and Table \ref{tab:deblur_set5}, averaging over samples from our method results in improved PSNR (but more blurring);
(3) RED, which uses ADMM for optimization, relies on hyper-parameter selection that affects convergence (as discussed extensively in \cite{Romano17}). On the contrary, our algorithm adjusts step-size automatically using the denoiser (which must be ``blind'', operating on images of unknown noise level), with only two primary hyper-parameters ($h_0$ and $\beta$), and performance is robust to choices of these (Appendix D);
(4) Our method requires that the denoiser be trained on Gaussian-noise contaminated images to minimize squared error and must operate "blind" (without knowledge of noise level), whereas RED relies on a set of three additional properties. 

Our method is also similar to recent work that uses Score Matching \citep{hyvarinen2005estimation} to draw samples from an implicit prior. This line of work is rooted in the connection between denoising autoencoders and score matching which was described in \cite{Vincent11}. The idea that denoising autoencoders learn the gradient of log density, and its connection to an underlying data manifold has been explored in \citep{alain2014regularized, Bengio13, vincent2008extracting}. More recently, \citep{saremi2019neural, Saremi18} directly train a neural network to estimate energy (a scalar value related to density) and equated its gradient to the score of the data. Finally, in a breakthrough paper, Song and Ermon \citep{Song19} trained a sequence of denoisers with decreasing levels of noise and used Langevin dynamics to sample from the underlying probability distributions in succession, using the sample from each stage to initialize the process at the next stage. Our work differs in several important ways: (1) Our derivation of the relationship between the denoiser mapping and the density is direct and significantly simpler (i.e. a few lines of proof in Section \ref{Miyasawa}), exploiting a result from the classical statistical literature on Empirical Bayes estimation~\cite{Miyasawa61}. The theoretical connection between score matching and denoinsing autoencoders \cite{Vincent11} is approximate and significantly more complex. More importantly, our method makes explicit the noise level, whereas the score matching and denoising score matching derivations hold only in the limit of small noise. 
(2) Our method assumes and uses a single (universal) blind denoiser, rather than a family of denoisers trained for a pre-specified set of noise levels. The noise level schedule, hence the step size, is automatically adjusted by the denoiser, based on distance to the manifold. 
(3) Our algorithm is efficient - we use stochastic gradient ascent to maximize probability, rather than MCMC methods (e.g., Langevin dynamics) to draw proper samples from each of a discrete sequence of densities.
(4) Finally, we've demonstrate the generality of our algorithm by using it as a general solver for five different linear inverse problems, whereas the Score Matching methods are generally focused on unconditional image synthesis.

\section{Discussion}
We've described a framework for using the prior embedded in a denoiser to solve inverse problems. Specifically, we developed a stochastic coarse-to-fine gradient ascent algorithm that uses the denoiser to draw high-probability samples from its implicit prior, and a constrained variant that can be used to solve {\em any} linear inverse problem. The derivation relies on the denoiser being optimized for mean squared error in removing additive Gaussian noise of unspecified amplitude. Denoisers can be trained using discriminative learning (nonlinear regression) on virtually unlimited amounts of unlabeled data, and thus, our method extends the power of supervised learning to a much broader set of problems, without further training.

As mentioned previously, the performance of our method is not well-captured by comparisons to the original image (using, for example, PSNR or SSIM). Performance should ultimately be tested using experiments with human observers, which might be approximated using using a no-reference perceptual quality metric (e.g., \cite{ma2018blind}). Handling of nonlinear inverse problems with convex measurements (e.g. recovery from quantized representation, such as JPEG) is a natural extension of the method, in which the algorithm would need to be modified to incorporate projection onto convex sets.
Finally, our method for image generation offers a means of visualizing the implicit prior of a denoiser, which arises from the combination of architecture, optimization, regularization, and training set. As such, it might offer a means of experimentally isolating and elucidating the effects of these components.

{
 \bibliographystyle{unsrt}
 \bibliography{universalInverse}
}
\newpage
\appendix

\input{supplementary_material.tex}
\end{document}

%% file: supplementary_material.tex
\section{Description of BF-CNN denoiser}
\label{sec:bfcnn}
Examples in this paper were computed using the publicly-available implementation of the BF-CNN denoiser~\citep{MohanKadkhodaie19b}, which is trained to minimize MSE for images corrupted with additive Gaussian white noise.

\paragraph{Architecture.} The network contains $20$ bias-free convolutional layers, each consisting of $3 \times 3$ filters and $64$ channels, batch normalization, and a ReLU nonlinearity. Note that to construct a bias-free network, we remove all sources of additive bias, including the mean parameter of the batch-normalization, in every layer.

\paragraph{Training Scheme.} We follow the  training procedure described in \citep{MohanKadkhodaie19b}. The network is trained to denoise images corrupted by i.i.d. Gaussian noise with standard deviations drawn from the range $[0, 0.4]$ (relative to image intensity range $[0,1]$). The training set consists of overlapping patches of size $40 \times 40$ cropped from the Berkeley Segmentation Dataset~\citep{MartinFTM01}. Each original natural image is of size $180 \times 180$. Training is carried out on batches of size 128, for 70 epochs. 

\section{Block diagram of Universal Linear Inverse Sampler}
\begin{figure}[H]
    \centering
    \includegraphics[width=0.63\linewidth]{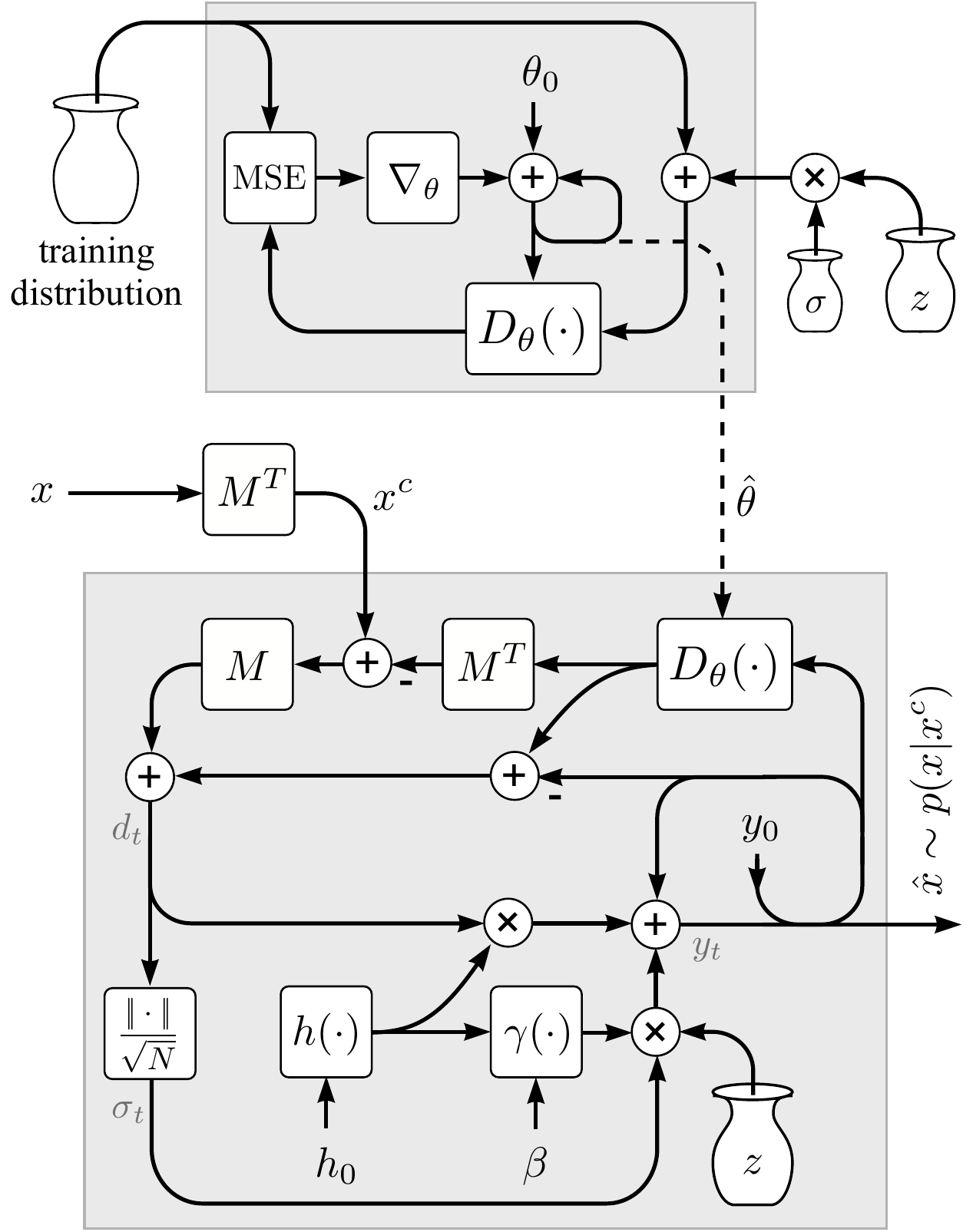}
    \caption{{Block diagrams for denoiser training, and Universal Inverse Sampler.}  {\bf Top:} A parametric blind denoiser,  $D_\theta(\cdot)$, is trained to minimize mean squared error when removing additive Gaussian white noise ($z$) of varying amplitude ($\sigma$) from images drawn from a training distribution.  The trained denoiser parameters, $\hat{\theta}$, constitute an implicit model of this distribution.  {\bf Bottom:} The trained denoiser is embedded within an iterative computation to draw samples from this distribution, starting from initial image $y_0$, and conditioned on a low-dimensional linear measurement of a test image: $\hat{x} \sim p(x|x^c)$, where $x^c = M^T x$. If measurement matrix $M$ is empty, the algorithm draws a sample from the unconstrained distribution.  Parameter $h_0 \in [0,1]$ controls the step size, and $\beta \in [0,1]$ controls the stochasticity (or lack thereof) of the process. }
    \label{fig:BlockDiagram}
\end{figure}

\section{Visualization of Universal Inverse Sampler on a 2D manifold prior}

\begin{figure}[H]
\centering
\def\fsz{0.24\linewidth}
\includegraphics[width=\fsz]{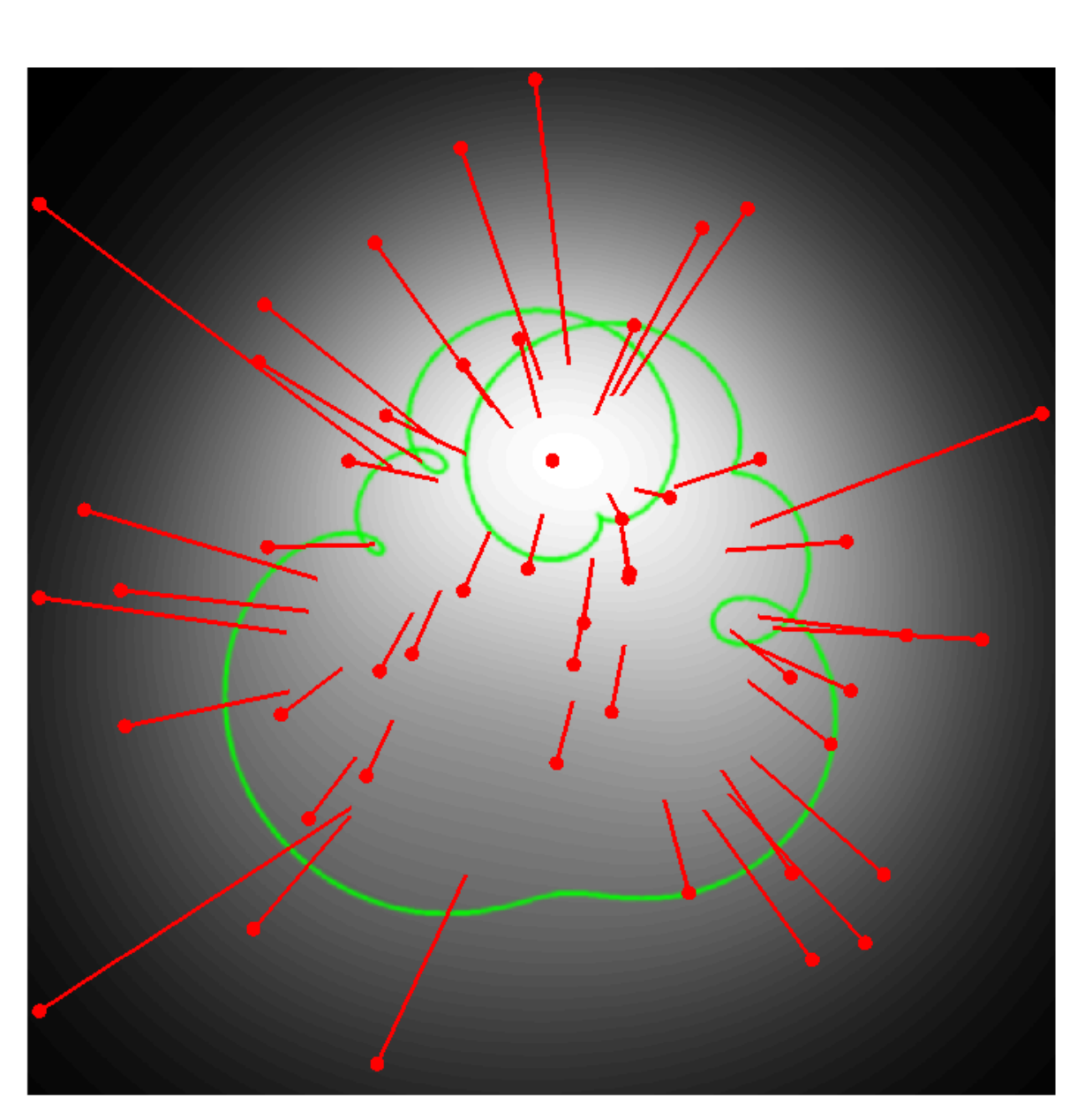} 
\includegraphics[width=\fsz]{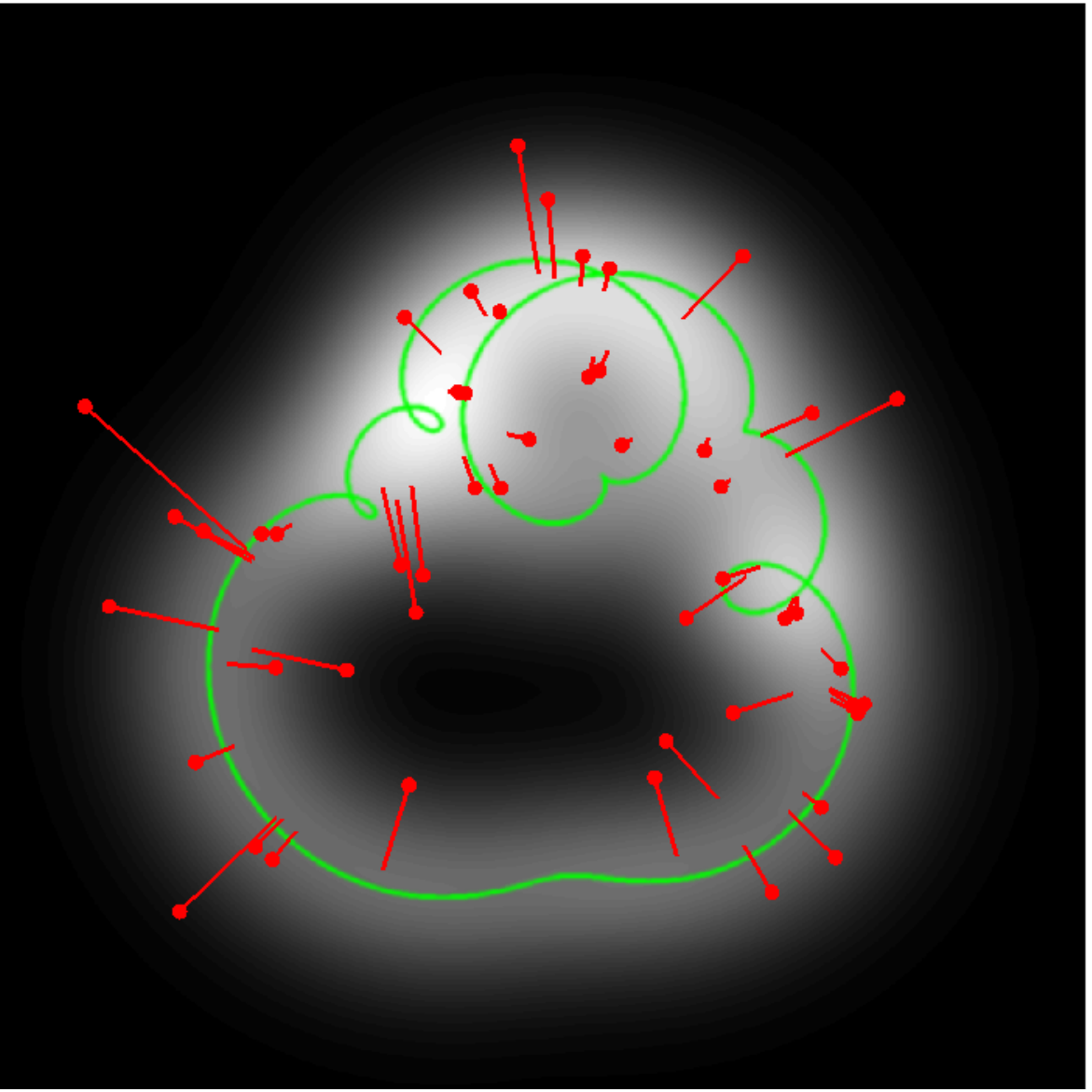} 
\includegraphics[width=\fsz]{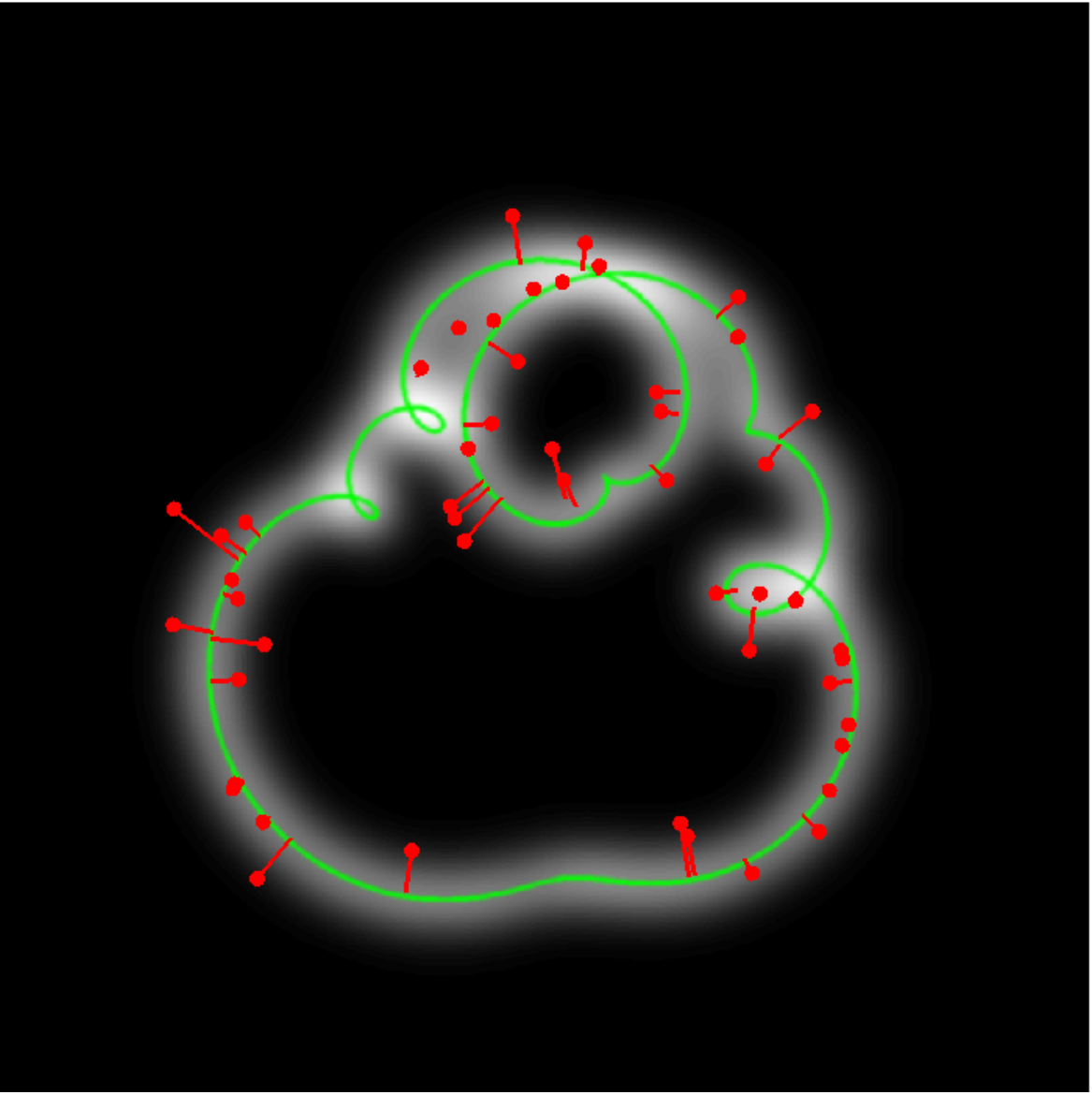} 
\hfill
\includegraphics[width=\fsz]{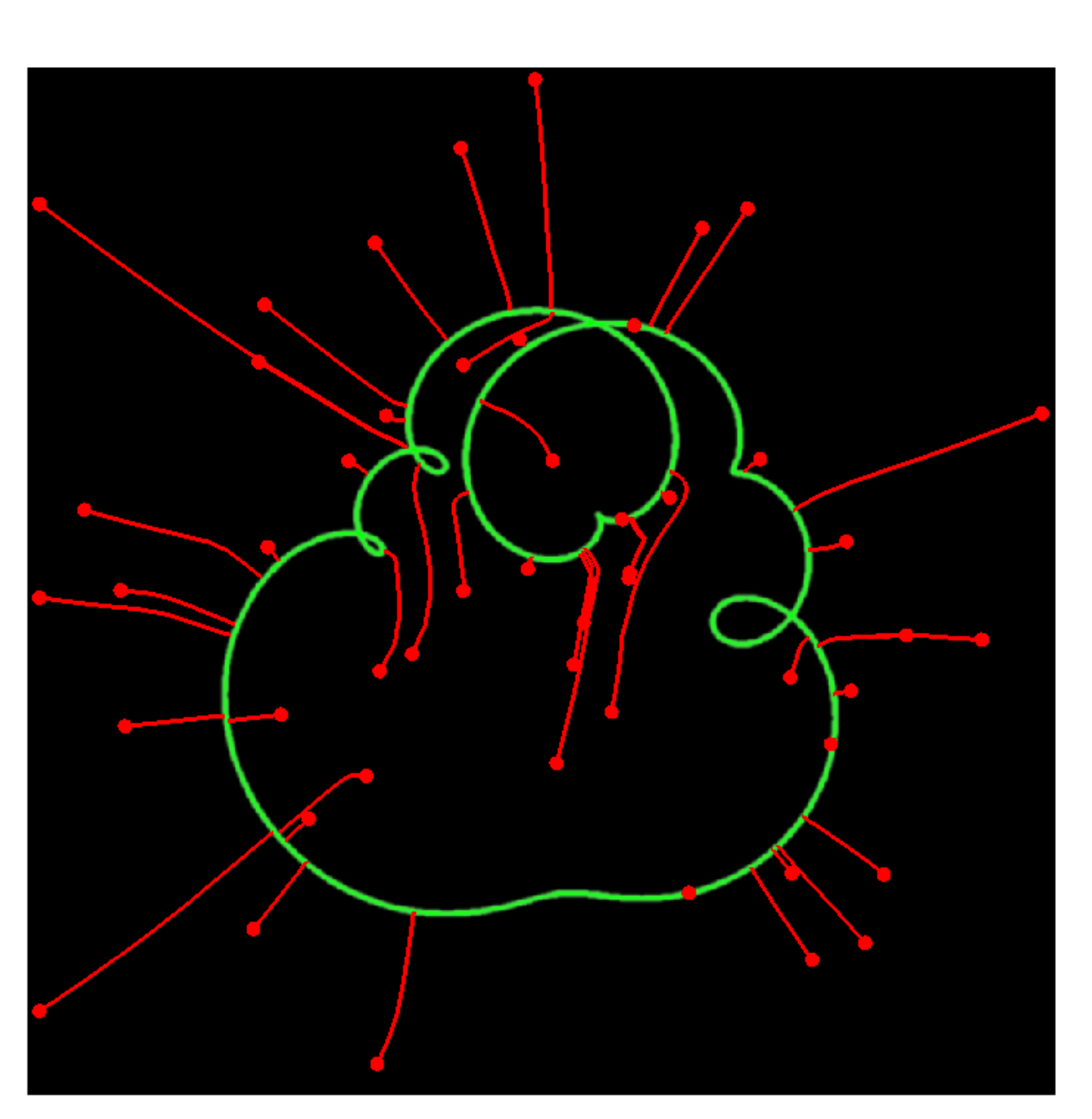} \\[-0.5ex]
\caption{Two-dimensional simulation/visualization of the Universal Inverse Sampler.
Fifty example signals $x$ are sampled from a uniform prior on a manifold (green curve).  
First three panels show, for three different levels of noise, the noise-corrupted measurements of the signals (red points), the associated noisy signal distribution $p(y)$ (indicated with underlying grayscale intensities), and the least-squares optimal denoising solution $\hat{x}(y)$ for each (end of red line segments), as defined by Eq.~(\ref{eq:BLS}), or equivalently, Eq.~(\ref{eq:miyasawa}).
Right panel shows trajectory of our iterative coarse-to-fine inverse algorithm (Algorithm 2, depicted in Figure \ref{fig:BlockDiagram}), starting from the same initial values $y$ (red points) of the first panel. Algorithm parameters were $h_0=0.05$ and $\beta = 1$ (i.e., no injected noise).
Note that, unlike the least-squares solutions, the iterative trajectories are curved, and always arrive at solutions on the signal manifold. }
\label{fig:2Dvisualization}
\end{figure}

\section{Convergence}
\begin{figure}[H]
\centering
\includegraphics[width=0.9\linewidth]{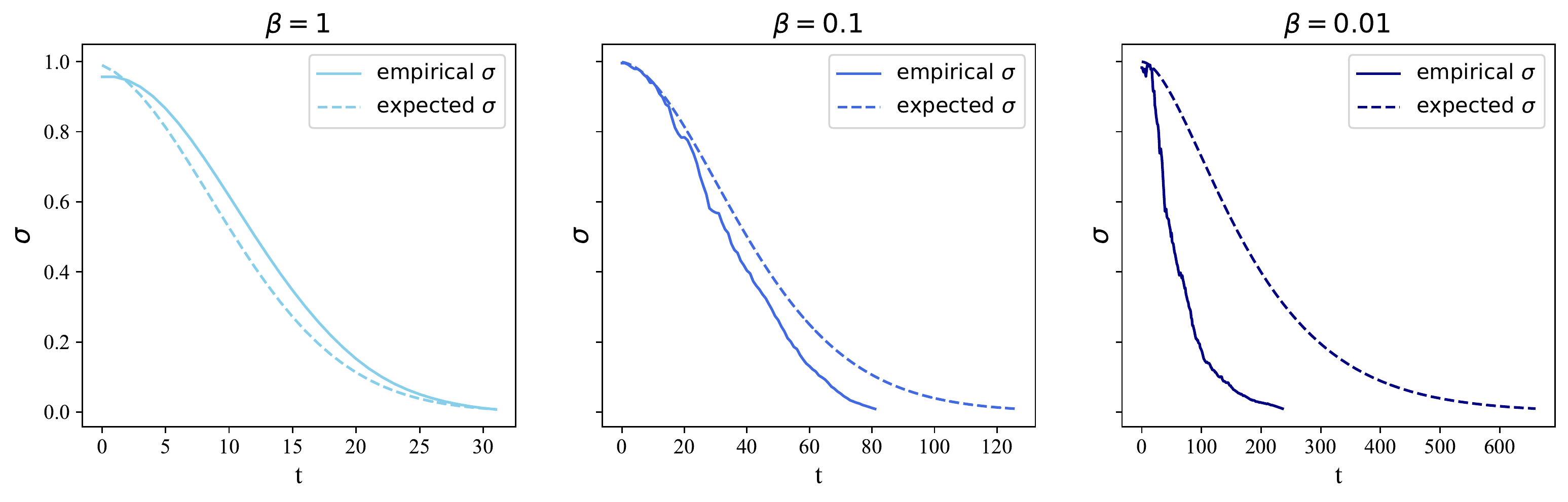} \\[-0.5ex]
\caption{Comparison of the computed effective noise, $\sigma_t = \frac{||d_t||}{\sqrt{N}}$, and the noise expected from the schedule $ \sigma_t= (1 - \beta h_t) \sigma_{t-1}$, where $h_t = h_0 \frac{t}{1+ h_0(t-1) } $. When $\beta = 1$, the effective noise estimated by the denoiser falls at approximately the rate expected by the schedule. As $\beta$ decreases (i.e. for non-zero injected noise), convergence is slower, but faster than the expected rate.  }
\label{fig:convergence_expected}
\end{figure}

Figure \ref{fig:convergence_expected} illustrates the convergence of our iterative sampling algorithm, expressed in terms of the effective noise standard deviation $\sigma=\frac{||d_t||}{\sqrt{N}}$ averaged over synthesis of three images, for three different levels of the stochasticity parameter $\beta$.  Convergence is well-behaved and efficient in all cases.  
As expected, with smaller $\beta$ (larger amounts of injected noise), effective standard deviation falls more slowly, and convergence takes longer. 
For $\beta=1$ (no injected noise), the convergence is approximately what is expected from the formulation of the alorithm (Eq.~\ref{eq:sigma_schedule}).  For larger amounts of injected noise, the algorithm converges faster than expected, we believe because a portion of the additive noise is parallel to the manifold, so does not contribute to calculated variance.

\comment{
\begin{figure}
\centering
\includegraphics[width=0.6\linewidth]{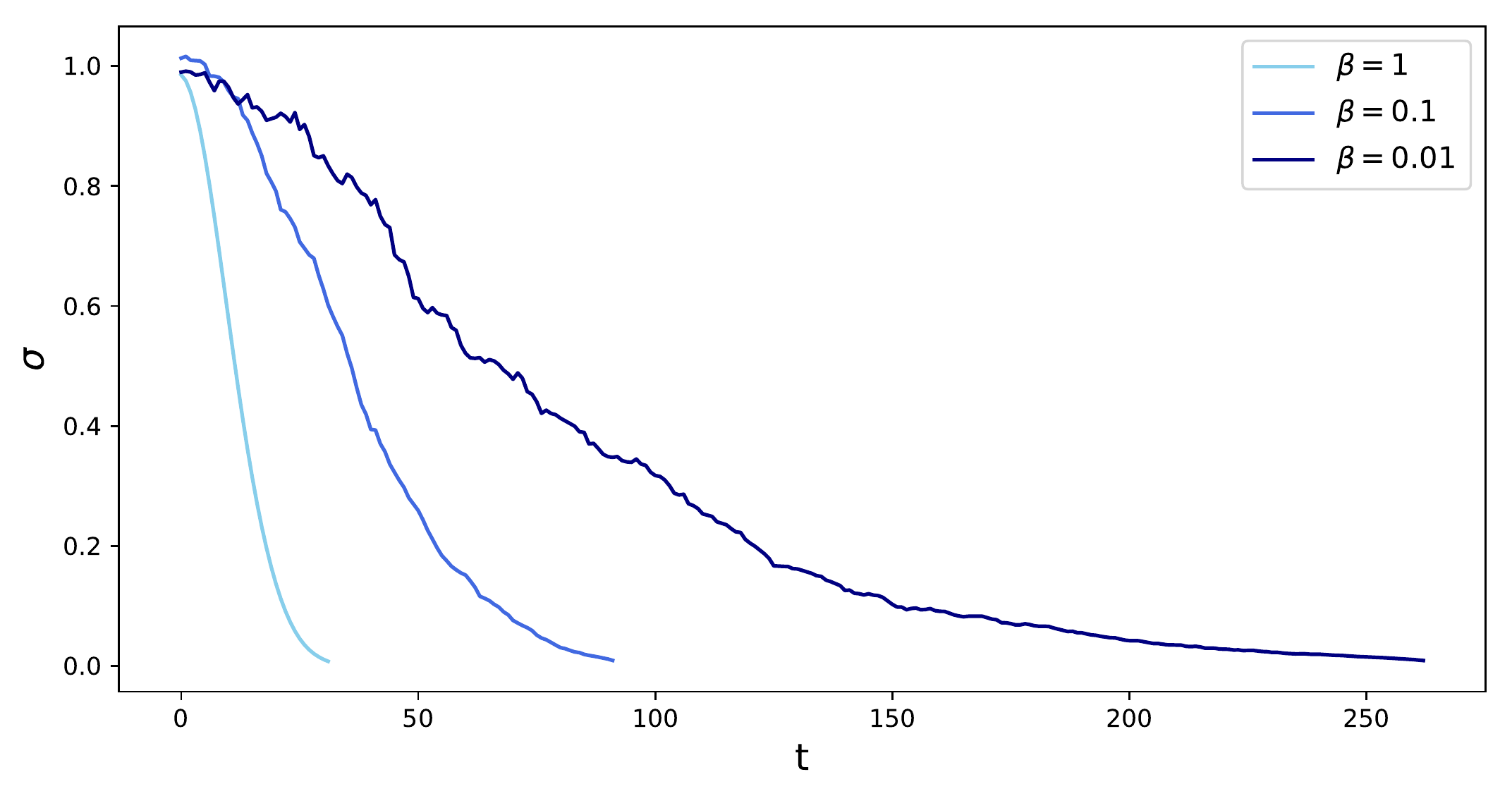} \\[-0.5ex]
\caption{Convergence of $\sigma$ for three synthesized patches with different values of $\beta$. Injecting more noise in each iteration (i.e., smaller $\beta$) slows down the convergence.} 
\label{fig:convergence_sigma}
\end{figure}
}

In addition to the total effective noise, we can compare the evolution of the removed noise versus injected noise. Figure \ref{fig:convergence_parts} shows the reduction in effective standard deviation, $h_t \sigma_{t} = h_t \frac{||d_t||}{\sqrt{N}}$ in each iteration, along with the standard deviation of the added noise, $\gamma_t$. The amount of noise added relative to the amount removed is such that effective noise drops as $\sigma_t = (1-\beta h_t) \sigma_{t-1}$. When $\beta=1$, the addedtive noise is zero, $\gamma_t=0$, and the convergence of $\sigma_t$ is the fastest. When $\beta = 0.01$, a lot of noise is added in each iteration, and the convergence is the slowest.

\begin{table}
  \caption{Run time (in seconds) for different applications, averaged over images in Set12, on an NVIDIA  RTX 8000 GPU.}
  \label{tab:runtime}
  \centering
  \begin{tabular}{lllll}
    \toprule
    Inpainting $30\times 30$  & Missing pixels 10\% &  SISR  4:1  &  Deblurring 10\% &  CS 10\%\\
    \midrule
           Inpainting $30\times 30$  & 7.8 &  SISR  4:1  &  8.4 &   \\
    \bottomrule
  \end{tabular}
\end{table}

\section{Sampling from the implicit prior of a denoiser trained on MNIST}


\begin{figure}[h]
\centering
\def\fsz{0.32\linewidth} \def\nsp{\hspace*{-.005\linewidth}}
\begin{tabular}{ccc}
  \nsp\includegraphics[width=\fsz]{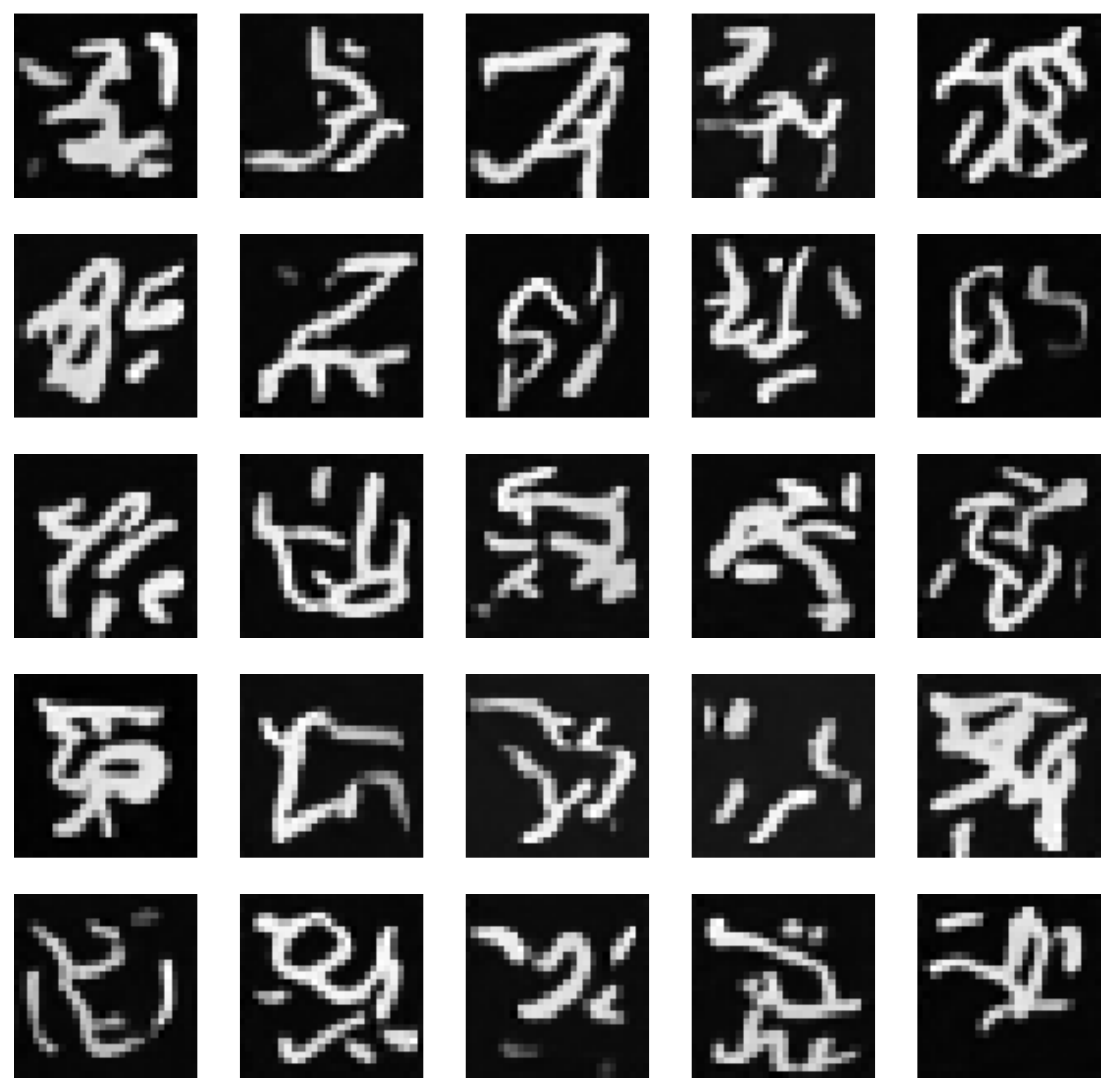}\nsp & 
  \nsp\includegraphics[width=\fsz]{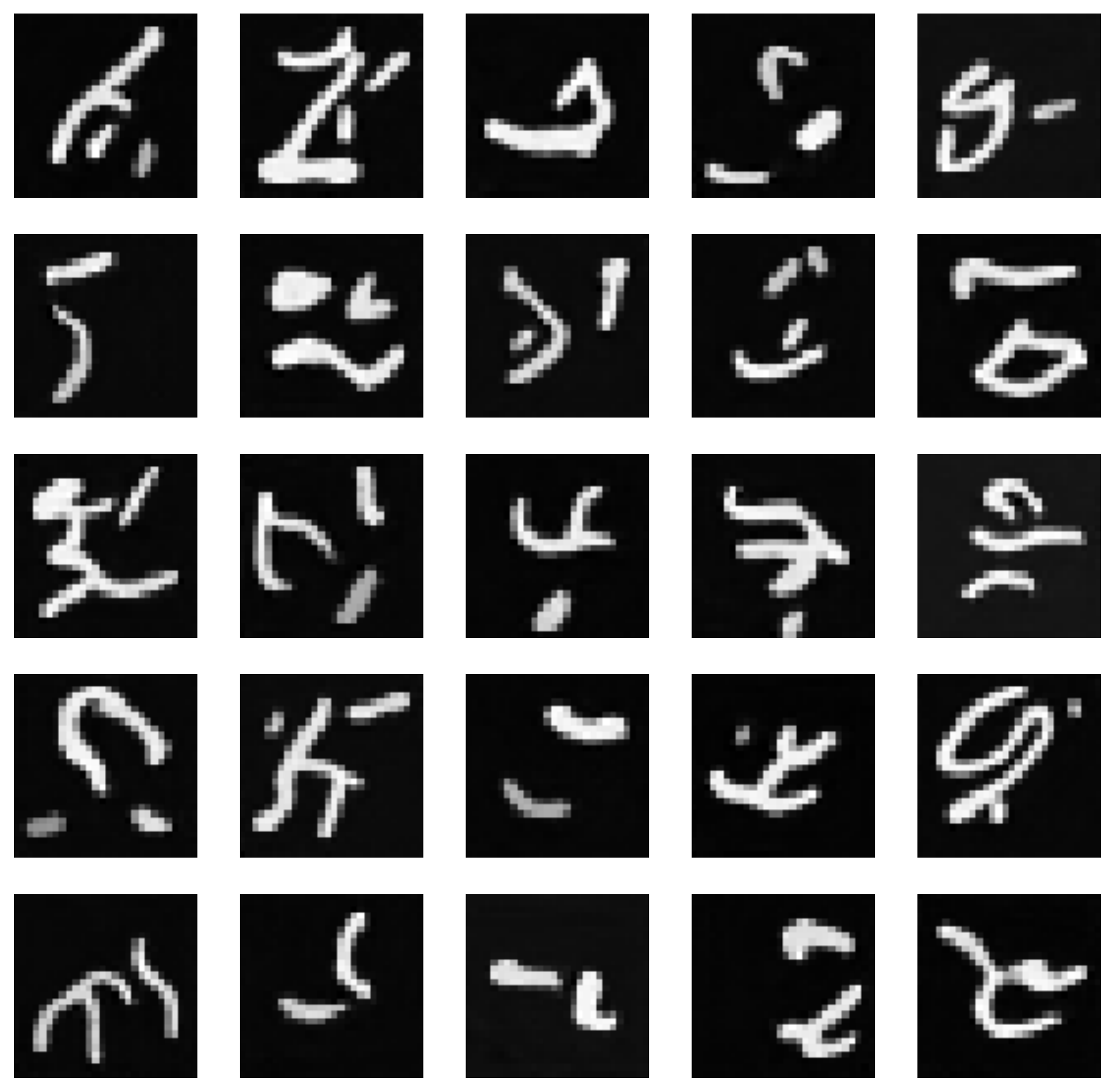}\nsp&
  \nsp\includegraphics[width=\fsz]{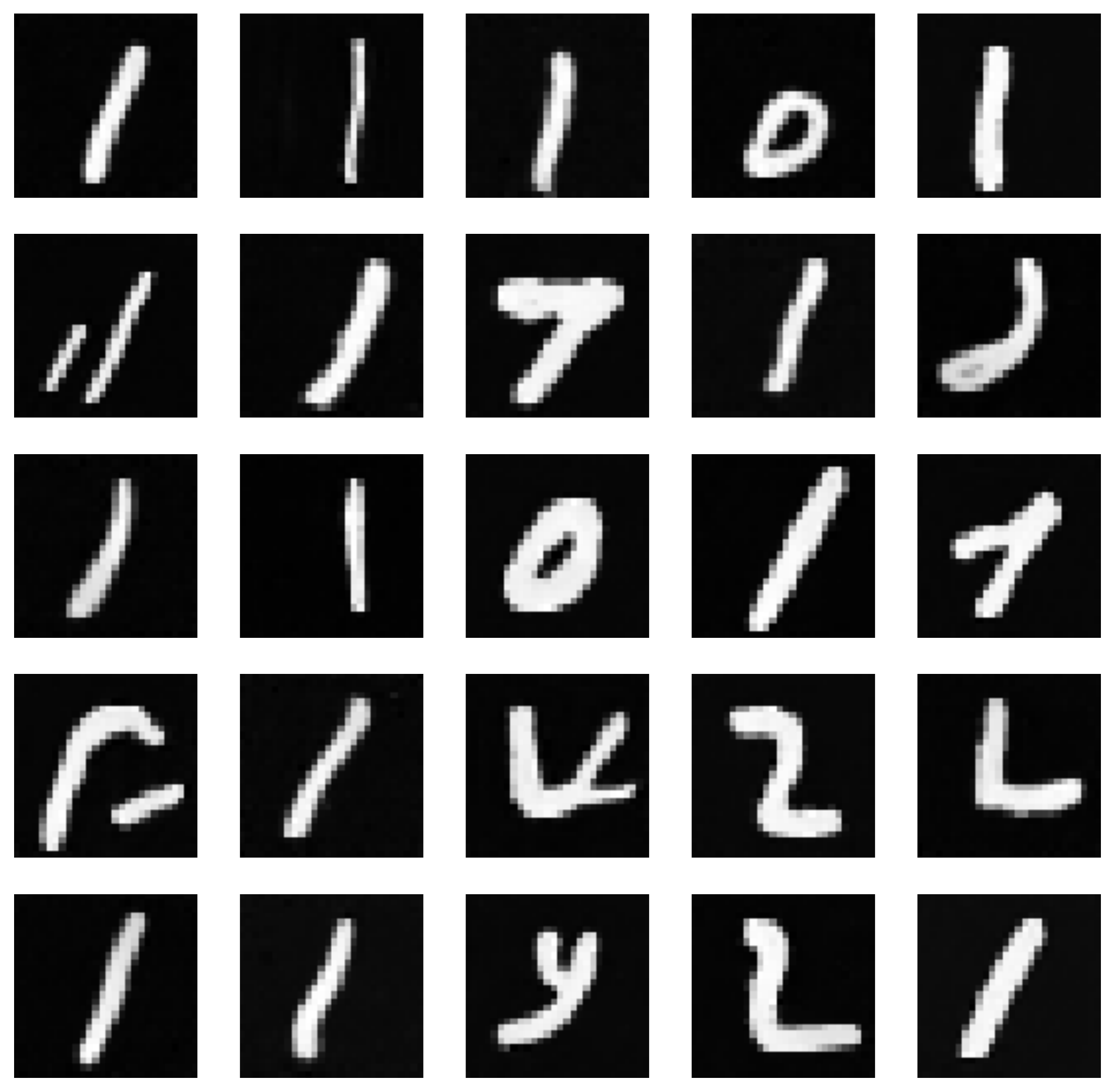}\nsp \\  
\end{tabular}
\caption{Training BF-CNN on the MNIST dataset of handwritten digits~\cite{lecun2010mnist} results in a different implicit prior (compare to Figure~\ref{fig:synthesis_samples}). Each panel shows 16 samples drawn from the implicit prior, with different levels of injected noise (increasing from left to right, $\beta \in \{1.0,\ 0.3,\ 0.01\}$).
} 
\label{fig:mnist_beta}
\end{figure}  